\RequirePackage{fix-cm}
\documentclass[twocolumn]{svjour3}          
\smartqed  
\makeatletter
\def\cl@chapter{\@elt {theorem}}
\makeatother
\usepackage{amsmath}
\usepackage{amssymb}
\usepackage{bm}
\usepackage{multirow}

\usepackage{booktabs}
\usepackage{xspace}
\usepackage{diagbox}

\usepackage[pagebackref,breaklinks,colorlinks]{hyperref}
\usepackage{url}
\usepackage{graphicx}
\usepackage{lipsum}
\usepackage{microtype}
\usepackage{nicefrac}
\usepackage{verbatim}
\usepackage{bbding}
\usepackage{enumitem}

\usepackage{epstopdf}
\usepackage{times}
\usepackage{epsfig}
\usepackage{array}
\usepackage{diagbox}
\usepackage[noblocks]{authblk}
\usepackage{float}
\usepackage[misc]{ifsym}
\usepackage{setspace}

\usepackage[capitalize]{cleveref}
\crefname{section}{Sec.}{Secs.}
\Crefname{section}{Section}{Sections}
\Crefname{table}{Table}{Tables}
\crefname{table}{Tab.}{Tabs.}

\usepackage{xcolor}

\makeatletter
\DeclareRobustCommand\onedot{\futurelet\@let@token\@onedot}
\def\@onedot{\ifx\@let@token.\else.\null\fi\xspace}
\def\eg{\emph{e.g}\onedot} 
\def\ie{\emph{i.e}\onedot}

\def\wrt{w.r.t\onedot} 

\renewcommand{\paragraph}{%
	\@startsection{paragraph}{4}{\z@}%
	{0.1em \@plus 0.5ex \@minus 0.2ex}{-1em}%
	{\normalsize\bf}%
}
\newcommand\rurl[1]{%
  \href{https://#1}{\nolinkurl{#1}}%
}
\makeatother

\definecolor{C1}{RGB}{187,151,39}

\newcommand{\bbm}{\bm{m}}
\newcommand{\bs}{\bm{s}}

\newcommand{\bx}{\bm{x}}
\newcommand{\by}{\bm{y}}

\begin{document}

\title{Universal Representations: A Unified Look at Multiple Task and Domain Learning \thanks{This work was partly supported by the EPSRC programme grant Visual AI EP/T028572/1 and a Huawei Technologies R\&D (UK) project.}
}

\author{Wei-Hong Li$^1$       \and
        Xialei Liu$^{2}$   \and
        Hakan Bilen$^{1}$
}


\institute{\Letter \; w.h.li@ed.ac.uk
           \\
           {$^{1}$ VICO Group, School of Informatics, University of Edinburgh, United Kingdom}
           \\
           {$^{2}$ School of Computer Science, Nankai University, China}
}

\date{Received: date / Accepted: date}

\maketitle

\begin{abstract}
We propose a unified look at jointly learning multiple vision tasks and visual domains through \emph{universal representations}, a single deep neural network. 
Learning multiple problems simultaneously involves minimizing a weighted sum of multiple loss functions with different magnitudes and characteristics and thus results in unbalanced state of one loss dominating the optimization and poor results compared to learning a separate model for each problem.
To this end, we propose distilling knowledge of multiple task/domain-specific networks into a single deep neural network after aligning its representations with the task/domain-specific ones through small capacity adapters.
We rigorously show that universal representations achieve state-of-the-art performances in learning of multiple dense prediction problems in NYU-v2 and Cityscapes, multiple image classification problems from diverse domains in Visual Decathlon Dataset and cross-domain few-shot learning in MetaDataset.
Finally we also conduct multiple analysis through ablation and qualitative studies.
\end{abstract}

\keywords{multi-task learning \and multi-domain learning \and cross-domain few-shot learning \and universal representation learning \and balanced optimization \and dense prediction}

\section{Introduction}\label{sec:intro}

\begin{figure*}[ht!]
\begin{center}
\includegraphics[width=1.0\linewidth]{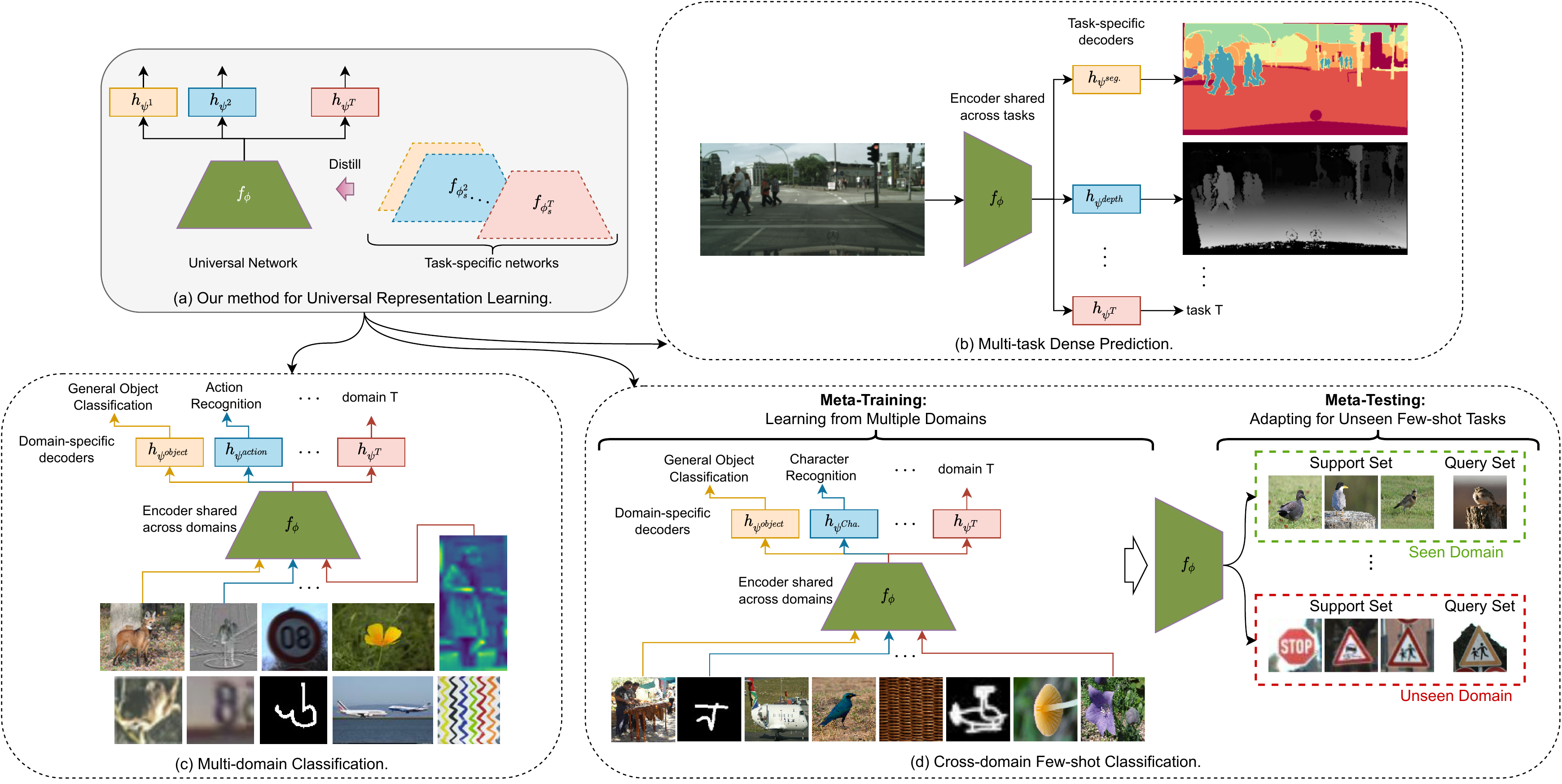}
\end{center}
\vspace{-0.3cm}
\caption{We propose a Universal Representation Learning framework in (a) that generalizes over multi-task dense prediction tasks (b), multi-domain many-shot learning (c), cross-domain few-shot learning (d).}
\label{fig:applications}
\end{figure*}
A major limitation of state-of-the-art image interpretation systems is their narrow scope. 
Different models are learned to recognize faces~\cite{taigman2014deepface,parkhi2015deep,schroff2015facenet,zhong2016faces}, textures~\cite{cimpoi2014describing}, sketches~\cite{eitz2012humans} and drawings~\cite{Lake1332}, various fine-grained flower~\cite{nilsback2008automated}, bird~\cite{wah2011caltech}, fungi categories~\cite{brigit2018fungi}, detect~\cite{ren2015faster,liu2016ssd} and segment~\cite{dai2016instance}  object categories, and perform various low and mid-level tasks such as depth~\cite{eigen2014depth}, surface normal estimation~\cite{wang2015designing}, so on.

In contrast, humans in the early years of their development develop powerful internal visual representations that are subject to small refinements in response to later visual experience \cite{atkinson2002developing,maurer2001visual,lewis2005multiple}.
Once these visual representations are formed, they are \emph{universal} and later employed in many diverse vision tasks from reading text, recognizing faces to interpreting visual art forms.

Presence of universal representations in computer vision~\cite{bilen2017universal}, has important implications.
First, it means that vision has limited complexity.
A growing number of visual domains and tasks\footnote{While domain and and task definitions vary in previous work and are used interchangeably, in our experiments each domain denotes a data domain, dataset such as ImageNet, Omniglot and each task denotes either different prediction task such as semantic segmentation and depth estimation, and also same prediction task such as image classification, albeit, over different sets of categories. A subtle difference between two settings is that for a single image there can be multiple tasks defined, however, only a single domain.} can be modeled with a bounded number of representations.
As a result, one can use a compact set of representations for learning multiple domains and tasks, and efficiently share features and computations across them, which is crucial in platforms with limited computational resources such as mobile devices and autonomous cars.
Second, as we obtain more complete universal representations, learning of new domains and tasks can be easier and performed efficiently from only few samples by transfer learning.

In practice, learning universal representations requires to address several challenges.
First, modelling diverse visual data demands deep network architectures that can simultaneously learn representations while selectively sharing only the relevant representations across multiple tasks and domains. 
To this end, previous multi-task works proposed controlling representation sharing across tasks through latent connections~\cite{misra2016cross,ruder2019latent}, constructing branched deep neural networks based on task affinities~\cite{vandenhende2019branched}, custom attention mechanisms~\cite{liu2019end}, neural architecture search~\cite{liang2018evolutionary,bruggemann2020automated,guo2020learning}, developing progressive communication across multiple tasks through recurrent networks~\cite{bilen2016integrated,zhang2018joint,vandenhende2020mti}, multi-scale feature sharing~\cite{xu2018pad,vandenhende2020mti}.
Other previous works assume that features extracted from pretrained deep networks on ImageNet provide basis for universal representations and adapt them with a set of compact adapters to various domains~\cite{rebuffi2017learning,rebuffi2018efficient,rosenfeld2018incremental,deecke2021visual}.
In cross-domain few-shot classification, where the goal is to generalize to unseen tasks and domains from few samples, features from multiple domain-specific networks are considered as universal features and transferred to previously unseen domains and tasks after a post selection step~\cite{dvornik2020selecting,liu2020universal}.

The second challenge is to develop training algorithms to learn representations that achieve good performance not only in one of the tasks or domains but in all of them. 
This problem is especially visible when the training involves jointly minimizing a set of loss functions (\ie one for each task) with significantly different difficulty levels, magnitudes, and characteristics. 
Thus a naive strategy of uniformly weighing multiple losses can lead to sub-optimal performances and searching for optimal weights in a continuous hyperparameter space can be prohibitively expensive.
Previous works~\cite{chen2018gradnorm,sener2018multi,kendall2018multi,guo2018dynamic,liu2019end} address the \emph{unbalanced loss optimization} problem by weighing loss functions based on the task-dependent uncertainty of the model at training time~\cite{kendall2018multi}, proposing Pareto optimal solution~\cite{sener2018multi}, eliminating conflicting gradient components between the tasks~\cite{yu2020gradient}.
Although these methods are shown to improve over the uniform weighing loss strategy in some benchmarks, they do not consistently outperform the baseline that simply weighs each loss function with a constant scalar~\cite{vandenhende2021multi}.

In this work, we focus on the second challenge.
Inspired from knowledge distillation~\cite{romero2014fitnets,hinton2015distilling}, we approach the problem from a different perspective and propose a general methodology for universal representation learning that can be applied to a diverse set of problems including multi-task and multi-domain learning in few- and many-shot  settings.
We propose a two-stage procedure for universal representation learning where we first train a set of task or domain-specific models and freeze their parameters, and then distill their knowledge to a universal representation network while simultaneously training it over multiple-tasks/domains. 
In contrast to the standard knowledge distillation, in our setting each ``teacher'' network is trained for either significantly different task (\eg semantic segmentation, depth estimation) and/or domain (\eg flowers, handwritten characters), and encodes significantly different representations.
Hence naively distilling their representations into a single network would result in poor performance.
To this end, we propose aligning the universal network with the individual ones via small task-specific adapters before the distillation, and using specific loss functions that are invariant to certain transformations between representations.
Our method has multiple key advantages over the previous work.
First, in contrast to relying solely on weighing the individual loss functions (\eg \cite{kendall2018multi}) or modifying the direction of their gradients (\eg \cite{yu2020gradient}) that are limited to prevent one task dominating or interfering the rest, we propose more explicit control on the model parameters through knowledge distillation such that representations from all tasks/domains are included in the universal representations.
Second, unlike task/domain-specific loss functions with different characteristics which are difficult to balance, the distillation loss function is same for all tasks/domains and hence provides balanced optimization by design.
Third, unlike \cite{dvornik2020selecting,liu2020universal} that employ multiple feature extractors, our model learns a single set of universal representations (a single feature extractor) over multiple domains which has a fixed computational cost regardless of the number of domains at inference.
Finally, our method can be successfully incorporated to various state-of-the-art multi-task/domain customized network architectures~\cite{vandenhende2020mti,rebuffi2018efficient} and loss balancing strategies~\cite{kendall2018multi,liu2021towards,liu2021conflict}.

We illustrate our universal representation learning method and its applications to three standard vision problems in \cref{fig:applications}.
The common step for all the applications is to first train a task or domain specific model and then distill their knowledge to a single universal network (see \cref{fig:applications}(a)).
We show that the universal representations (depicted by the green network) can successfully be employed in jointly learning (i) multiple dense vision problems such as semantic segmentation, depth estimation (see \cref{fig:applications}(b)), (ii) multiple image classification problems from diverse datasets such as ImageNet~\cite{deng2009imagenet}, Omniglot~\cite{Lake1332}, FGVC Aircraft~\cite{maji2013fine} (see \cref{fig:applications}(c)), (iii) learning to classify images from few training samples of unseen tasks and domains (see \cref{fig:applications}(d)).
In all applications, the computations and representations are largely shared across tasks and domains through the universal network, while light-weight task or domain-specific heads are used to obtain the predictions by mapping the universal representations to the task output space.

\quad

In summary, our core contribution is a generic framework for universal representation learning that can be employed in very diverse problems including dense multi-task prediction, and multi-domain many-shot and few-shot image classification tasks.
We show that the learned universal representations generalize well not only to unseen samples in previously seen domains and tasks but also to unseen tasks and domains in few-shot setting.
We rigorously evaluate our method and show that it outperforms the state-of-the-art multi-task dense prediction methods in NYU-v2~\cite{silberman2012indoor} and CityScapes~\cite{cordts2016cityscapes}, cross-domain few-shot image classification methods in MetaDataset~\cite{triantafillou2019meta}, and multi-domain many-shot image classification methods in Visual Decathlon benchmark~\cite{rebuffi2017learning}.
We also propose an efficient learning strategy that enables to learn representations from subsets of tasks in parallel and finally merging them to universal representations. 
We extensively analyze the performance of our method over various design choices including different deep network architectures, adaptor types and loss functions and for knowledge distillation.

This work is an extended version of our prior contributions~\cite{li2020knowledge,li2021universal} that focus on dense multi-task prediction and cross-domain few-shot learning problems. 
The new contributions are:
\begin{itemize}
    \item a unified look at universal representations for a diverse set of problems in multi-domain and multi-task learning,
    \item a hierarchical distillation strategy that allows for learning representations in parallel,
    \item evaluation in a new problem, multi-domain many-shot learning,
    \item evaluation of our method with the state-of-the-art multi-task dense prediction architectures~\cite{xu2018pad,liu2019end,vandenhende2020mti},
    \item more extensive analysis on adaptor types and loss functions,
    \item more extensive and recent literature review.
\end{itemize}

The rest of the paper is organized as follows: \Cref{sec:rel} provides an extensive overview of related work in multi-task learning, multi-domain learning and knowledge distillation.
\Cref{sec:method} reviews the background learning formulation for single and multiple task learning.
\Cref{sec:universal} introduces universal representation learning for multi-task dense prediction, multi-domain classification and cross-domain few-shot classification.
\Cref{sec:exp} provides a rigorous analysis of the design choices and evaluates the performance of the proposed models in multiple standard benchmarks for the three problems.
\Cref{sec:conclusion} concludes the paper with future remarks and limitations.

\section{Related Work}\label{sec:rel}

\subsection{Multi-task Learning}
Multi-task Learning (MTL)~\cite{caruana1997multitask} aims at learning a single model that can infer all desired task outputs given an input. 
We refer to \cite{ruder2017overview,zhang2017survey,vandenhende2021multi} for more comprehensive literature review.
As discussed above, the prior works can be broadly divided into two groups.
The first one focuses on improving network architecture via more effective information sharing across tasks~\cite{misra2016cross,kokkinos2017ubernet,ruder2019latent,liu2019end,vandenhende2019branched,liang2018evolutionary,bruggemann2020automated,guo2020learning,bragman2019stochastic,strezoski2019many,xu2018pad,zhang2019pattern,bruggemann2021exploring,bilen2016integrated,zhang2018joint,vandenhende2020mti,xu2018pad}.
The second group aims to address the unbalanced optimization that is caused by jointly optimizing multiple task loss functions with varying characteristics through either actively changing weight of each loss term~\cite{kendall2018multi,liu2019end,guo2018dynamic,chen2018gradnorm,lin2019pareto,sener2018multi,liu2021towards} and/or modifying the gradients of loss functions w.r.t. the shared network weights to alleviate the conflicts among tasks~\cite{yu2020gradient,liu2021conflict,chen2020just,chennupati2019multinet++,suteu2019regularizing}.

Our method is complementary to the first line of work. In fact, we show in \cref{sec:exp} that our method can be used to boost the state-of-the-art multi-task architectures in dense prediction problems.
While our goal is aligned with the one of the second group, we propose a significantly different strategy based on knowledge distillation to solve the unbalanced loss optimization problem. 
To this end, we first train a task-specific model for each task in an offline stage and freeze their parameters. 
We then train the universal representation (multi-task) network for minimizing task-specific loss and also for producing the same features with the task-specific networks. 
As each task-specific network encodes different features, we introduce small task-specific adapters to project the universal features to the task-specific features and then minimize the discrepancy between task-specific and universal features. 
In contrast to prior works that either rely solely on weighing the individual loss functions (\eg \cite{kendall2018multi}) or modifying their gradients for parameter updates (\eg \cite{yu2020gradient}) that are limited to prevent one task dominating or interfering the rest, our method provides a more direct control on the model parameters through knowledge distillation such that representations from all tasks are included in the universal representations.
Second, unlike task-specific loss functions with different characteristics which are difficult to balance, the distillation loss function is same for all tasks and hence provides a balanced optimization by design.
In addition, in this paper, we show that our method can also generalize to learning multiple diverse visual domains for standard image classification~\cite{rebuffi2017learning} and few-shot classification~\cite{triantafillou2019meta}.

\paragraph*{Multi-task Self-supervised Learning.}
Our work is also loosely related to a recent line of self-supervised learning methods that learns representations from unlabelled data from multiple pretext tasks~\cite{doersch2017multi}, \eg predicting rotation, or learning them from unlabelled data which are sampled from multiple domains (datasets)~\cite{zoph2020rethinking,ghiasi2021multi}. 
Unlike these works that focus on pretraining representations from unlabelled data and transferring them to a downstream task at a time (one model for one downstream task), we focus on learning models that can jointly perform multiple tasks.

\subsection{Multi-domain learning (MDL)}
A parallel line of research is learning representations jointly on multiple domains ~\cite{bilen2017universal,rosenfeld2018incremental,rebuffi2018efficient,deecke2021visual}.
Unlike~\cite{ganin2016domain,tzeng2017adversarial,hoffman2018cycada,xu2018deep,peng2019moment,sun2019unsupervised} that focus on domain adaptation, this line of work aims at learning a single set of universal representations over multiple tasks and visual domains.
Bilen and Vedaldi~\cite{bilen2017universal} proposed to learn a compact multi-domain representation for standard image classification in multiple visual domains using domain-specific scaling parameters. 
This idea was later extended to the use of domain-specific adapters~\cite{rebuffi2017learning,rosenfeld2018incremental,rebuffi2018efficient} and latent domains learning without access to domain annotations by learning gating functions to select domain-specific adapters for the given images~\cite{deecke2021visual}. 
In this paper, we also learn a single set of universal (multi-domain) representations by sharing most of the computation across domains (\eg the feature encoder is shared across all domains, followed by multiple domain-specific classifiers). 
However, unlike \cite{rebuffi2017learning,rosenfeld2018incremental,rebuffi2018efficient,deecke2021visual} that use representations pretrained only on ImageNet~\cite{deng2009imagenet} as the universal ones, and then learn additional domain-specific representations resulting, our method is capable of learning a single set of universal representations from multiple domains which requires less number of parameters and it is also a significantly harder task due to the challenges in the multi-loss optimization. 
In addition, those works do not scale up to multi-task learning in a single domain, as they require running network with the corresponding task-specific adaptors for each task separately, while ours requires only a single forward computation for all tasks.

\subsection{Cross-Domain Few-shot learning}
Our work is also related to the few-shot classification scenarios that aim at adapting a classifier to previously unseen tasks and domains from few labeled samples. 
Earlier works~\cite{koch2015siamese,vinyals2016matching,snell2017prototypical,finn2017model,nichol2018first} focus on evaluating their methods in homogeneous learning tasks, \eg Omniglot~\cite{Lake1332}, miniImageNet~\cite{vinyals2016matching} where both the meta-train and meta-test examples are sampled from a single data distribution (or dataset), and perform poorly in the more challenging cross-domain few-shot tasks, where test data is sampled from an unknown or previously unseen domain~\cite{triantafillou2019meta}.
We refer to \cite{wang2020generalizing,hospedales2020meta} for comprehensive review of early works.

Recent few-shot techniques~\cite{dvornik2020selecting,liu2020universal,bateni2020improved,requeima2019fast} leverage powerful representations learned over multiple domains and focus on few-shot learning from multiple domains that generalizes to unseen domains at test time in the recently proposed MetaDataset~\cite{triantafillou2019meta}.
CNAPS~\cite{requeima2019fast} consists of an adaptation network that modulates the parameters of both a feature extractor and classifier for new categories by encoding the data distribution of few training samples.
Simple CNAPS~\cite{bateni2020improved} extends CNAPS by replacing its parametric classifier with a non-parametric classifier based on Mahalanobis distance and shows that adapting the classifier from few samples is not necessary for good performance.
SUR~\cite{dvornik2020selecting} and URT~\cite{liu2020universal} further show that adaptation for the feature extractor can also be replaced by a feature selection mechanism.
In particular, both~\cite{dvornik2020selecting,liu2020universal} learn a separate deep network for each training dataset in an offline stage, employ them to extract multiple features for each image, and then select the optimal set of features either based on a similarity measure~\cite{dvornik2020selecting} or on an attention mechanism~\cite{liu2020universal}.
Despite their good performance, SUR and URT are computationally expensive and require multiple forward passes through multiple networks during inference time. 
Our method also uses multi-domain features but in a more efficient way, by learning a single network over multiple domains. 
Our method requires significantly less network capacity and compute load than theirs.

\subsection{Knowledge distillation}
Our work is related to knowledge distillation (KD) methods~\cite{hinton2015distilling,li2020knowledge,ma2019graph,phuong2019towards,romero2014fitnets,tian2019contrastive} that distill the knowledge of an ensemble of large teacher models to a small student neural network at the classifier~\cite{hinton2015distilling} and intermediate layers~\cite{romero2014fitnets}.
Born-Again Neural Networks~\cite{furlanello2018born} uses KD proposes to consecutively distill knowledge from an identical teacher network to a student network, which is further applied to few-shot learning in \cite{tian2020rethinking} and multi-task learning in \cite{clark2019bam}.

While our method can also be seen as a distillation method, our goal differs significantly.
In contrast to the standard KD that aims to learn a single task/domain network from multiple teachers, our goal is to learn a multi-task/domain network.
The difference is subtle.
Multi-task/domain learning typically involves solving an unbalanced optimization, while aligning the predictions of the multi-task/domain (student) network with the task-specific (teacher) networks does not necessarily alleviate this issue, as this alignment problem leads to another unbalanced optimization problem due to varying dimensionality of task outputs and loss functions required for matching different tasks' predictions \cite{clark2019bam}, \eg, a kl-divergence loss for classification and l2-norm loss for regression.
While alignment of intermediate representations are studied for KD in \cite{romero2014fitnets}, such an alignment is substantially harder when the representations vary significantly across different teacher networks.
We demonstrate that mapping the student (or universal) representations to teacher's representation space before the alignment is crucial.
Finally, we also show that intermediate representation matching across very diverse domains can indeed be improved by using a loss function that is invariant to linear transformations, inspired from Centered Kernel Alignment (CKA) similarity~\cite{kornblith2019similarity}.

\section{Background}\label{sec:method}

\begin{figure*}[ht!]
\begin{center}
\includegraphics[width=1.0\linewidth]{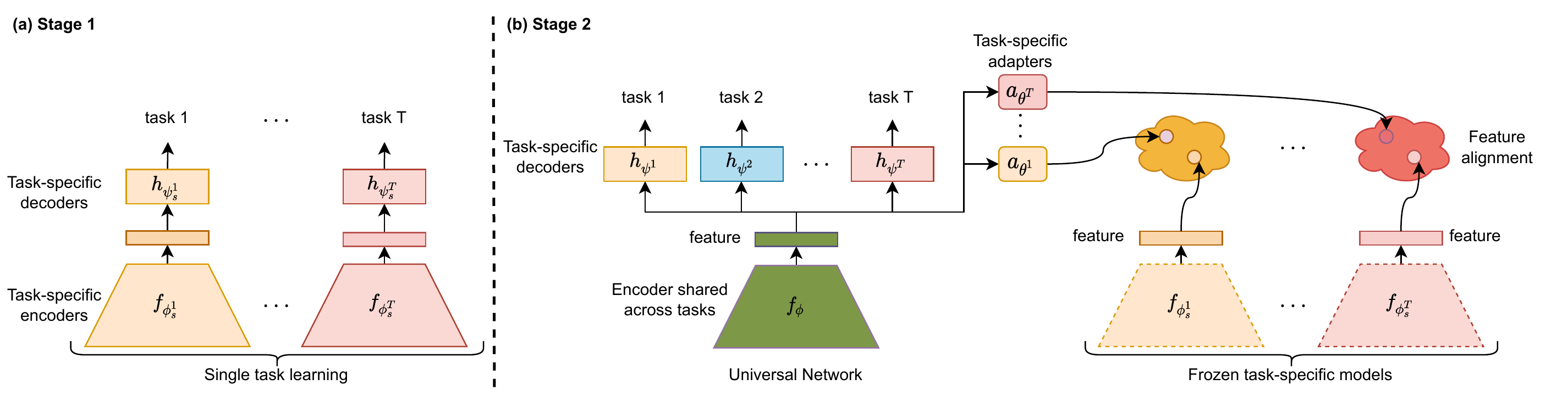}
\end{center}
\vspace{-0.3cm}
\caption{\textbf{Illustration of universal representation learning}. In the first stage (a), we learn a task-specific deep network for each task. In the second stage (b), our goal is to learn a multi-task network that shares the feature encoder across all tasks and build multiple task-specific decoders on top of the feature encoder such that it performs well on these tasks compared to task-specific models trained in (a). To achieve this, we train such multi-task network by jointly minimizing task-specific losses and aligning the feature between the multi-task network and task-specific network. For the feature alignment, we introduce a set of task-specific adapters to transform the feature from multi-task network to task-specific space before the alignment with task-specific features.}
\label{fig:framework}
\end{figure*}

In this section, we review the problem setting for single-task and multi-task learning to provide the required background for the universal representation learning.
Let $\mathcal{D}$ be a training set consisting of $N$ RGB training images and their respective labels.
We consider two general label settings, multi-task learning (MTL) and multi-domain learning (MDL).
In MTL, we assume that training images are sampled from a single distribution, \ie dataset, and each training image $\bx$ is associated with labels for $T$ tasks, $y=\{\by^1,\by^2,\dots,\by^T\}$.
This is a common setting for dense prediction problems where multiple tasks such as semantic segmentation and depth estimation are performed on the same image.
In MDL, the training set contains samples from $T$ different domains, where each image is associated only with a single domain and its domain-specific task.
The standard MDL benchmarks~\cite{rebuffi2017learning,triantafillou2019meta} contain images from diverse datasets (\eg ImageNet, Omniglot, VGG Flowers), each with a classification task over mutually an exclusive set of categories.
Hence, an image $\bx$ associated with domain $t$ is labelled only with the domain-specific task $\by^t$.
In both MTL and MDL settings, our goal is to learn a function $\hat{y}^{t}$ for each task $t$ in MTL and for each domain $t$ in MDL that accurately predicts the ground-truth label $\by^{t}$ of previously unseen images.

Note that while in MTL different tasks involve solving different problems such as semantic segmentation and depth estimation, in MDL they involve solving the same problem, \eg image classification, however over a different set of categories.
We do not focus on other scenarios where images from different domains are associated with the same task (\eg digit recognition from hand-written notes and real street-numbers).

\subsection{Single-task Learning}\label{sec:stl}
Single-task learning (STL) involves learning a task-specific function $\hat{y}^t_s$ \footnote{The subscript $s$ in $\hat{y}^t_s$ indicates single-task learning} independently for each task by optimizing a task-specific loss function $\ell^{t}(\hat{y}^t_s,\by^t)$ (\eg cross-entropy for classification), which measures the mismatch between the ground-truth label and prediction as following:
\begin{equation}\label{eq:singletask}
	\min_{\phi_s^t,\psi_s^t}\frac{1}{N}\sum_{n=1}^{N}\ell^{t}(\hat{y}^{t}_s(\bx_{n}), \by^{t}_{n}), \quad \forall t \in \{1,\dots,T\}
\end{equation} where each $\hat{y}^t_s$ is composed of i) a feature encoder $f_{\phi_s^t}: \mathbb{R}^{3 \times H\times W} \rightarrow \mathbb{R}^{C \times H'\times W'}$ parameterized by $\phi_s^t$ that takes in an $H\times W$ dimensional RGB image and outputs a $H'\times W'$ dimensional feature map with $C$ channels, where $C>3$, $H'<H$ and $W'<W$; ii) a decoder $h_{\psi_s^{t}}: \mathbb{R}^{C \times H'\times W'} \rightarrow \mathbb{R}^{O^{t} \times H^{t} \times W^{t}}$ that decodes the extracted feature to predict the label for the task $t$, \ie $\hat{y}_s^{t}(\bx)=h_{\psi_s^{t}}\circ f_{\phi_s^t}(\bx)$ where $O^{t}$, $H^{t}$, $W^{t}$ are the dimensions of the output space for task $t$ and $\psi_s^{t}$ are its parameters. 

\subsection{Multi-task Learning}\label{sec:mtl}
A more efficient design is to share a significant portion of the computations and parameters across the tasks via a common feature encoder $f_{\phi}: \mathbb{R}^{3 \times H\times W} \rightarrow \mathbb{R}^{C \times H'\times W'}$, \ie convolutional neural network parameterized by $\phi$ that takes in an image $\bx$ and produces a $H'\times W'$ dimensional $C$ feature maps.
The parameters $\phi$ are shared across all the tasks. 
In this setting, $f_{\phi}$ is followed by $T$ task-specific decoders $h_{\psi^{t}}: \mathbb{R}^{C \times H'\times W'} \rightarrow \mathbb{R}^{O^{t} \times H^{t} \times W^{t}}$, each with its own task-specific weights $\psi^{t}$ that decodes the extracted feature to predict the label for the task $t$, \ie $\hat{y}^{t}(\bx^{t})=h_{\psi^{t}}\circ f_{\phi}(\bx^{t})$. 
A common way of learning $\hat{y}^{t}$ for all tasks is to jointly optimize the shared and task-specific parameters as following:
\begin{equation}\label{eq:mtl}
    \min_{\phi, \{\psi^t\}\vert_{t=1}^T}\frac{1}{N}\sum_{n=1}^{N}\sum_{\by^t_n \in y_n}\lambda^t\ell^{t}(\hat{y}^{t}(\bx_{n}), \by^{t}_{n}),
\end{equation} where $\lambda^t$ is a scaling hyperparameter for task $t$ to balance the loss functions among the tasks. 
However, obtaining good universal representations through solving \cref{eq:mtl} is a challenging problem, as it requires to leverage commonalities between the tasks while balancing their loss functions and minimizing interference (negative transfer \cite{chen2018gradnorm,yu2020gradient}) between them. 
Hence solving \cref{eq:mtl} often leads to lower results than the ones of task-specific models, where each task is independently learned. 


\section{Universal Representation Learning}\label{sec:universal}

Motivated by these challenges, previous methods mainly focus on dynamically balancing the loss functions through loss weights ($\lambda^t$) or manipulating gradients of each task \wrt the shared encoder to alleviate the conflicts between them while optimizing \cref{eq:mtl}.
However, as reported in \cite{vandenhende2021multi}, existing solutions fail to improve over carefully tuning these hyperparameters.
We hypothesize that modifying hyperparameters and/or gradients provide only a limited control on the learned representations, and propose a different view on this problem, a two stage procedure inspired by the knowledge distillation methods~\cite{romero2014fitnets,hinton2015distilling}. 
Assuming that single-task learning often performs well when sufficient training data is available, we argue that single-task representations provide powerful representations and hence they provide good approximations to universal representations.

To this end, we first train task-specific deep networks $\{\hat{y}_s^t\}_{t=1}^{T}$ as described in \cref{sec:stl}, where each network consists of a task-specific feature encoder $f_{\phi_s^t}$ and decoder $h_{\psi_s^t}$ with parameters $\phi_s^t$ and $\psi_s^t$ respectively. 
In the second stage, we freeze their weights, the task-specific feature extractors $f_{\phi_s^t}$ and decoders $h_{\psi_s^t}$, and transfer their knowledge to learn a single set of universal representations by minimizing the distance between the task-specific and universal representations for given training samples (see~\cref{fig:framework}).

As minimizing the distance between a single set of universal representations and multiple single-task representations would not yield a satisfactory solution, \ie the average of the single-task representations, we instead first map the universal representations to each task-specific representation space through small task-specific adapters, and then compute the distance in this space.
In addition, we consider minimizing the distance between the outputs of the universal and single-task networks as in \cite{hinton2015distilling}.
With the introduction of these two distillation terms, \cref{eq:mtl} can be rewritten as:
\begin{equation}\label{eq:mtld}
    \begin{aligned}
        & \min_{\phi, \{\psi^t, \theta^t\}\vert_{t=1}^T} \frac{1}{N}\sum_{n=1}^{N}\sum_{\by^t_n \in y_n}\biggl(\lambda^t\ell^{t}(\hat{y}^{t}(\bx_{n}), \by^{t}_{n}) + \\
        & \lambda_f^t\ell_{f}(a_{\theta^{t}} \circ f_{\phi}(\bx^{t}_n), f_{\phi_s^t}(\bx^{t}_n)) + \lambda_{p}^{t}\ell_{p}^t(\hat{y}^t(\bx_{n}^{t}), \hat{y}_s^t(\bx_{n}^{t}))\biggr),
    \end{aligned}
\end{equation} where $\lambda_{f}^{t}$ and $\lambda_{p}^{t}$ are task-specific hyperparameters for distilling representations and predictions respectively.
$a_{\theta^{t}}$: $\mathbb{R}^{C\times H' \times W'}\rightarrow\mathbb{R}^{C \times H' \times W'}$ is the adapter for task $t$ which is parameterized by $\theta^{t}$, $\ell_{f}$ and $\ell_{p}$ are distance functions in the task representation space $t$.
While a single distance function is used for distilling representations (\ie  $\ell_{f}$), distilling predictions may require a task-specific distance function (\ie $\ell_{p}^{t}$). 
We provide these details in \cref{sec:exp}.
The adapters are jointly trained along with the network parameters.
Note that we discard the task-specific networks and adapters at test time, only use the universal network to predict the labels of unseen images.
Hence, the inference time of our method is fixed and does not depend on the number of tasks/domains.

\paragraph*{Hierarchical Distillation with Task Grouping.}
Solving \cref{eq:mtld} requires aligning the universal representations with multiple single-task representations.
Although obtaining single-task representations involves only forward pass of their input, running multiple single-task networks in training can get memory and compute intensive, as the number of tasks ($T$) grows.
Here we propose a hierarchical strategy that enables solving \cref{eq:mtld} in multiple stages by decomposing it in independent optimization problems.
To this end, we first divide all tasks into different task groups which can be done randomly or in accordance with task similarity. 
Here we only explore random selection, as identifying task similarity is a challenging task by itself and an actively studied problem \cite{zamir2018taskonomy}.
For each group $g$, we learn a single network by distilling the single-task representations of the tasks that belong to the group $g$ like in \cref{eq:mtld}.
Importantly each group training can be run in parallel to accelerate the training. 
Once the group-specific networks are learned, we employ them to obtain the final universal representations as in \cref{eq:mtld}.
Note that it is also possible to use a deeper hierarchy with nested groupings.
However, we focus on only single level of grouping to validate the idea.
We evaluate this strategy in the Visual Decathlon benchmark~\cite{rebuffi2017learning} (see \cref{sec:exp:analysis}), as it involves learning representations over a large set of domains.

Next we describe how the universal representations are learned for different scenarios including multi-task dense prediction, multi-domain classification and cross-domain few-shot classification problems.

\subsection{Multi-Task Dense Prediction}\label{sec:lmdpt}
In multi-task dense prediction setting in a single domain, each image $\bx$ is associated with labels for all tasks. 
The spatial dimensions of labels for each task is equal to the image size -- hence it is called dense or pixelwise prediction -- and is the same for all tasks.
We consider semantic segmentation, monocular depth estimation and surface normal prediction in our experiments. 

In this setting, we only minimize the difference between intermediate representations of the universal network and task-specific ones, and do \emph{not} minimize the difference between the predictions of the universal and single-task networks as in~\cite{clark2019bam,hinton2015distilling,romero2014fitnets} and set $\lambda^t_p$ to zero.
In our preliminary experiments, we observed that matching the predictions leads to a significant drop in the final performance.
As each task prediction has different magnitude range and characteristic, we argue that jointly minimizing these distances along with other loss terms in \cref{eq:mtld} leads to a challenging unbalanced optimization.

Let $\bbm^t=a_{\theta^{t}} \circ f_{\phi}(\bx) \in \mathbb{R}^{C \times H'\times W'}$ and $\bs^t=f_{\phi_s^t}(\bx) \in \mathbb{R}^{C \times H'\times W'}$ denote representations obtained from the universal and single-task encoder for a given image $\bx_n$ and task $t$, respectively.
We normalize two feature maps with L2 Norm: $\tilde{\bbm}_{chw}={\bbm_{chw}} / ||\bbm_{\cdot hw}||_2$ and $\tilde{\bs}_{chw}={\bs_{chw}} / ||\bs_{\cdot hw}||_2$, where $\bbm_{chw}$ indicates a hidden unit at $c,h,w$ in $\bbm$. 
We then measure the distance between two normalized feature maps by using the Euclidean distance function for $\ell_{f}$ as following:
\begin{equation}\label{eq:kddense}
	\ell_{f}(\bbm,\bs)=\sum_{w=1}^{W}\sum_{h=1}^{H}\sum_{c=1}^{C}\left\lVert\tilde{\bbm}_{chw}-\tilde{\bs}_{chw}\right\lVert_2^2.
\end{equation} 
We investigate different designs for $a_{\theta}$ and $\ell_{f}$ and show that using linear adapters for $a_{\theta}$ with Euclidean distance function for $\ell_{f}$ obtains the best performance in \cref{sec:exp:analysis}.

\subsection{Multi-Domain Classification}\label{sec:lmvd}
We also consider a multi-domain scenario as in \cite{rebuffi2017learning}, where the training set $\mathcal{D}$ contains $T$ subdatasets, each sampled from a different domain.
Each image is associated with only one domain and hence one task.
The associated task is known in both train and test time as in \cite{rebuffi2017learning}.
Like the multi-dense prediction problem, our goal is to learn a single network with a shared feature encoder $f_{\phi}$ across the domains, thus the tasks.
Unlike the multi-dense prediction problem, the output of the feature encoder is a vector (\ie $H'=1$ and $W'=1$) and also the predictions (\ie $H^t=1$ and $W^t=1$).
In particular, the feature encoder $f_{\phi}$ is a convolutional neural network followed by a average global pooling layer as in~\cite{he2016deep}.

Following the two stage procedure, we first independently train a set of domain-specific deep networks $\{\hat{y}_{s}^{t}\}_{t=1}^{T}$ by \cref{eq:singletask} where each consists of a specific feature encoder $f_{\phi_{s}^{t}}$ and classifier $h_{\psi_{s}^{t}}$ with parameters $\phi_{s}^{t}$ and $\psi_{s}^{t}$ respectively in the first stage. 
Unlike the previous setting that trains multiple task-specific networks on the same training set, this setting involves training each domain network on a different training set from a different domain. 
In the second stage, we then learn the universal network over training images of multiple domains using \cref{eq:mtld}. Rather than setting $\lambda_p$ in~\cref{eq:mtld} to zero as in~\cref{sec:lmdpt}, here we use KL divergence loss for $\ell_p$ in~\cref{eq:mtld} to align predictions of single-task and universal networks.

Though we use domain-specific adapters to map the universal features to the domain-specific space, learning a single set of representations over substantially diverse domains still remains challenging, requires to model complex non-linear relations between and hence a more elaborate distance function ($\ell_f$) than the Euclidean one.
To this end, we propose to adopt the Centered Kernel Alignment (CKA)~\cite{kornblith2019similarity} similarity index with the Radial Basis Function (RBF) kernel that is originally proposed as an analysis tool to measure similarities between neural network representations and shown to be invariant to various transformations, capable of capturing meaningful non-linear similarities between representations of higher dimension than the number of data points.
Differently from the original goal, we use CKA as a loss function to minimize the distance between universal and domain-specific representations rather than an analysis tool.

Next we briefly describe CKA. 
Given a set of images $\{\bx^{t}_1,\dots,\bx^{t}_B\}$, let $\mathbf{M}=[a_{\theta^{t}}\circ f_{\phi}(\bx^{t}_1), \dots, a_{\theta^{t}}\circ f_{\phi}(\bx^{t}_B)]^\top\in \mathrm{R}^{B \times C}$ and $\mathbf{S}=[f_{\phi_{s}^{t}}(\bx^{t}_1), \dots, f_{\phi_{s}^{t}}(\bx^{t}_B)]^\top \in \mathrm{R}^{B \times C}$ denote the features that are computed by the multi-domain network adapted by $a_{\theta^{t}}$ and domain-specific networks respectively.
We first compute the RBF kernel matrices $\mathbf{P}$ and $\mathbf{T}$ of $\mathbf{M}$ and $\mathbf{S}$ respectively and then use two kernel matrices $\mathbf{P}$ and $\mathbf{T}$ to measure CKA similarity between $\mathbf{M}$ and $\mathbf{S}$:
\begin{equation}\label{eq:kd}
    \text{CKA}(\mathbf{M}, \mathbf{S}) = \text{tr}(\mathbf{P}\mathbf{H}\mathbf{T}\mathbf{H})/\sqrt{\text{tr}(\mathbf{P}\mathbf{H}\mathbf{P}\mathbf{H})\text{tr}(\mathbf{T}\mathbf{H}\mathbf{T}\mathbf{H})},
\end{equation} where $\text{tr}(\cdot)$ and $\mathbf{H}$ denote the trace of a matrix and centering matrix $\mathbf{H}_n=\mathbf{I}_n-\frac{1}{n}\mathbf{1}\mathbf{1}^\top$ respectively. 
The loss $\ell_{f}(\mathbf{M}, \mathbf{Y})$ can be derived as $\ell_{f}(\mathbf{M}, \mathbf{S})=1-\text{CKA}(\mathbf{M}, \mathbf{S})$ as \emph{dissimilarity} between the multi-domain and domain-specific features.
As the original CKA similarity requires the computation of the kernel matrices over the whole datasets, which is not scalable to large datasets, we follow \cite{nguyen2020wide} and compute them over each minibatch in our training.
We refer to \cite{kornblith2019similarity,nguyen2020wide} for more details.

\subsection{Cross-Domain Few-shot Classification}\label{sec:mdfsl}
We also apply our method to cross-domain few-shot classification problem that aims at learning to classify samples from a small training set with only few images for each class from an unknown visual domain.
Like the previous setting at \cref{sec:lmvd}, we are given a large training set $\mathcal{D}$ that consists of images from $T$ subdatasets, each focusing on a classification task over a set of mutually exclusive categories.
Unlike the previous case, the goal is not to accurately classify an unseen image that belongs to one of the previously seen categories and domains but to a previously unseen category from either a previously seen or unseen domain, hence it is substantially more challenging.
In particular, for the unseen task, we are given a support set $\mathcal{S}$ that contains few image and label pairs, and a query set $\mathcal{Q}$ that contains samples to be classified.
In other words, we would like to learn a classifier on the support set that can accurately predict the labels of the query set.

As in \cite{dvornik2020selecting,liu2020universal}, we solve this problem in two stages, meta-training and meta-test.
In meta-training, we learn our universal representation network on $\mathcal{D}$ as in \cref{sec:lmvd} by following the two step procedure where we also use CKA loss function (\ie \cref{eq:kd}) for $\ell_f$ and KL divergence loss for $\ell_p$ in \cref{eq:mtld}.
In meta-test, we transfer the learned universal network, further adapt its representations and learn a classifier for the new task on $\mathcal{S}$.
In contrast to \cite{dvornik2020selecting,liu2020universal} that learn multiple domain-specific networks in an offline stage, then employ them to extract multiple features for each image, and then select the most relevant ones at test time, we learn a single universal network that performs well in $T$ domains by distilling the knowledge of the domain-specific networks at train time, but use only the universal network at test time.
This has two key advantages over \cite{dvornik2020selecting,liu2020universal}.
First using a single feature extractor, which has the same capacity with each domain-specific one, is significantly more efficient in terms of run-time and number of parameters in the meta-test stage.
Second learning to find the most relevant features for a given support and query set in \cite{liu2020universal} is not trivial and may also suffer from overfitting to the small number of datasets in the training set, while the multi-domain representations, by definition, automatically contain the required information from the relevant domains.

In particular, after the universal representation learning, we freeze and use the universal network to extract features, and learn a task-specific linear mapping $m_{\varphi}:\mathrm{R}^C\rightarrow \mathrm{R}^C$, which is parameterized by $\varphi$, on the support set $\mathcal{S}$ with the non-parametric nearest centroid classifier (NCC)~\cite{mensink2013distance,snell2017prototypical}:
\begin{equation} \label{eq:adapt}
\min_{\varphi}\frac{1}{\vert\mathcal{S}\vert}\sum_{(\bx, \by) \in \mathcal{S}} \ell_{\text{ce}}(\text{NCC} \circ m_{\varphi} \circ f_{\phi}(\bx),\by)
\end{equation} where $\ell_{\text{ce}}$ is cross-entropy loss.
NCC computes an average of feature vector over the mapped features of support samples that belong to each category to obtain each class centroids, measures the class probability of each sample by applying softmax over its negative cosine distance to the class centroids.

\noindent\textbf{Discussion.} In contrast to the previous setting where the universal representations are learned over multiple domains from sufficiently large data, cross-domain few-shot learning requires obtaining the domain-specific knowledge from only few samples in an unseen domain, which is extremely challenging.
Hence, our hypothesis is that transferring universal representations should yield more effective learning of new domains, as the base assumption is that there is a bounded number of representations for vision problems.

\section{Experiments}\label{sec:exp}

In this section, we analyze and evaluate our method in three problems, i) learning multiple dense prediction tasks on two popular benchmarks, NYU-v2~\cite{silberman2012indoor} and Cityscapes~\cite{cordts2016cityscapes}) in \cref{sec:exp:dense}, ii) learning multiple diverse visual domains on the Visual Domain Decathlon~\cite{rebuffi2017learning} in \cref{sec:exp:vd} and iii) few-shot classification on MetaDataset~\cite{triantafillou2019meta} in \cref{sec:exp:fsl}.
Finally, we conduct extensive analysis over various design choices in \cref{sec:exp:analysis}.
\footnote{The code and models that are used in our experiments are available at \url{https://github.com/VICO-UoE/UniversalRepresentations}.}

\subsection{Learning multiple dense prediction tasks} \label{sec:exp:dense}
Here, we evaluate our method on learning universal representations for performing multiple dense prediction tasks on two standard multi-task learning benchmarks  NYU-v2~\cite{silberman2012indoor} and Cityscapes~\cite{cordts2016cityscapes} as in~\cite{liu2019end,liu2021conflict}. 

\paragraph*{Datasets and experimental setting.}
We follow the training and evaluation settings in~\cite{liu2019end,liu2021conflict} for both single-task and multi-task learning in both datasets. 
More specifically, \emph{NYU-V2}~\cite{silberman2012indoor} contains RGB-D indoor scene images, where we evaluate performances on 3 tasks, including 13-class semantic segmentation, depth estimation, and surface normals estimation. 
We use the true depth data recorded by the Microsoft Kinect and surface normals provided in \cite{eigen2015predicting} for depth and surface normal estimation as in~\cite{liu2019end}. All images are resized to $288 \times 384$ resolution as in \cite{liu2019end}. We follow the default setting in~\cite{silberman2012indoor,liu2019end} where 795 and 654 images are used for training and testing, respectively.
\emph{Cityscapes}~\cite{cordts2016cityscapes} consists of street-view images, which are labeled for two tasks: 7-class semantic segmentation\footnote{The original version of Cityscapes provides labels 7\&19-class semantic segmentation. We follow the 7-class semantic segmentation evaluation protocol as in \cite{liu2019end} to be able to compare to the related works.} and depth estimation. We resize the images to $128 \times 256$ to speed up the training as~\cite{liu2019end}.

In both NYU-v2 and Cityscapes, we follow the training and evaluation protocol in~\cite{liu2019end}.
We apply our method and all the baseline methods to two common multi-task architectures, encoder and decoder based ones. 
The encoder-based methods only share information in the encoder before decoding each task with an independent task-sepcific decoder while the decoder-based approaches also exchange information during the decoding stage~\cite{vandenhende2021multi}.
For the encoder-based, we use the SegNet~\cite{badrinarayanan2017segnet} as the backbone. 
As in ~\cite{liu2019end}, we use cross-entropy loss for semantic segmentation, l1-norm loss for depth estimation in Cityscapes, and cosine similarity loss for surface normal estimation in NYU-v2. 
We use the exactly same hyper-parameters including learning rate, optimizer and also the same evaluation metrics, mean intersection over union (mIoU), absolute error (aErr) and mean error (mErr) in the predicted angles to evaluate the semantic segmentation, depth estimation and surface normals estimation task, respectively in~\cite{liu2019end}. 

For the decoder-based, we build our method on PAD-Net~\cite{xu2018pad} and MTI-Net~\cite{vandenhende2020mti} that use multi-scale feature extractor (encoder) based on the HRNet-18~\cite{sun2019deep} initialized with ImageNet pretrained weight as the feature encoder. 
We use the same loss functions, evaluation metrics, and training and evaluation protocol as done for SegNet backbone. 
For our method, we use the uniform loss weights (\ie $\lambda^t=1$ for all tasks) for task-specific losses, unless stated otherwise. 
As we do \emph{not} minimize the difference between predictions of the universal and single-task networks, we set $\lambda^t_p$ in \cref{eq:mtld} to zero.
We then first split the train set as train and validation set to search $\lambda_f^t \in \{1, 2\}$ by cross-validation and train our network on the whole training set. 
We set $\lambda_f^t$ to 1 for semantic segmentation and depth and 2 for surface normal estimation. Please refer to the supplementary (Sec. A.1) for more details.

\paragraph*{Multi-task performance.}
In addition to the abovementioned evaluation metric for each task, following prior work~\cite{vandenhende2021multi,li2022learning}, we also report the multi-task performance $\bigtriangleup$MTL whih measures the average per-task drop in performance \wrt the single-task baseline:
\begin{equation}\label{eq:mtlp}
\bigtriangleup\text{MTL}=\frac{1}{T}\sum_{t=1}^{T}(-1)^{\ell_t}(P^{t}-P_{s}^{t})/P_{s}^{t},
\end{equation} where $\ell_t=1$ if a lower value of $P_{t}$ means better performance for metric of task $t$, and 0 otherwise. $P^{t}$ and $P_s^{t}$ are performance (\eg mIoU for semantic Segmentation) of the universal (multi-task) network and single-task network, respectively.

\subsubsection{Encoder-based Architecture}

\paragraph*{Compared methods.} 
Encoder-based architectures, including the vanilla MTL using \emph{SegNet}~\cite{badrinarayanan2017segnet} that shares the whole feature encoder across all tasks and consists of task-specific decoders, and \emph{MTAN}~\cite{liu2019end} which extends the vanilla MTL baseline by sharing the SegNet across tasks and using task-specific attention modules in each layer to extract task-specific features.
We compare our method to the single-task learning (STL) baseline, \ie train individual network per task, the vanilla multi-task learning network with uniform loss weights (MTL), and balanced optimization strategies, including Uncertainty~\cite{kendall2018multi}, GradNorm~\cite{chen2018gradnorm}, MGDA~\cite{sener2018multi}, DWA~\cite{liu2019end}, PCGrad~\cite{yu2020gradient}, GradDrop~\cite{chen2020just}, IMTL~\cite{liu2021towards} and CAGrad~\cite{liu2021conflict}. 
We also consider the BAM~\cite{clark2019bam} which is originally designed for natural language processing and adapt this method for visual dense prediction tasks by aligning dense predictions.
Importantly, this model performs knowledge distillation on predictions when learning the multi-task network and hence comparing this method sheds light onto importance of matching intermediate representations in the task-specific spaces.
Here we use KL-divergence loss, l1-norm loss and cosine similarity loss as knowledge distillation loss on predictions for semantic segmentation, depth estimation and surface normal estimation, respectively. 
We reproduce all methods in the same settings for fair comparison and the results of the compared methods are similar or better than the ones reported in the corresponding papers.

\paragraph*{Results on NYU-v2.}

\Cref{tab:nyuv2} depicts the results of our method and other compared approaches in NYU-v2. 
We see that the vanilla MTL (Uniform) using SegNet achieves better performance in depth estimation, however, its performance drops in surface normal estimation in comparison with STL. 
This indicates that joint optimization of multiple tasks with uniform loss weights leads to unbalanced results, and overall worse performance than STL models in $\Delta$MTL metric. 
While only few balanced optimization algorithms help to improve the MTL performance, IMTL-H and CAGrad obtains the best when applied to SegNet and MTAN. IMTL-H improves by balancing the pace at which tasks are learned via looking at the projection onto individual tasks of the average gradient \wrt the shared parameters while CAGrad achieves improvement by modifying the parameter update such that the update not only minimizes the average of task-specific losses but also decreases each task-specific loss. 
While BAM optimizing the multi-task learning network by knowledge distillation, it performs worse than the Uniform baseline as it aligns the predictions which requires to use different loss functions (\eg cross-entropy for segmentation) and requires solving another unbalanced optimization problem. 
This shows that simply distilling the predictions of the multiple single task network leads to poor performance.
Finally, our method outperforms these methods with either SegNet or MTAN backbones significantly, +7.84\% MTL performance improvement over the IMTL-H, the best baseline.
The results suggest that distilling features from multiple single-task networks provides a more effective learning of shared representations.

\begin{table}[h!]
	\centering
    \resizebox{0.46\textwidth}{!}
    {
		\begin{tabular}{clccccccccccc}

		    \toprule
		    Arch & Method & Seg. (mIoU) $\uparrow$ & Depth (aErr) $\downarrow$ & Norm. (mErr) $\downarrow$ & $\bigtriangleup \text{MTL}$ $\uparrow$ \\
		    \midrule
		    \multirow{13}{*}{SegNet} & STL & 40.54 & 0.6276 & {\bf 24.28} & +0.00 \\
		    \cmidrule{2-6}
		    & Uniform & 40.22 & 0.5196 & 29.09 & -1.13 \\
		    & Uncertainty~\cite{kendall2018multi} & 37.35 & 0.5014 & 26.74 & +0.72 \\
		    & GradNorm~\cite{chen2018gradnorm} & 40.15 & 0.5824 & 27.70 & -2.60 \\
		    & MGDA~\cite{sener2018multi} & 39.77 & 0.5669 & 29.05 & -3.95 \\
		    & DWA~\cite{liu2019end} & 40.27 & 0.5247 & 28.71 & -0.82 \\
		    & PCGrad~\cite{yu2020gradient} & 39.55 & 0.5236 & 28.54 & -1.13 \\
		    & GradDrop~\cite{chen2020just} & 39.25 & 0.5226 & 29.41 & -2.52 \\
		    & IMTL-H~\cite{liu2021towards} & 40.62 & 0.5224 & 26.14 & +3.11 \\
		    & CAGrad~\cite{liu2021conflict} & 37.99 & 0.5196 & 25.77 & +1.61 \\
		    & BAM~\cite{clark2019bam} & 37.72 & 0.5571 & 28.58 & -4.47 \\
		    \cmidrule{2-6}
		    & Ours & {\bf 45.52} & {\bf 0.4912} & 24.57 & {\bf +10.95} \\ 
		    \midrule
		    \multirow{13}{*}{MTAN} & STL & 38.69 & 0.5701 & 24.85 & +0.00 \\
		    \cmidrule{2-6}
		    & Uniform & 40.60 & 0.5238 & 26.87 & +1.65 \\
		    & Uncertainty~\cite{kendall2018multi} & 39.59 & 0.5382 & 26.21 & +0.83 \\
		    & GradNorm~\cite{chen2018gradnorm} & 38.97 & 0.5269 & 26.22 & +0.94 \\
		    & MGDA~\cite{sener2018multi} & 40.08 & 0.5410 & 26.84 & +0.23 \\
		    & DWA~\cite{liu2019end} & 40.34 & 0.5418 & 27.65 & -0.68 \\
		    & PCGrad~\cite{yu2020gradient} & 39.42 & 0.5290 & 27.17 & -0.07 \\
		    & GradDrop~\cite{chen2020just} & 39.19 & 0.5552 & 27.35 & -2.05  \\
		    & IMTL-H~\cite{liu2021towards} & 41.12 & 0.5200 & 25.73 & +3.86 \\
		    & CAGrad~\cite{liu2021conflict} & 42.17 & 0.5227 & 25.42 & +5.01  \\
		    & BAM~\cite{clark2019bam} & 40.49 & 0.5412 & 26.84 & +0.58 \\
		    \cmidrule{2-6}
		    & Ours & {\bf 43.91} & {\bf 0.5019} & {\bf 24.58} & {\bf +8.85} \\
			\bottomrule
		\end{tabular}%
			}
		\caption{Test performance on NYU-v2. We evaluate single task learning (STL) method and multi-task learning methods (MTL) on NYU-v2. Mean intersection over union (mIoU) for semantic segmentation, absolute error (aErr) for depth estimation, mean error (mErr) for surface normal estimation and multi-task performance ($\bigtriangleup \text{MTL}$) are reported.}
		\label{tab:nyuv2}
\end{table}%

\paragraph*{Results on Cityscapes.}
We also evaluate all methods in Cityscapes and report the results in \cref{tab:citys}. 
Similar to the results in NYU-v2, BAM obtains worse results compared with the STL methods (\eg -0.58\% MTL performance when using the vanilla MTL method with SegNet).
Among the loss balancing methods, Uncertainty and IMTL-H obtain the best MTL performance in both backbones (\ie SegNet and MTAN) but their performance is lower than the STL models in both tasks. 
This shows the difficulty of optimizing the MTL network in a balanced way in this problem. 
Our method obtains significant gains on both tasks than all the compared methods and also achieves better results than the STL results. 
The results again demonstrate that our method is able to optimize MTL model in a more balanced way and to achieve better overall results. In addition, our method has much less parameters (one network) than the STL models (two networks) in Cityscapes.

\begin{table}[h!]
	\centering
	
    \resizebox{0.46\textwidth}{!}
    {
		\begin{tabular}{clccccc}

		    \toprule
		     Arch & Method & Seg. (mIoU) $\uparrow$ & Depth (aErr) $\downarrow$ & $\bigtriangleup \text{MTL}$ $\uparrow$ \\
		    \midrule
		    \multirow{13}{*}{SegNet} & STL & 74.19 & 0.0122 & +0.00  \\
		    \cmidrule{2-5}
		    & Uniform & 73.82 & 0.0126 & -0.74 \\
		    & Uncertainty~\cite{kendall2018multi} & 72.74 & 0.0123 & -0.29 \\
		    & GradNorm~\cite{chen2018gradnorm} & 73.67 & 0.0130 & -2.75 \\
		    & MGDA~\cite{sener2018multi} & 73.86 & 0.0130 & -2.43 \\
		    & DWA~\cite{liu2019end} & 73.51 & 0.0126 & -1.13 \\
		    & PCGrad~\cite{yu2020gradient} & 73.55 & 0.0126 & -1.16 \\
		    & GradDrop~\cite{chen2020just} & 73.09 & 0.0125 & -0.96 \\
		    & IMTL-H~\cite{liu2021towards} & 73.26 & 0.0124 & -0.56 \\
		    & CAGrad~\cite{liu2021conflict} & 74.50 & 0.0136 & -4.34 \\
		    & BAM~\cite{clark2019bam} & 74.02 & 0.0123 & -0.58 \\
		    \cmidrule{2-5}
		    & Ours & {\bf 75.53} & {\bf 0.0119} & {\bf +2.21} \\
		    \midrule
		    \multirow{13}{*}{MTAN}& STL & 75.92 & 0.0119 & +0.00 \\
		    \cmidrule{2-5}
		    & Uniform & 75.31 & 0.0119 & -0.56 \\
		    & Uncertainty~\cite{kendall2018multi} & 74.95 & 0.0121 & -1.73 \\
		    & GradNorm~\cite{chen2018gradnorm} & 74.88 & 0.0123 & -2.70 \\
		    & MGDA~\cite{sener2018multi} & 75.84 & 0.0129 & -4.65 \\
		    & DWA~\cite{liu2019end} & 75.39 & 0.0121 & -1.42 \\
		    & PCGrad~\cite{yu2020gradient} & 75.62 & 0.0122 & -1.73 \\
		    & GradDrop~\cite{chen2020just} & 75.69 & 0.0123 & -2.15 \\
		    & IMTL-H~\cite{liu2021towards} & 75.33 & 0.0120 & -1.20 \\
		    & CAGrad~\cite{liu2021conflict} & 75.45 & 0.0124 & -2.69 \\
		    & BAM~\cite{clark2019bam} & 75.74 & 0.0122 & -1.44 \\
		    \cmidrule{2-5}
		    & Ours & {\bf 76.42} & {\bf 0.0117} & {\bf +0.81} \\
			\bottomrule
		\end{tabular}%
			}
		\caption{Testing results on Cityscapes. We evaluate single task learning (STL) method and multi-task learning methods (MTL) on Cityscapes. Mean intersection over union (mIoU) for semantic segmentation, absolute error (aErr) for depth estimation and multi-task performance ($\bigtriangleup \text{MTL}$) are reported.}
		\label{tab:citys}
\end{table}%

\paragraph*{Incorporating loss balancing to ours.}
While our method achieves consistent improvements over all the target tasks, solving \cref{eq:mtld} also involves minimizing a weighted sum of multiple loss terms.
Hence here we investigate whether our method can also benefit from dynamically setting weights of the individual loss terms in NYU-v2.
In particular, we use SegNet as backbone, and we dynamically update the weights of task-specific losses (\ie $\lambda^t$) with keeping the weight of distillation loss fixed ($\lambda_f$). 
We evaluate our method with each of three best performing loss balancing methods, \ie Uncertainty, IMTL-H and CAGrad and report the results in \cref{tab:nyuv2lossbalance}. 
We see that our method is complementary to these loss balancing methods and it significantly improves the performance of loss balancing methods (\eg Ours (Uncertainty) obtains about +12 improvement in MTL performance over the Uncertainty).
Also, we can see that by applying our method to loss balancing methods obtains better performance than using our method with the Uniform MTL baseline.

\begin{table}[h!]
	\centering
    \resizebox{0.46\textwidth}{!}
    {
		\begin{tabular}{clccccccccccc}

		    \toprule
		    Arch & Method & Seg. (mIoU) $\uparrow$ & Depth (aErr) $\downarrow$ & Norm. (mErr) $\downarrow$ & $\bigtriangleup \text{MTL}$ $\uparrow$ \\
		    \midrule
		    \multirow{10}{*}{SegNet} & STL & 40.54 & 0.6276 & 24.28 & +0.00 \\
		    \cmidrule{2-6}
		    & Uniform & 40.22 & 0.5196 & 29.09 & -1.13 \\
		    & Ours (Uniform) & {\bf 45.52} & {\bf 0.4912} & {\bf 24.57} & {\bf +10.95} \\ 
		    \cmidrule{2-6}
		    & Uncertainty~\cite{kendall2018multi} & 37.35 & 0.5014 & 26.74 & +0.72 \\
		    & Ours (Uncertainty) & {\bf 46.03} & {\bf 0.4780} & {\bf 24.04} & {\bf +12.80} \\ 
		    \cmidrule{2-6}
		    & IMTL-H~\cite{liu2021towards} & 40.62 & 0.5224 & 26.14 & +3.11 \\
		    & Ours (IMTL-H) & {\bf 46.33} & {\bf 0.5000} & {\bf 23.65} & {\bf +12.41} \\
		    \cmidrule{2-6}
		    & CAGrad~\cite{liu2021conflict} & 37.99 & 0.5196 & 25.77 & +1.61 \\
		    & Ours (CAGrad) & {\bf 45.58} & {\bf 0.5059} & {\bf 23.68} & {\bf +11.44} \\
			\bottomrule
		\end{tabular}%
			}
		\caption{Testing results on NYU-v2. We evaluate single task learning (STL) method and multi-task learning methods (MTL) on NYU-v2. Mean intersection over union (mIoU) for semantic segmentation, absolute error (aErr) for depth estimation, mean error (mErr) for surface normal estimation and multi-task performance ($\bigtriangleup \text{MTL}$) are reported.}
		\label{tab:nyuv2lossbalance}
\end{table}%

\subsubsection{Decoder-based Architectures}
We also apply our method to the decoder-based methods, PAD-Net~\cite{xu2018pad} and MTI-Net~\cite{vandenhende2020mti} which are particularly designed for MTL by exchanging information during the decoding stage and achieve state-of-the-art performances in MTL~\cite{vandenhende2021multi}.
Apart from these results, we also include results of the vanilla MTL method using the same backbone (HRNet-18~\cite{sun2019deep}) of PAD-Net and MTI-Net as baseline and we report all results in NYU-v2 and Cityscapes in \cref{tab:nyuv2dec} and \cref{tab:citysdec}.

\begin{table}[h!]
	\centering
    \resizebox{0.46\textwidth}{!}
    {
		\begin{tabular}{lcccc}

		    \toprule
		    Method & Seg. (mIoU) $\uparrow$ & Depth (aErr) $\downarrow$ & Norm. (mErr) $\downarrow$ & $\bigtriangleup \text{MTL}$ $\uparrow$ \\
		    \midrule
		    STL & 53.07 & 0.3608 & 20.90 & +0.00 \\
		    \midrule
		    MTL & 53.39 & 0.3626 & 23.35 & -3.87 \\
		    Ours (MTL) & {\bf 54.03} & {\bf 0.3565} & {\bf 21.43} & {\bf +0.15} \\
		    \midrule
		    PAD-Net~\cite{xu2018pad} & 54.07 & 0.3583 & 22.60 & -1.85 \\
		    Ours (PAD-Net) & {\bf 54.75} & {\bf 0.3537} & {\bf 21.80} & {\bf +0.28} \\
		    \midrule
		    MTI-Net~\cite{vandenhende2020mti} & 54.59 & 0.3353 & 21.90 & +1.72 \\
		    Ours (MTI-Net) & {\bf 56.48} & {\bf 0.3317} & {\bf 21.36} & {\bf +4.11} \\
			\bottomrule
		\end{tabular}%
			}
		\caption{Testing results on NYU-v2. We evaluate single task learning (STL) method and decoder-based multi-task learning methods (MTL) with HRNet on NYU-v2. Mean intersection over union (mIoU) for semantic segmentation, absolute error (aErr) for depth estimation, mean error (mErr) for surface normal estimation and multi-task performance ($\bigtriangleup \text{MTL}$) are reported.}
		\label{tab:nyuv2dec}
\end{table}%

\begin{table}[h!]
	\centering
    \resizebox{0.46\textwidth}{!}
    {
		\begin{tabular}{lccccc}

		    \toprule
		    Method & Seg. (mIoU) $\uparrow$ & Depth $\downarrow$ & $\bigtriangleup \text{MTL}$ $\uparrow$ \\
		    \midrule
		    STL & 76.92 & 0.0111 & +0.00\\
		    \midrule
		    MTL & 76.67 & 0.0115 & -1.86 \\
		    Ours (MTL) & {\bf 77.20} & {\bf 0.0111} & {\bf +0.06} \\
		    \midrule
		    PAD-Net~\cite{xu2018pad} & 77.83 & 0.0109 & +1.47 \\
		    Ours (PAD-Net) & {\bf 77.84} & {\bf 0.0107} & {\bf +2.48} \\
		    \midrule
		    MTI-Net~\cite{vandenhende2020mti} & 76.61 & 0.0111 & -0.17 \\
		    Ours (MTI-Net) & {\bf 77.10} & {\bf 0.0109} & {\bf +0.77} \\
			\bottomrule
		\end{tabular}%
			}
		\caption{Testing results on Cityscapes. We evaluate single task learning (STL) method and decoder-based multi-task learning methods (MTL) with HRNet on Cityscapes. Mean intersection over union (mIoU) for semantic segmentation, absolute error (aErr) for depth estimation and multi-task performance ($\bigtriangleup \text{MTL}$) are reported.
		}
		\label{tab:citysdec}
\end{table}%

First we see that decoder based methods achieves better performance than the encoder based ones, as they use more powerful customized architectures and initialized with pre-trained ImageNet weights.
Similar to encoder-based methods, the vanilla MTL method obtains better performance in Segmentation than STL while it performs worse in depth and surface normal estimation than STL models in NYU-v2 due to the unbalanced optimization. In Cityscapes, it performs worse in both tasks than STL baselines. 
Our method when applied to the vanilla MTL method using HRNet-18 backbone improves the performance over the vanilla MTL method and achieves a balanced MTL performance (\ie better or comparable results than STL) in both datasets.

We see in \cref{tab:nyuv2dec} and \cref{tab:citysdec} that the decoder-based methods (PAD-Net and MTI-Net) obtain better performance than the vanilla MTL method by first employing a multi-task network to make initial task predictions, and then leveraging features from these initial predictions to improve each task output (MTI-Net obtains +1.72 MTL performance in NYU-v2). 
Here, PAD-Net improves over the vanilla MTL by aggregating information from the initial task predictions of other tasks by spatial attention for estimating the final task output while MTI-Net extends the PAD-Net to a multi-scale procedure by making initial task predictions and distilling information at each individual scale (of feature).
However, they still suffer from the unbalanced optimization problem (\eg MTI-Net obtains worse performance in surface normal estimation in NYU-v2 and semantic segmentation in Cityscapes, respectively). 
Building our method on these decoder-based method helps to boost their performance (Ours (MTI-Net) vs MTI-Net: +4.11\% vs +1.72\%). 
These results indicate that our method can be used with various architectures and enable more balanced performance over multiple tasks, and boost their overall performance.

\begin{table*}[t!]
	\centering
    \resizebox{0.93\textwidth}{!}
    {
		\begin{tabular}{c|c|cccccccccc|c|c|c}

		    \toprule
		    Model & \#params & ImNet & Airc. & C100 & DPed & DTD & GTSR & Flwr & OGlt & SVHN & UCF & avg $\uparrow$ & S $\uparrow$ & $\bigtriangleup \text{MDL}$ $\uparrow$ \\
		    \#images &  & 1.3m & 7k & 50k & 30k & 4k & 40k & 2k & 26k & 70k & 9k & & &  \\
		    \midrule
		    Feature (ImNet)~\cite{rebuffi2017learning} & 1 & 59.67 & 23.31 & 63.11 & 80.33 & 45.37 & 68.16 & 73.69 & 58.79 & 43.54 & 26.80 & 54.28 & 544 & -\ - \\
		    Scratch~\cite{rebuffi2017learning} & 10 & 59.87 & 57.10 & 75.73 & 91.20 & 37.77 & 96.55 & 56.30 & 88.74 & 96.63 & 43.27 & 70.23 & 1625 & -\ - \\
		    Finetune~\cite{rebuffi2017learning} & 10 & 59.87 & 60.34 & 82.12 & 92.82 & 55.53 & 97.53 & 81.41 & 87.69 & 96.55 & 51.20 & 76.51 & 2500 & +0.00 \\
		    \midrule
		    Serial RA~\cite{rebuffi2017learning} & 2 & 59.67 & 56.68 & 81.20 & 93.88 & 50.85 & 97.05 & 66.24 & 89.62 & 96.13 & 47.45 & 73.88 & 2118 & -3.95 \\
		    DAN~\cite{rosenfeld2018incremental} & 2.17 & 57.74 & 64.12 & 80.07 & 91.30 & 56.54 & 98.46 & 86.05 & 89.67 & 96.77 & 49.38 & 77.01 & 2851 & +0.60 \\
		    Piggyback~\cite{mallya2018piggyback} & 1.28 & 57.69 & 65.29 & 79.87 & 96.99 & 57.45 & 97.27 & 79.09 & 87.63 & {\bf 97.24} & 47.48 & 76.60 & 2838 & +0.00 \\
		    MDL & 1 & 59.30 & 67.21 & 79.60 & 97.08 & 57.66 & 97.28 & 83.88 & 87.41 & 96.39 & 45.76 & 77.16 & 2866 & +0.75 \\
		    Parallel RA~\cite{rebuffi2018efficient} & 2 & 60.32 & 64.21 & 81.91 & 94.73 & 58.83 & {\bf 99.38} & 84.68 & 89.21 & 96.54 & 50.94 & 78.07 & 3412 & +2.20 \\
		    Parallel RA SVD~\cite{rebuffi2018efficient} & 1.5 & 60.32 & 66.04 & 81.86 & 94.23 & 57.82 & 99.24 & 85.74 & 89.25 & 96.62 & {\bf 52.50} & 78.36 & 3398 & +2.70 \\
		    \midrule
		    Ours & 1 & 61.44 & 74.59 & 80.10 & 97.20 & 59.31 & 98.31 & 86.00 & 89.71 & 96.90 & 48.96 & 79.25 & 3560 & +4.00 \\
		    Ours (RA) & 2 & {\bf 61.97} & {\bf 77.17} & {\bf 82.31} & {\bf 97.58} & {\bf 60.69} & 98.75 & {\bf 87.28} & {\bf 90.40} & 97.14 & 51.89 & {\bf 80.52} & {\bf 4005} & {\bf +5.96} \\
			\bottomrule
		\end{tabular}%
			}
		\caption{Universal Representation Learning on Visual Decathlon. Accuracy on the test sets of individual dataset, average accuracy of 10 datasets (avg), evaluation score (S), multi-domain learning performance ($\bigtriangleup \text{MDL}$) and the number of parameters (\#params) \wrt a single task network are reported.}
		\label{tab:vd}
\end{table*}%

\subsection{Multi-domain Learning} \label{sec:exp:vd}
Here we evaluate our method on learning universal representations for multiple image classification tasks over multiple diverse domains in Visual Decathlon Benchmark~\cite{rebuffi2017learning}.

\paragraph*{Dataset.} The \emph{Visual Decathlon Benchmark}~\cite{rebuffi2017learning} consists of 10 different well-known datasets:  including ILSVRC\_2012~(ImNet)~\cite{russakovsky2015imagenet}, FGVC-Aircraft~(Airc.)~\cite{maji2013fine}, CIFAR-100~(C100)~\cite{krizhevsky2009learning}, Daimler Mono Pedestrian Classification Benchmark~(DPed)~\cite{munder2006experimental}, Describable Texture Dataset (DTD)~\cite{cimpoi2014describing}, German Traffic Sign Recognition~(GTSR)~\cite{houben2013detection}, Flowers102~(Flwr)~\cite{nilsback2008automated}, Omniglot~(OGlt)~\cite{Lake1332}, Street View House Numbers~(SVHN)~\cite{netzer2011reading}, UCF101~(UCF)~\cite{soomro2012dataset}. 
In this benchmark~\cite{rebuffi2017learning}, images are resized to a common resolution of roughly 72 pixels to accelerate training and evaluation by the organizers.

\paragraph*{Implementation details.} We follow~\cite{rebuffi2017learning,rebuffi2018efficient}, use the official train/val/test splits, evaluation protocol, also use the ResNet-26~\cite{he2016deep} as the backbone for domain-specific network and universal network. 
In our universal network, the backbone (\ie ResNet-26) is shared across all domains and followed by domain-specific linear classifiers. 
We use the same data augmentation (random crop, flipping) and SGD as optimizer, and train domain-specific networks and our universal network for 120 epochs as in~\cite{rebuffi2017learning,rebuffi2018efficient}. 
Here we set the loss weights to 1 (\ie $\lambda^t=1$ for all tasks) and perform cross-validation to search loss weights ($\lambda^t_f,\lambda^t_p$) in $\{0.1, 1, 10\}$ for knowledge distillations on features and predictions, and set $\lambda_f$ and $\lambda_p$ to 10 for ImageNet, 0.1 for DPed, and 1 for other datasets. Please refer to the supplementary (Sec. A.2) for more details.

\paragraph*{Results.}
In Visual Decathlon, we compare our method to \emph{Feature} \ie a feature extractor on ImageNet and learn classifiers on top of the feature extractor for other domains, and single domain learning models that are learned from \emph{Scratch} or \emph{Finetune} from the ImageNet pretrained feature extractor. We also compared our method with existing approaches, including Serial Residual Adapters (RA)~\cite{rebuffi2017learning}, Parallel RA and Parallel RA SVD~\cite{rebuffi2018efficient}, DAN~\cite{rosenfeld2018incremental} and Piggyback~\cite{mallya2018piggyback}. Results are from the corresponding papers.

We report the results on the test split on each domains by the official online evaluation~\cite{rebuffi2018efficient} in~\cref{tab:vd}, including testing accuracy in individual datasets, average accuracy over 10 datasets (avg), decathlon evaluation score (S)~\cite{rebuffi2018efficient}, number of parameters (\#params) \wrt one single task network. We also consider the multi-domain performance (\ie $\bigtriangleup \text{MDL}$) as described in \cref{eq:mtlp}. First, while using ImageNet features requires only 1$\times$ parameters, they do not generalize well to other datasets when large domain gap is present (\eg SVHN). 
In contrast, single-task learning model obtained by either learning from scratch or finetuning achieves significantly better performance (\eg Finetune obtains 76.51 average accuracy and 2500 score) with the expense of 10 times more parameters. 

We use Finetune as the baseline as in~\cite{rebuffi2017learning,rebuffi2018efficient} and compute $\bigtriangleup \text{MDL}$ metric for existing methods and ours. 
We can see that Serial RA which learns a set of domain-specific residual adapters for each task with a ImageNet pretrained feature extractor greatly reduce the number of parameter to 2$\times$ while it obtains slightly worse performance than Finetune (\eg 73.88 vs 76.51 average accuracy for Serial RA and Finetune, respectively ). 
The performance is further improved by DAN which constrains newly learned filters to be linear combinations of existing ones when adapting a pretrained model for other domains, \ie DAN obtains 77.01 average accuracy, +0.60 MDL performance and only requires 2.17 parameters). 
Piggyback learns binary domain-specific masks to select effective filters to adapt a pretrained model for each domains (it obtains 76.60 average accuracy, +0.00 MDL performance and further reduce the number of parameters to 1.28). 
Connecting the RAs in parallel to the backbone (Parallel RAs) boosts the performance of the serial configuration while keeping the same computation cost (78.07 in average accuracy and +2.20 in MDL performance). 
The authors show that the performance can be further improved by decomposing residual adapters to low rank adapters through (78.36 average accuracy, +2.70 MDL performance and 1.5 parameters). 

\begin{table*}[t!]
	\centering
    \resizebox{1.0\textwidth}{!}
    {
		\begin{tabular}{cccccccccc}

		    \toprule
		    \multirow{2}{*}{Test Dataset} & Proto-MAML & BOHB-E & CNAPS & Simple CNAPS & SUR & URT & \multirow{2}{*}{Best SDL} & \multirow{2}{*}{MDL} & \multirow{2}{*}{Ours} \\
		     & \cite{triantafillou2019meta} & \cite{saikia2020optimized} & \cite{requeima2019fast} & \cite{bateni2020improved} & \cite{dvornik2020selecting} & \cite{liu2020universal} & & & \\
		    \midrule
		    ImageNet & $46.5\pm1.1$ & $51.9\pm1.1$ & $50.8\pm1.1$ & $58.4\pm1.1$ & $56.2\pm1.0$ & $56.8\pm1.1$ & $55.8\pm1.0$ & $53.4\pm1.1$ & ${\bf 58.8\pm1.1}$  \\
			Omniglot & $82.7\pm1.0$ & $67.6\pm1.2$ & $91.7\pm0.5$ & $91.6\pm0.6$ & $94.1\pm0.4$ & $94.2\pm0.4$ & $93.2\pm0.5$ & $93.8\pm0.4$ & ${\bf 94.5\pm0.4}$  \\
			Aircraft & $75.2\pm0.8$ & $54.1\pm0.9$ & $83.7\pm0.6$ & $82.0\pm0.7$ & $85.5\pm0.5$ & $85.8\pm0.5$ & $85.7\pm0.5$ & $86.6\pm0.5$ & ${\bf 89.4\pm0.4}$  \\
			Birds & $69.9\pm1.0$ & $70.7\pm0.9$ & $73.6\pm0.9$ & $74.8\pm0.9$ & $71.0\pm1.0$ & $76.2\pm0.8$ & $71.2\pm0.9$ & $78.5\pm0.8$ & ${\bf 80.7\pm0.8}$  \\
			Textures & $68.2\pm0.8$ & $68.3\pm0.8$ & $59.5\pm0.7$ & $68.8\pm0.9$ & $71.0\pm0.8$ & $71.6\pm0.7$ & $73.0\pm0.6$ & $71.4\pm0.7$ & ${\bf 77.2\pm0.7}$  \\
			Quick Draw & $66.8\pm0.9$ & $50.3\pm1.0$ & $74.7\pm0.8$ & $76.5\pm0.8$ & $81.8\pm0.6$ & $82.4\pm0.6$ & ${\bf 82.8\pm0.6}$ & $81.5\pm0.6$ & $82.5\pm0.6$  \\
			Fungi & $42.0\pm1.2$ & $41.4\pm1.1$ & $50.2\pm1.1$ & $46.6\pm1.0$ & $64.3\pm0.9$ & $64.0\pm1.0$ & $65.8\pm0.9$ & $61.9\pm1.0$ & ${\bf 68.1\pm0.9}$  \\
			VGG Flower & $88.7\pm0.7$ & $87.3\pm0.6$ & $88.9\pm0.5$ & $90.5\pm0.5$ & $82.9\pm0.8$ & $87.9\pm0.6$ & $87.0\pm0.6$ & $88.7\pm0.6$ & ${\bf 92.0\pm0.5}$  \\
			Traffic Sign & $52.4\pm1.1$ & $51.8\pm1.0$ & $56.5\pm1.1$ & $57.2\pm1.0$ & $51.0\pm1.1$ & $48.2\pm1.1$ & $47.4\pm1.1$ & $51.0\pm1.0$ & ${\bf 63.3\pm1.1}$  \\
			MSCOCO & $41.7\pm1.1$ & $48.0\pm1.0$ & $39.4\pm1.0$ & $48.9\pm1.1$ & $52.0\pm1.1$ & $51.5\pm1.1$ & $53.5\pm1.0$ & $49.6\pm1.1$ & ${\bf 57.3\pm1.0}$  \\
			MNIST & - & - & - & $94.6\pm0.4$ & $94.3\pm0.4$ & $90.6\pm0.5$ & $89.8\pm0.5$ & $94.4\pm0.3$ & ${\bf 94.7\pm0.4}$  \\
			CIFAR-10 & - & - & - & ${\bf 74.9\pm0.7}$ & $66.5\pm0.9$ & $67.0\pm0.8$ & $67.3\pm0.8$ & $66.7\pm0.8$ & $74.2\pm0.8$  \\
			CIFAR-100 & - & - & - & $61.3\pm1.1$ & $56.9\pm1.1$ & $57.3\pm1.0$ & $56.6\pm0.9$ & $53.6\pm1.0$ & ${\bf 63.5\pm1.0}$  \\
		    \midrule
		    Average Rank & 7.8 & 8.1 & 6.6 & 5.2 & 5.0 & 4.4 & 4.8 & 4.6 & 1.3 \\
			\bottomrule
		\end{tabular}%
			}
		\caption{\textbf{Comparison to baselines and state-of-the-art methods on MetaDataset}. Mean accuracy, 95\% confidence interval are reported. The first eight datasets are seen during training and the last five datasets are unseen and used for test only. Average rank is computed according to first 10 datasets as some methods do not report results on last three datasets.}
		\label{tab:fslcurrmethod}
\end{table*}%

Finally, we show that our method (Ours) successfully learns a single feature extractor shared across all domains only with 1$\times$ parameters, the same number of parameters with a single domain network.
Our model obtains better results than the Finetune baseline and existing methods in most domains (Ours obtains 79.25 average accuracy and +4.00 MDL performance). 
This clearly shows that learning representations from all domains jointly produces more general features than ImageNet representations.
However, this is challenging due to the optimization issues.
The difference between our model and vanilla MDL model shows that simply optimizing over multiple domain-specific loss functions is not sufficient to obtain good representations and representation distillation is crucial.
We also show that RAs can be incorporated to our universal network. 
Jointly learning a shared ResNet-26 backbone with residual adapters (\ie Ours (RA)) boosts the performance, \eg Ours (RA) obtains the best or second best performance in most datasets (\eg Airc., DTD, etc), best average accuracy (80.52), best MDL performance (+5.96) while only requires 2 in parameters cost and best score (4005). 

\subsection{Cross-domain Few-shot Learning} \label{sec:exp:fsl}
Here, we evaluate our method to few-shot learning on recent MetaDataset~\cite{triantafillou2019meta}.

\paragraph*{Dataset.} The \emph{MetaDataset}~\cite{triantafillou2019meta} is a few-shot classification benchmark that initially consisted of ten datasets: ILSVRC\_2012 (ImageNet)~\cite{russakovsky2015imagenet},  Omniglot~\cite{Lake1332}, FGVC-Aircraft (Aircraft)~\cite{maji2013fine}, CUB-200-2011 (Birds)~\cite{wah2011caltech}, Describable Textures (DTD)~\cite{cimpoi2014describing}, QuickDraw~\cite{jongejan2016quick}, FGVCx Fungi (Fungi)~\cite{brigit2018fungi}, VGG Flower (Flower)~\cite{nilsback2008automated}, Traffic Signs~\cite{houben2013detection} and MSCOCO~\cite{lin2014microsoft} was further expanded to 13 datasets with the addition of MNIST~\cite{lecun1998gradient}, CIFAR-10/-100~\cite{krizhevsky2009learning}.

We follow the standard procedure in~\cite{triantafillou2019meta} and use the first eight datasets for meta-training, in which each dataset is further divided into train, validation and test set with disjoint classes. 
While the evaluation within these datasets is used to measure the generalization ability in the seen domains, the remaining five datasets are reserved as unseen domains in meta-test for measuring the cross-domain generalization ability.

\paragraph*{Implementation details.}
In all experiments we build our method on ResNet-18~\cite{he2016deep} backbone for both single-domain and multi-domain networks.
In the multi-domain network, we share all the layers but the last classifier across the domains.
For training single-domain models, we strictly follow the training protocol in~\cite{dvornik2020selecting}, use a SGD optimizer with a momentum and the cosine annealing learning scheduler with the same hyperparameters.
For our multi-domain network, we use the same optimizer and scheduler as before, train it for 240,000 iterations.
We set $\lambda_f$ and $\lambda_p$ in \cref{eq:mtld} to 4 for ImageNet and 1 for other datasets and use early-stopping based on cross-validation over the validations sets of 8 training datasets.
We refer to supplementary (Sec. A.3) for more details.

\begin{table*}[t!]
	\centering
    \resizebox{1.0\textwidth}{!}
    {
		\begin{tabular}{c|cc|cc|cc|cc|cc|cc|cc|cc||cc|cc|cc|cc|cc}

		    \toprule
		    Test Dataset & \multicolumn{2}{c|}{ImageNet} & \multicolumn{2}{c|}{Omniglot} & \multicolumn{2}{c|}{Aircraft} & \multicolumn{2}{c|}{Birds} & \multicolumn{2}{c|}{Textures} & \multicolumn{2}{c|}{Quick Draw} & \multicolumn{2}{c|}{Fungi} & \multicolumn{2}{c||}{VGG Flower} & \multicolumn{2}{c|}{Traffic Sign} & \multicolumn{2}{c|}{MSCOCO} & \multicolumn{2}{c|}{MNIST} & \multicolumn{2}{c|}{CIFAR-10} & \multicolumn{2}{c}{CIFAR-100}\\
		    \midrule
		    Recall@$k$ & 1 & 2 & 1 & 2 & 1 & 2 & 1 & 2 & 1 & 2 & 1 & 2 & 1 & 2 & 1 & 2 & 1 & 2 & 1 & 2 & 1 & 2 & 1 & 2 & 1 & 2 \\
		    \midrule
		    Sum & $22.1$ & $30.3$ & $84.7$ & $91.8$ & $69.7$ & $80.7$ & $45.9$ & $59.7$ & $66.3$ & $78.2$ & $77.4$ & $84.3$ & $31.9$ & $42.9$ & $85.1$ & $92.1$ & $94.6$ & $97.2$ & $62.6$ & $71.2$ & $98.3$ & $99.2$ & $54.0$ & $68.9$ & $27.8$ & $37.4$ \\
		    Concate & $20.2$ & $28.0$ & $84.4$ & $91.5$ & $44.3$ & $58.1$ & $35.5$ & $48.8$ & $68.8$ & $78.2$ & $73.0$ & $80.8$ & $30.7$ & $40.4$ & $83.4$ & $91.3$ & ${\bf 95.1}$ & ${\bf 97.3}$ & $60.7$ & $69.8$ & ${\bf 98.7}$ & ${\bf 99.3}$ & $49.7$ & $65.3$ & $25.4$ & $34.6$ \\
		    MDL & $29.8$ & $39.6$ & ${\bf 89.8}$ & ${\bf 94.3}$ & $80.3$ & $87.1$ & $63.2$ & $75.9$ & $67.0$ & $77.1$ & $79.5$ & $85.4$ & $40.2$ & $51.7$ & $86.9$ & $93.3$ & $89.5$ & $94.1$ & $63.6$ & $72.6$ & $97.6$ & $98.8$ & $58.9$ & $72.9$ & $31.6$ & $42.0$ \\
		    Simple CNAPS~\cite{bateni2020improved} & $34.0$ & $43.8$ & $84.9$ & $91.6$ & $70.5$ & $82.5$ & $55.9$ & $70.5$ & $64.8$ & $76.9$ & $75.3$ & $83.0$ & $29.1$ & $39.0$ & $88.1$ & $94.1$ & $79.9$ & $86.9$ & $65.2$ & $73.8$ & $97.5$ & $98.8$ & ${\bf 66.2}$ & ${\bf 79.3}$ & $33.2$ & $44.2$ \\
		    Ours & ${\bf 36.1}$ & ${\bf 46.2}$ & $89.7$ & ${\bf 94.3}$ & ${\bf 83.3}$ & ${\bf 90.4}$ & ${\bf 66.7}$ & ${\bf 78.9}$ & ${\bf 70.2}$ & ${\bf 80.8}$ & ${\bf 79.9}$ & ${\bf 86.5}$ & ${\bf 44.5}$ & ${\bf 56.2}$ & ${\bf 90.0}$ & ${\bf 94.6}$ & $87.9$ & $93.0$ & ${\bf 67.4}$ & ${\bf 76.3}$ & $97.0$ & $98.4$ & $62.1$ & $76.5$ & ${\bf 35.1}$ & ${\bf 46.1}$ \\
			\bottomrule
		\end{tabular}%
			}
		\caption{\textbf{Global retrieval performance on MetaDataset}. Here we evaluate our method in a non-episodic retrieval task to further compare the generalization ability of our universal representations. }
		\label{tab:fslrecall}
\end{table*}%

\paragraph*{Baselines and compared methods.}
First we compare our method to our own baselines, i) the best single-domain model (Best SDL) where we use each single-domain network as the feature extractor and test it for few-shot classification in each dataset and pick the best performing model (see supplementary Sec. B.2.1 for the complete results). This involves evaluating 8 single-domain networks on 13 datasets, serves a very competitive baseline, ii) the vanilla multi-domain learning baseline (MDL) that is learned by optimizing \cref{eq:mtl} without the proposed distillation method.
As additional baselines, we include the best performing method in~\cite{triantafillou2019meta}, \ie Proto-MAML~\cite{triantafillou2019meta}, and as well as the state-of-the-art methods, BOHB-E~\cite{saikia2020optimized}, CNAPS~\cite{requeima2019fast}, SUR~\cite{dvornik2020selecting}, URT~\cite{liu2020universal}, and the Simple CNAPS~\cite{bateni2020improved}\footnote{Results of Proto-MAML~\cite{triantafillou2019meta}, BOHB-E~\cite{saikia2020optimized}, and CNAPS~\cite{requeima2019fast} are obtained from \href{https://github.com/google-research/MetaDataset}{MetaDataset}.}.
For evaluation, we follow the standard protocol in~\cite{triantafillou2019meta}, randomly sample 600 tasks for each dataset, and report average accuracy and 95\% confidence score in all experiments. 
We reproduce results by training and evaluating SUR~\cite{dvornik2020selecting}, URT~\cite{liu2020universal}, and Simple CNAPS~\cite{bateni2020improved} using their code for fair comparison as recommended by \href{https://github.com/google-research/MetaDataset}{MetaDataset}.

\paragraph*{Results on MetaDataset.}
As described in \cite{triantafillou2019meta}, we sample each task with varying number of ways and shots and report the results in \cref{tab:fslcurrmethod}.
Our method outperforms the state-of-the-art methods in seven out of eight seen datasets and four out of five unseen datasets. 
We also compute average rank as recommended in~\cite{triantafillou2019meta}, our method ranks 1.3 in average and the state-of-the-art SUR and URT rank 5.0 and 4.4, respectively.
In detail, we obtain significantly better results than the second best approach on Aircraft (+2.8), Birds (+2.1), Texture (+4.2), and VGG Flower (+1.5) for seen domains and Traffic Sign (+6.1)\footnote{The accuracy of all methods on Traffic Sign is different from the one in the original papers as one bug has been fixed in MetaDataset repository. See \url{https://github.com/google-research/MetaDataset/issues/54} for more details. As mentioned in the MetaDataset repository, we further update the evaluation protocol and report the updated results of all methods in the supplementary (Sec. B.2.4).} and MSCOCO (+3.8). 
The results show that jointly learning a single set of representations provides better generalization ability than fusing the ones from multiple single-domain feature extractors as done in SUR and URT.
Notably, our method requires less parameters and computations to run during inference than SUR and URT, as it runs only one universal network to extract features, while both SUR and URT need to pass the query set to multiple single-domain networks.

We also see that our method outperforms two strong baselines, Best SDL and MDL in all datasets except in QuickDraw.
This indicates that i) universal representations are superior to the single-domain ones while generalizing to new tasks in both seen and unseen domains, while requiring significantly less number of parameters (1 vs 8 neural networks), ii) our distillation strategy is essential to obtain good multi-domain representations.
While MDL outperforms the best SDL in certain domains by transferring representations across them, its performance is lower in other domains than SDL, possibly due to negative transfer across the significantly diverse domains.
Surprisingly, MDL achieves the third best in average rank, indicating the benefit of multi-domain representations.

\paragraph*{Global retrieval.}
Here we go beyond the few-shot classification experiments and evaluate the generalization ability of our representations that are learned in the multi-domain network in a retrieval task, inspired from metric learning literature~\cite{oh2016deep,yu2019learning}.
To this end, for each test image, we find the nearest  images in entire test set in the feature space and test whether they correspond to the same category.
For evaluation metric, we use Recall@$k$ which considers the predictions with one of the $k$ closest neighbors with the same label as positive.
In \cref{tab:fslrecall}, we compare our method with Simple CNAPS in Recall@1 and Recall@2 (see supplementary Sec. B.2.7 for more results). URT and SUR require adaptation using support set and no such adaptation in retrieval task is possible, we replace them with two baselines that concatenate or sum features from multiple domain-specific networks. 
Our method achieves the best performance in ten out of thirteen domains with significant gains in Aircraft, Birds, Textures and Fungi.
This strongly suggests that our multi-domain representations are the key to the success of our method in the previous few-shot classification tasks. We also provide additional experiments in supplementary (Sec. B.2).

\begin{table*}[t!]
	\centering
	
    \resizebox{0.93\textwidth}{!}
    {
		\begin{tabular}{c|c|cccccccccc|c|c|c}

		    \toprule
		    Model & \#params & ImNet & Airc. & C100 & DPed & DTD & GTSR & Flwr & OGlt & SVHN & UCF & avg $\uparrow$ & S $\uparrow$ & $\bigtriangleup \text{MDL}$ $\uparrow$ \\
		    \#images &  & 1.3m & 7k & 50k & 30k & 4k & 40k & 2k & 26k & 70k & 9k & & &  \\
		    \midrule
		    Ours (group 1) & 1 & 61.59 & 74.11 & {\bf 80.41} & {\bf 97.91} & 58.09 & {\bf 98.57} & {\bf 86.78} & 89.80 & 96.99 & {\bf 50.15} & {\bf 79.44} & {\bf 3759} & {\bf +4.22} \\ 
		    Ours (group 2) & 1 & {\bf 61.65} & 74.44 & 80.01 & 97.20 & 58.78 & 98.22 & 86.60 & 89.54 & {\bf 97.01} & 49.91 & 79.33 & 3570 & +4.15 \\ 
		    Ours (group 3) & 1 & 61.20 & 75.58 & 79.92 & 97.29 & 58.24 & 98.06 & 86.14 & {\bf 89.92} & 96.83 & 49.41 & 79.26 & 3513 & +4.02 \\ 
		    \midrule
		    Ours & 1 & 61.44 & {\bf 74.59} & 80.10 & 97.20 & {\bf 59.31} & 98.31 & 86.00 & 89.71 & 96.90 & 48.96 & 79.25 & 3560 & +4.00 \\
			\bottomrule
		\end{tabular}%
			}
		\caption{Universal Representation Learning with task grouping on Visual Decathlon Benchmark. Accuracy on the test sets of individual dataset, average accuracy of 10 datasets (avg), evaluation score (S), multi-domain learning performance ($\bigtriangleup \text{MDL}$) and the number of parameters (\#params) \wrt a single task network are reported.}
		\label{tab:vdtg}
\end{table*}%

\subsection{Hierarchical Distillation with Task Grouping}\label{sec:exp:group}
As introduced in \cref{sec:universal}, here we first randomly group tasks and learn a single network per group, and then distill their knowledge to learn universal representations. 
In addition to training of networks per task, networks per group can be trained in parallel.
This strategy allows our method to better scale to the scenarios with a big number of tasks and domains by requiring few group-specific networks at the final stage.
Note that this strategy does not necessarily reduce the total training time but reduces the number of feature extractors for training the universal representations.
Here, we evaluate this strategy on Visual Decathlon under three random different groupings and report the performance of obtained universal representations in  \cref{tab:vdtg}.
Note that as ImageNet is a large diverse dataset, we treat it as a single group, and randomly assign the remaining datasets to three other groups.
In each grouping, we divide 10 domains/tasks into 4 groups\footnote{In `group 1', four tasks groups are \{ImNet\}, \{Airc., C100, DPed\}, \{DTD, GTSR, Flwr\}, \{OGlt, SVHN, UCF\}. They are \{ImNet\}, \{Airc., SVHN, UCF\}, \{DPed, GTSR, Flwr\}, \{C100, DTD, OGlt\} for `group 2' and \{ImNet\}, \{DPed., C100, SVHN\}, \{GTSR, OGlt, UCF\}, \{Airc, DTD, Flwr\} for `group 3'.}. 
The results show that training the universal network with task grouping obtains comparable, even slightly better results than learning it without task grouping. 
In addition, different groupings achieve similar performance in average to each other, while the first group obtains the best result.

\subsection{Further Analysis} \label{sec:exp:analysis}
In this section, we provide an extensive analysis over various adapter types, loss functions for knowledge distillation for multi-task learning and multi-domain learning. 

\paragraph*{Effect of adapters.}
As explained in \cref{sec:universal}, we employ adapters to align the universal representations with each task-specific representation (see $\alpha_{\theta^{t}}$ in \cref{eq:mtld}).
Here, we evaluate our method without any adapters (by directly matching universal representations with task-specific ones), also with two different adapter parameterizations including \emph{linear} adapters (\ie each adapter is constructed by a linear $1 \times 1$ convolutional layer, this is the default setting in \cref{sec:exp:dense}, \cref{sec:exp:vd}, \cref{sec:exp:fsl}), \emph{nonlinear} adapters (\ie each adapter consists of two linear convolutional layers and a ReLU layer between them). 
We report their results on NYU-v2 dataset in \cref{tab:nyuv2adapt}. 
From the results, we can see that, though directly align features without the adapters improves performance on all tasks over the vanilla MTL baseline (Uniform), it still performs significantly worse than using either linear or non-linear adapters. 
This verifies that the adapters helps aligning feature between the multi-task network with features of different single-task network. 
We also observe that, using linear and nonlinear adapters obtains comparable results and using linear adapters is sufficient.
We hypothesize that there is a tradeoff between the complexity of adapters and informativeness of aligned features.
For instance, using deep multi-layer adapters would overfit to the data and align the pairs very accurately, hence lead to inferior representation transfer.
Thus we argue that the linear adapters provide a good complexity/performance tradeoff.

\begin{table}[h!]
	\centering
    \resizebox{0.46\textwidth}{!}
    {
		\begin{tabular}{clccccccccccc}

		    \toprule
		    Arch & Type & Method & Seg. (IoU) $\uparrow$ & Depth (aErr) $\downarrow$ & Norm. (mErr) $\downarrow$ & $\bigtriangleup \text{MTL}$ $\uparrow$ \\
		    \midrule
		    \multirow{6}{*}{SegNet} & STL & SL & 40.54 & 0.6276 & 24.28 & +0.00 \\
		    \cmidrule{2-7}
		    & \multirow{4}{*}{MTL} & Uniform & 40.22 & 0.5196 & 29.09 & -1.13 \\
		    \cmidrule{3-7}
		    & & Ours (w/o adapter) & 41.64 & 0.5086 & 25.88 & +5.03 \\ 
		    & & Ours (nonlinear) & 44.84 & {\bf 0.4881} & 24.59 & +10.52 \\
		    & & Ours (linear) & {\bf 45.52} & 0.4912 & {\bf 24.57} & {\bf +10.95} \\
			\bottomrule
		\end{tabular}%
			}
		\caption{Testing results on NYU-v2. `linear' means we use a linear convolutional layer for adapters and `nonlinear' means we use non-linear adapters (\ie each adapter consists of two linear convolutional layers and a ReLU layer between them). `w/o adapter' means aligning features without any adapters}
		\label{tab:nyuv2adapt}
\end{table}%

\paragraph*{Loss functions for knowledge distillation.}
Here we evaluate various loss functions for distilling intermediate representations (\ie $\ell_{f}(\cdot)$ in \cref{eq:mtld}) including standard ones such as L2, cosine distance, and also Attention Transfer (AT)~\cite{komodakis2017paying} that align the spatial attention maps computed by averaging the feature maps along the channel dimension and CKA.
Here, we use linear adapters for aligning features between multi-task and single-task networks before measuring their discrepancy with these loss functions. 

\begin{table}[h!]
	\centering
    \resizebox{0.46\textwidth}{!}
    {
		\begin{tabular}{clccccccccccc}

		    \toprule
		    Arch & Type & Loss & Seg. (IoU) $\uparrow$ & Depth (aErr) $\downarrow$ & Norm. (mErr) $\downarrow$ & $\bigtriangleup \text{MTL}$ $\uparrow$ \\
		    \midrule
		    \multirow{7}{*}{SegNet} & STL & SL & 40.54 & 0.6276 & 24.28 & +0.00 \\
		    \cmidrule{2-7}
		    & \multirow{5}{*}{MTL} & -/- & 40.22 & 0.5196 & 29.09 & -1.13 \\
		    \cmidrule{3-7}
		    & & AT & 39.93 & 0.5060 & 28.81 & -0.26 \\
		    & & Cosine & 44.22 & 0.4969 & 25.45 & +8.37 \\
		    & & L2 & {\bf 45.52} & {\bf 0.4912} & {\bf 24.57} & {\bf +10.95} \\
			\bottomrule
		\end{tabular}%
			}
		\caption{Testing results on NYU-v2. `MTL' and `STL' means multi-task learning and single-task learning, respectively. `AT', `KL', `Cosine' and `L2' means using Attention Transfer, KL-divergence, Cosine and L2 loss functions respectively.}
		\label{tab:nyuv2loss}
\end{table}%

We first evaluate these loss functions for multiple dense prediction problem in NYU-v2 and report the results in \cref{tab:nyuv2loss} where our default loss function is L2. 
Here, we apply knowledge distillation with different loss function to the vanilla MTL method with SegNet~\cite{badrinarayanan2017segnet} as backbone (Note that CKA loss function is not included as it requires too large memory cost to operate on feature maps). 
From the results, we can see that, AT obtains the worst performance among these loss functions as it aligns the averaged features where some information is lost, but it still outperforms the vanilla MTL with uniform loss weights.
While Cosine loss function performs better than AT, using L2 loss function obtains the best results.

Compared to learning multiple dense prediction tasks in a single domain, learning universal representations from multiple visually-diverse domains is a more challenging problem.
Hence we use CKA as loss function for representation distillation, \ie $\ell_f$. 
Here we evaluate the effect of CKA in MetaDataset and compare it to different distillation loss functions, and report their performances in \cref{tab:fsllossf}.
In this study, we set $\lambda_p$ to zero and do not match prediction of the universal network with of the domain-specific ones.
Among these loss functions, the best results are obtained with CKA loss in all domains.
Although the universal representations are first mapped to the domain-specific spaces via adapters, L2 and cosine loss functions are not sufficient to match features from very diverse domains and further aligning features with CKA is significantly beneficial.

\begin{table}[t!]
	\centering
    \resizebox{0.425\textwidth}{!}
    {
		\begin{tabular}{cccccc}

		    \toprule
		    Test Dataset & L2 & COSINE & CKA \\
		    \midrule
		    ImageNet & $55.7\pm1.1$ & $57.0\pm1.1$& ${\bf 59.0\pm1.0}$ \\
		    Omniglot & $94.0\pm0.4$ & $94.1\pm0.4$& ${\bf 94.7\pm0.4}$ \\
		    Aircraft & $87.4\pm0.5$ & $88.3\pm0.5$& ${\bf 88.9\pm0.5}$ \\
		    Birds & $78.5\pm0.7$ & $77.5\pm0.8$& ${\bf 80.4\pm0.7}$ \\
		    Textures & $72.8\pm0.6$ & $73.2\pm0.7$& ${\bf 74.5\pm0.7}$ \\
		    Quick Draw & $81.2\pm0.6$ & $80.8\pm0.6$& ${\bf 81.9\pm0.6}$ \\
		    Fungi & $65.7\pm0.9$ & $65.9\pm0.9$& ${\bf 66.4\pm0.9}$ \\
		    VGG Flower & $87.5\pm0.6$ & $85.0\pm0.6$& ${\bf 91.3\pm0.5}$ \\
		    \midrule
		    Traffic Sign & $61.6\pm1.1$ & $59.5\pm1.1$& ${\bf 63.2\pm1.1}$ \\
		    MSCOCO & $53.4\pm1.0$ & $53.8\pm1.1$& ${\bf 56.6\pm1.0}$ \\
		    MNIST & ${\bf 94.7\pm0.3}$ & $93.2\pm0.5$& ${\bf 94.7\pm0.4}$ \\
		    CIFAR-10 & $71.1\pm0.8$ & $68.1\pm0.8$& ${\bf 73.8\pm0.7}$ \\
		    CIFAR-100 & $59.1\pm1.0$ & $58.1\pm1.0$& ${\bf 62.1\pm1.0}$ \\
			\bottomrule
		\end{tabular}%
			}
		\caption{\textbf{Quantitative analysis of knowledge distillation loss functions for $\ell_f$}. Mean accuracy, 95\% confidence interval are reported. COSINE denotes negative cosine similarity. All the loss functions are applied to measure the difference between intermediate representations of neural networks. All results are obtained with feature adaptation during meta-test stage.}
		\label{tab:fsllossf}
\end{table}%

We then evaluate individual contributions of distillation through representations and predictions while using CKA and KL-divergence respectively in \cref{tab:fsllossfp}.
Compared to only applying KL loss on predictions (`Ours w/o $\ell_f$'), only aligning representations with CKA loss function (`Ours w/o $\ell_p$') performs better in most domains. 
Finally, combining $\ell_f$ (CKA) with $\ell_p$ (KL divergence), \ie `Ours ($\ell_f+\ell_p$)', gives the best performance over the multi-domain models that are trained with the individual loss functions. 

\begin{table}[h!]
	\centering
    \resizebox{0.46\textwidth}{!}
    {
		\begin{tabular}{cccccc}

		    \toprule
		    Test Dataset & Ours w/o $\ell_p$ & Ours w/o $\ell_f$ & Ours ($\ell_f+\ell_p$) \\
		    \midrule
		    ImageNet & ${\bf 59.0\pm1.0}$ & $57.0\pm1.1$& $58.8\pm1.1$ \\
		    Omniglot & ${\bf 94.7\pm0.4}$ & $94.5\pm0.4$ & $94.5\pm0.4$\\
		    Aircraft & $88.9\pm0.5$ & $89.3\pm0.4$ & ${\bf 89.4\pm0.4}$\\
		    Birds & $80.4\pm0.7$ & $78.6\pm0.8$ & ${\bf 80.7\pm0.8}$\\
		    Textures & $74.5\pm0.7$ & $73.3\pm0.7$ & ${\bf 77.2\pm0.7}$\\
		    Quick Draw & $81.9\pm0.6$ & $81.6\pm0.6$ & ${\bf 82.5\pm0.6}$\\
		    Fungi & $66.4\pm0.9$ & $67.6\pm0.9$ & ${\bf 68.1\pm0.9}$\\
		    VGG Flower & $91.3\pm0.5$ & $89.6\pm0.5$ & ${\bf 92.0\pm0.5}$\\
		    \midrule
		    Traffic Sign & $63.2\pm1.1$ & $62.5\pm1.2$ & ${\bf 63.3\pm1.2}$\\
		    MSCOCO & $56.6\pm1.0$ & $55.6\pm1.1$ & ${\bf 57.3\pm1.0}$\\
		    MNIST & $94.7\pm0.4$ & ${\bf 95.3\pm0.4}$ & $94.7\pm0.4$\\
		    CIFAR-10 & $73.8\pm0.7$ & $72.9\pm0.8$ & ${\bf 74.2\pm0.8}$\\
		    CIFAR-100 & $62.1\pm1.0$ & $60.8\pm1.0$ & ${\bf 63.6\pm1.0}$\\
			\bottomrule
		\end{tabular}%
			}
		\caption{\textbf{Quantitative analysis of knowledge distillation loss functions on representations and predictions}. Mean accuracy, 95\% confidence interval are reported. `Ours w/o $\ell_p$' and `Ours w/o $\ell_f$' means we only apply CKA function on representations and apply KL divergence on predictions for knowledge distillation, respectively. `Ours ($\ell_f+\ell_p$)' is our model using both CKA on features and KL on predictions. All results are obtained with feature adaptation during meta-test stage.}
		\label{tab:fsllossfp}
\end{table}%

\begin{figure}[h!]
\begin{center}
\includegraphics[width=1.0\linewidth]{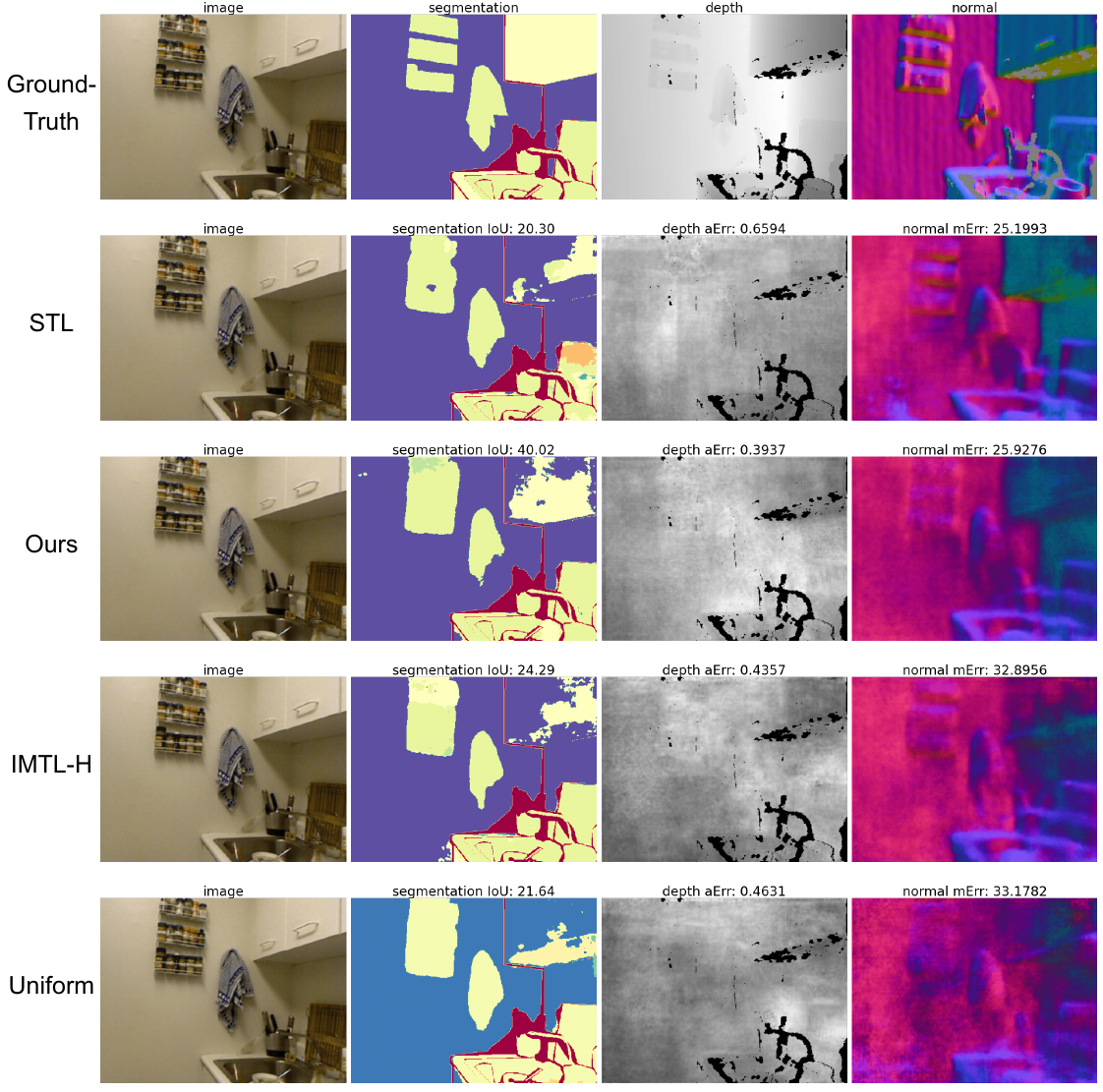}
\end{center}
\vspace{-0.3cm}
\caption{\textbf{Qualitative results on NYU-v2}. The fist column shows the RGB image, the second column plots the ground-truth or predictions with the IoU ($\uparrow$) score of all methods for semantic segmentation, the third column presents the ground-truth or predictions with the absolute error ($\downarrow$), and we show the prediction of surface normal with mean error ($\downarrow$) in the last column.}
\label{fig:predsnyuv2}
\end{figure}

\subsection{Qualitative results}
Here, we analyze our method and qualitatively compare our method to STL, MTL with Uniform loss weights and the best compared method, \ie IMTL-H~\cite{liu2021towards} for multi dense prediction problem on NYU-v2 with SegNet backbone (see \cref{fig:predsnyuv2}, see supplementary Sec. B.1 for more examples). 
We can see that, Uniform baseline obtains improvement on segmentation and depth estimation over STL, while it performs worse in surface normal estimation.
Though dynamically balancing the loss values with IMTL-H improves the overall performance, it still performs worse in surface normal estimation. Finally by distilling representations from single-task learning model to the universal network, our method can produce better or comparable results to STL, \ie our method produces similar outputs for surface normal as STL and more accurate predictions for segmentation and depth estimation as our method enables a balanced optimization of universal network and a task can be benefited from another one. 
This indicates the effectiveness of our method on learning shared representations for multiple dense predictions.

\begin{figure}
\begin{center}
\includegraphics[width=1.0\linewidth]{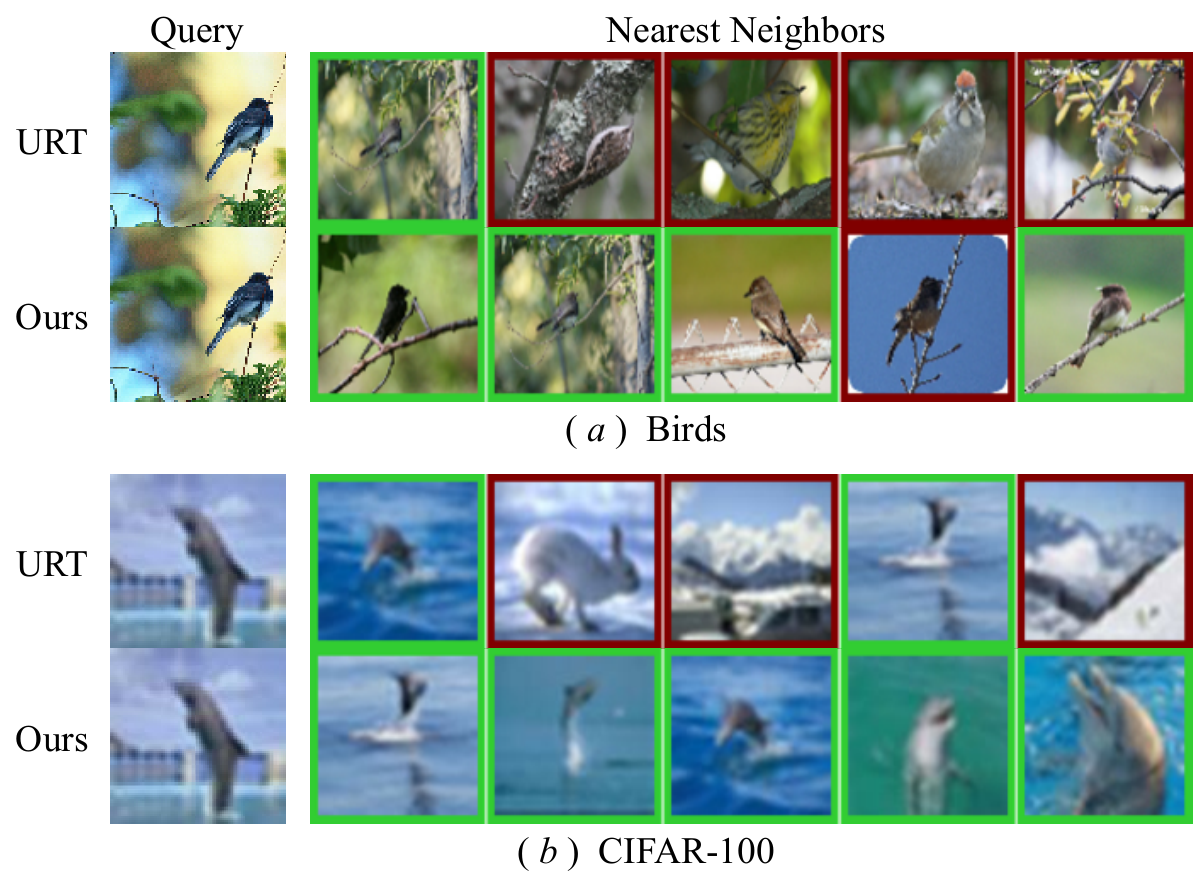}
\end{center}
\vspace{-0.3cm}
\caption{\textbf{Qualitative analysis of our method in two datasets}. Green and red colors indicate correct and false predictions respectively.}
\label{fig:visual}
\end{figure}

We also qualitatively analyze our method and compare it to the state-of-the-art URT~\cite{liu2020universal} in \cref{fig:visual} for cross-domain few-shot learning in MetaDataset.
In particular, we illustrate the nearest neighbors in two different datasets given a query image (see supplementary Sec. B.2.6 for more examples).
While URT retrieves images with more similar colors, shapes and backgrounds, our method is able to retrieve semantically similar images and finds more correct neighbors than URT. 
It again suggests that our method is able to learn more semantically meaningful and general representations.

\section{Conclusion}\label{sec:conclusion}
We showed that learning general features from multiple tasks and domains is an important step for better generalization in various computer vision problems including multiple dense prediction, multi-domain image classification and cross-domain few-shot learning problems.
By distilling representations from multiple task-specific or domain-specific networks, we can successfully learn a single set of universal representations after aligning them via small task/domain-specific adapters.
These representations are compact and generalize better to unseen samples, tasks and domains in multiple benchmarks.

So far, we focuesd on multi-task or multi-domain learning. We would like to extend the proposed URL method to problems that involve multiple tasks and multiple domains at the same time such as semantic segmentation and depth estimation from different cities, time of day (\eg day or night time), weather (\eg clear, rainy, snowy), or domains. 

\clearpage
\appendix

\section{Implementation Details}

\subsection{Multi-task Dense Prediction}\label{suppsec:mdp}
We evaluate our method on learning universal representations for performing multiple dense prediction tasks on two standard multi-task learning benchmarks  NYU-v2~\cite{silberman2012indoor} and Cityscapes~\cite{cordts2016cityscapes} as in~\cite{liu2019end,liu2021conflict}. Here, we provide more details about our implementation.

We follow the training and evaluation settings in~\cite{liu2019end,liu2021conflict} for both single-task and multi-task learning in both datasets. 
More specifically, \emph{NYU-V2}~\cite{silberman2012indoor} contains RGB-D indoor scene images, where we evaluate performances on 3 tasks, including 13-class semantic segmentation, depth estimation, and surface normals estimation. 
We use the true depth data recorded by the Microsoft Kinect and surface normals provided in \cite{eigen2015predicting} for depth estimation and surface normal estimation as in~\cite{liu2019end}. All images are resized to $288 \times 384$ resolution as in \cite{liu2019end}. We follow the default setting in~\cite{silberman2012indoor,liu2019end} where 795 and 654 images are used for training and testing, respectively.
\emph{Cityscapes}~\cite{cordts2016cityscapes} consists of street-view images, which are labeled for two tasks: 7-class semantic segmentation\footnote{The original version of Cityscapes provides labels 7\&19-class semantic segmentation. We follow the 7-class semantic segmentation evaluation protocol as in \cite{liu2019end} to be able to compare to the related works.} and depth estimation. We resize the images to $128 \times 256$ to speed up the training as~\cite{liu2019end}. 

In both NYU-v2 and Cityscapes, we follow the training and evaluation protocol in~\cite{liu2019end}.
We consider two backbone cases. 
For encoder-based one, we use the SegNet~\cite{badrinarayanan2017segnet} as the backbone. 
As in ~\cite{liu2019end}, we use cross-entropy loss for semantic segmentation, l1-norm loss for depth estimation in Cityscapes, and cosine similarity loss for surface normal estimation in NYU-v2. 
We use the exactly same hyper-parameters including learning rate, optimizer and also the same evaluation metrics, mean intersection over union (mIoU), absolute error (aErr) and mean error (mErr) in the predicted angles to evaluate the semantic segmentation, depth estimation and surface normals estimation task, respectively in~\cite{liu2019end}. More specifically, we use Adam as the optimizer and train the model for 200 epochs as in~\cite{liu2019end} with the learning rate of 0.0001 which is halved at the 100-th epoch. The batch size is set to 2 and 8 for NYU-v2 and Cityscapes, respectively. We use the same augmentation as in~\cite{liu2019end}, such as random crop, flipping.

For the decoder-based methods, as the MTI-Net~\cite{vandenhende2020mti} requires multi-scale feature extractor (encoder), we follow \cite{vandenhende2020mti,vandenhende2021multi}, use the HRNet-18 backbone~\cite{sun2019deep} initialized with ImageNet pretrained weight as the feature encoder. As in~\cite{vandenhende2020mti,vandenhende2021multi}, the batch size is set to 8 for both NYU-v2 and Cityscapes.
We use the same loss functions, evaluation metrics, and training and evaluation protocol as the encoder-based methods. 

In our method, we use the uniform loss weights (\ie $\lambda^t=1$ for all tasks) for task-specific losses, unless stated otherwise. As we do \emph{not} minimize the difference between predictions of the universal and single-task networks, we set $\lambda^t_p$ in Eq. (3) to zero.
We then first split the train set as train and validation set to search $\lambda_f^t \in \{1, 2\}$ by cross-validation and train our network on the whole training set. We set $\lambda_f^t$ to 1 for semantic segmentation and depth and 2 for surface normal estimation. For the optimization of adapters, we use Adam as optimizer with the learning rate of 0.01 and weight decay of 0.0001 and anneal the learning rate to 0 using cosine scheduler.

\subsection{Multi-domain Learning}\label{suppsec:mdmsl}

\paragraph*{Dataset.} The \emph{Visual Decathlon Benchmark}~\cite{rebuffi2017learning} consists of 10 different well-known datasets:  including ILSVRC\_2012~(ImNet)~\cite{russakovsky2015imagenet}, FGVC-Aircraft~(Airc.)~\cite{maji2013fine}, CIFAR-100~(C100)~\cite{krizhevsky2009learning}, Daimler Mono Pedestrian Classification Benchmark~(DPed)~\cite{munder2006experimental}, Describable Texture Dataset~(DTD)~\cite{cimpoi2014describing}, German Traffic Sign Recognition (GTSR)~\cite{houben2013detection}, Flowers102~(Flwr)~\cite{nilsback2008automated}, Omniglot (OGlt)~\cite{Lake1332}, Street View House Numbers~(SVHN)~\cite{netzer2011reading}, UCF101 (UCF)~\cite{soomro2012dataset}. In this benchmark~\cite{rebuffi2017learning}, images are resized to a common resolution of roughly 72 pixels to accelerate evaluation.

\paragraph*{Implementation details.} Each dataset has train/val/test splits and we follow the standard training and evaluation protocol in~\cite{rebuffi2017learning,rebuffi2018efficient}. For ImageNet, we random crop and center crop a $72 \times 72$ patch for training and evaluation respectively. For other datasets, as the aspect ratio in Airc. and DPed is quite different from other datasets, we first resize the images in both datasets to $72 \times 72$, and we random crop and center crop a $64 \times 64$ patch for training and evaluation as in~\cite{rebuffi2017learning}.

We follow~\cite{rebuffi2017learning,rebuffi2018efficient}, use the ResNet-26~\cite{he2016deep} as the backbone for domain-specific network and universal network. 
In our universal network, the backbone (\ie ResNet-26) is shared across all domains and followed by domain-specific linear classifiers. We use the same data augmentation as in~\cite{rebuffi2017learning,rebuffi2018efficient}, such as random crop, flipping. We use SGD as optimizer with weight decay and train domain-specific networks and our universal network for 120 epochs as in~\cite{rebuffi2017learning,rebuffi2018efficient}. 

For domain-specific models, \ie, learning one model per domain, we first train a model on ImageNet with learning rate of 0.1 and weight decay of 0.0005 and we finetune it for other datasets with weight decay of 0.0005. For finetuning on each dataset (except ImageNet), we set learning rate to 0.1 for Airc., Flwr, OGlt, SVHN, UCF and 0.01 for C100, DPed, DTD, GTSR. 

Here we set the loss weights to 1 (\ie $\lambda^t=1$ for all tasks) and perform cross-validation to search loss weights ($\lambda^t_f,\lambda^t_p$) in $\{0.1, 1, 10\}$ for knowledge distillations on features and predictions, and set $\lambda_f$ and $\lambda_p$ to 10 for ImageNet, 0.1 for DPed, and 1 for other datasets. We optimize our universal network and vanilla MDL using SGD as optimizer as in~\cite{rebuffi2017learning,rebuffi2018efficient} with the learning rate of 0.01 and weight decay of 0.0001 for 120 epochs. The learning rate is scaled by 0.1 at 80-th and 100-th epoch as in~\cite{rebuffi2017learning,rebuffi2018efficient}. We optimize the parameters of adapters by using the Adam as optimizer with the learning rate of 0.01 which is annealed to zero using cosine scheduler and weight decay of 0.0005. We evaluate our method though the official online evaluation provided by~\cite{rebuffi2017learning}.

\subsection{Cross-domain Few-shot Learning}\label{suppsec:mdfsl}
In all experiments we build our method on ResNet-18~\cite{he2016deep} backbone for both single-domain and multi-domain networks.

\paragraph*{Training details of single-domain models}\label{appsec:sdl}
We train one ResNet-18 model for each training dataset. For optimization, we follow the training protocol in \cite{dvornik2020selecting}. Specifically, we use SGD optimizer and cosine annealing for all experiments with a momentum of 0.9 and a weight decay of $7\times 10^{-4}$. The learning rate, batch size, annealing frequency, maximum number of iterations are shown in \cref{apptab:hyperparams}. To regularize training, we also use the exact same data augmentations as in \cite{dvornik2020selecting}, \eg random crops and random color augmentations.

\begin{table}[h!]
	\centering
    \resizebox{0.46\textwidth}{!}
    {
		\begin{tabular}{ccccc}
		    \toprule
		    Dataset & learning rate & batch size & annealing freq. & max. iter. \\
		    \midrule
		    ImageNet & $3\times 10^{-2}$ & 64 & 48,000 & 480,000\\
		    Omniglot & $3\times 10^{-2}$ & 16 & 3000 & 50,000\\
		    Aircraft & $3\times 10^{-2}$ & 8 & 3000 & 50,000\\
		    Birds & $3\times 10^{-2}$ & 16 & 3000 & 50,000\\
		    Textures & $3\times 10^{-2}$ & 32 & 1500 & 50,000\\
		    Quick Draw & $1\times 10^{-2}$ & 64 & 48,000 & 480,000\\
		    Fungi & $3\times 10^{-2}$ & 32 & 15,000 & 480,000\\
		    VGG Flower & $3\times 10^{-2}$ & 8 & 1500 & 50,000\\
			\bottomrule
		\end{tabular}%
			}
		\caption{Training hyper-parameters of single domain learning.}
		\label{apptab:hyperparams}
\end{table}%

\paragraph*{Training details of our method}

In the multi-domain network, we share all the layers but the last classifier across the domains.
To train the multi-domain network, we use the same optimizer with a weight decay of $7\times 10^{-4}$ and a scheduler as single domain learning model for learning 240,000 iterations. The learning rate is 0.03 and the annealing frequency is 48,000. Similar to~\cite{triantafillou2019meta} that the training episodes have 50\% probability coming from the ImageNet data source, each training batch for our multi-domain network consists of 50\% data coming from ImageNet. In other words. The batch size for ImageNet is $64\times 7$ and is $64$ for the other 7 datasets.

We first set loss weight $\lambda^t$ of domain-specific losses as 1 for all domains. We set $\lambda_f$ and $\lambda_p$ as 4 for ImageNet and 1 for other datasets, respectively. And we linearly anneal $\lambda$ by $\lambda \leftarrow \lambda \times (1 - \frac{t}{K})$, where, $t$ is the current iteration and $K$ is the total number of iterations to anneal $\lambda$ to zero. Here, $K=k\times (anneal.~freq.)$, where $anneal.~freq$. is 48, 000 in this work. We search the $k=\{1, 2, 3, 4, 5\}$ based on cross-validation over the validation sets of 8 training datasets and $k$ is 5 (\ie $K=240,000$) for ImageNet, is 2 for Omniglot, Quick Draw, Fungi and is 1 for other datasets. For all experiments, early-stopping is performed based on cross-validation over the validations sets of 8 training datasets.

For the optimization of feature adaptation during meta-test stage, we initialize $\vartheta$ as an indentity matrix, which allows the NCC to use the original features produced by our universal network and optimize $\vartheta$ from a good start point. Similar to the optimization in~\cite{dvornik2020selecting}, we optimize $\vartheta$ for 40 iterations using Adadelta~\cite{zeiler2012adadelta} as optimizer with a learning rate of 0.1 for first eight datasets and 1 for the last five datasets.

\section{More results}
\subsection{Multi-task Dense Prediction}
\begin{figure}[h!]
\begin{center}
\includegraphics[width=0.98\linewidth]{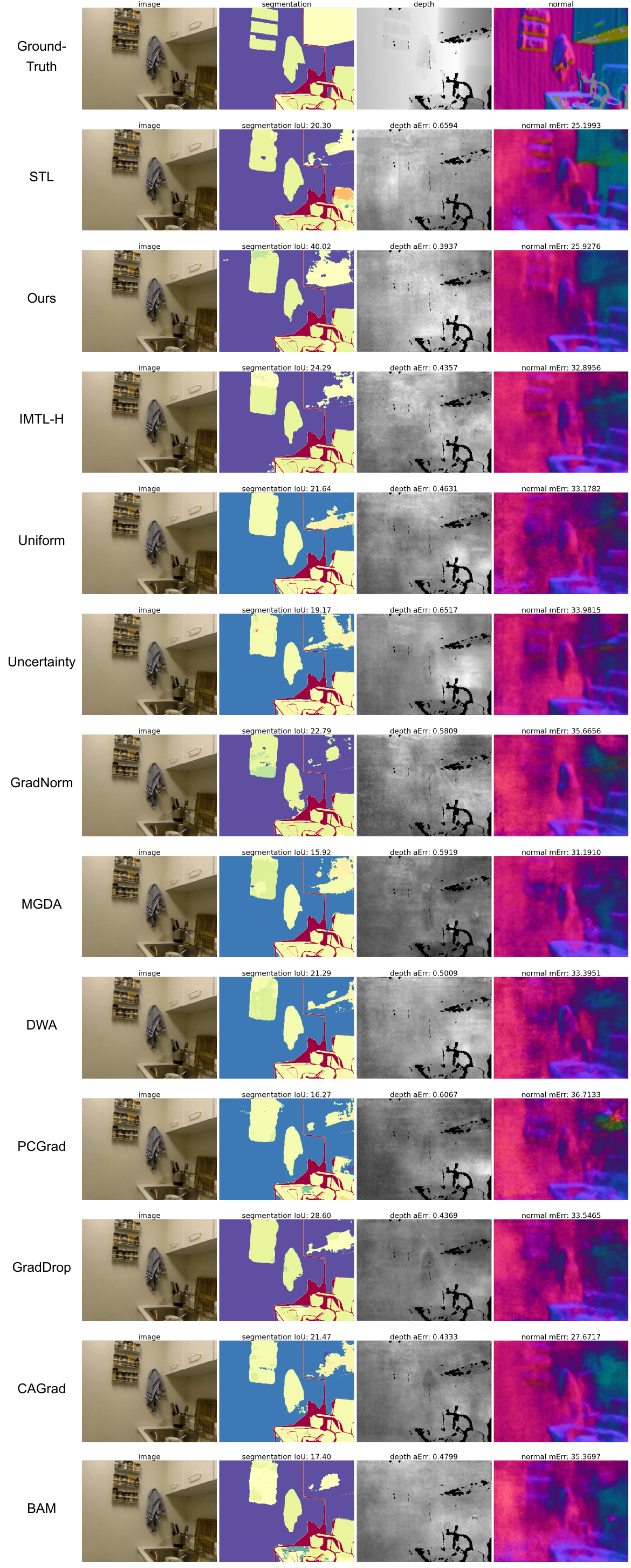}
\end{center}
\vspace{-0.3cm}
\caption{\textbf{Qualitative results on NYU-v2}. The fist column shows the RGB image, the second column plots the ground-truth or predictions with the IoU ($\uparrow$) score of all methods for semantic segmentation, the third column presents the ground-truth or predictions with the absolute error ($\downarrow$), and we show the prediction of surface normal with mean error ($\downarrow$) in the last column.}
\label{suppfig:predsnyuv2}
\end{figure}

Here, we analyze our method and qualitatively compare our method to STL, MTL with Uniform loss weights and the best compared method, \ie IMTL-H~\cite{liu2021towards} and other related methods for multi dense prediction problem on NYU-v2 with SegNet backbone (see \cref{suppfig:predsnyuv2}). 
We can see that, Uniform baseline obtains improvement on segmentation and depth estimation over STL, while it performs worse in surface normal estimation.
Though dynamically balancing the loss values with IMTL-H improves the overall performance, it still performs worse in surface normal estimation. Finally by distilling representations from single-task learning model to the universal network, our method can produce better or comparable results to STL, \ie our method produces similar outputs for surface normal as STL and more accurate predictions for segmentation and depth estimation as our method enables a balanced optimization of universal network and a task can be benefited from another one. 
This indicates the effectiveness of our method on learning shared representations for multiple dense predictions.

\subsection{Cross-domain Few-shot learning}\label{suppsec:mdfslresults}

In this section, we first evaluate each single-domain model for few-shot classification on each test dataset. We then show complete results on varying-way five-shot and five-way one-shot settings. We also evaluate the effect of the adaptors for aligning features in knowledge distillation. As the code of MetaDataset has been updated, we report results using the updated evaluation protocol from MetaDataset and compare our method with Cross-Transformer~\cite{doersch2020crosstransformers} and Transductive CNAPS~\cite{jiang2020transductive} methods. Finally more qualitative results and global retrieval results are reported. 

\subsubsection{Complete results of single domain learning}

\begin{table*}[t]
	\centering
    \resizebox{0.8\textwidth}{!}
    {
		\begin{tabular}{c|cccccccc}

		    \toprule
		    \diagbox{Test Dataset}{Train Dataset} & ImageNet & Omniglot & Aircraft & Birds & Textures & Quick Draw & Fungi & Vgg Flower\\
		    \midrule
		    ImageNet & ${\bf 55.8\pm1.0}$& $17.1\pm0.6$& $21.7\pm0.7$& $25.4\pm0.8$& $24.2\pm0.8$& $24.1\pm0.8$& $32.9\pm0.9$& $25.0\pm0.8$ \\
		    Omniglot & $67.4\pm1.2$& ${\bf 93.2\pm0.5}$& $58.2\pm1.2$& $58.7\pm1.4$& $57.3\pm1.4$& $78.4\pm1.0$& $57.6\pm1.3$& $54.6\pm1.3$ \\
		    Aircraft & $49.5\pm0.9$& $16.8\pm0.5$& ${\bf 85.7\pm0.5}$& $31.4\pm0.8$& $26.0\pm0.7$& $23.8\pm0.6$& $31.0\pm0.7$& $24.6\pm0.6$ \\
		    Birds & ${\bf 71.2\pm0.9}$& $13.0\pm0.6$& $19.9\pm0.7$& $65.0\pm0.9$& $19.6\pm0.7$& $16.7\pm0.7$& $42.8\pm1.0$& $28.9\pm0.8$ \\
		    Textures & ${\bf 73.0\pm0.6}$& $25.0\pm0.5$& $38.6\pm0.7$& $42.2\pm0.7$& $54.9\pm0.7$& $38.6\pm0.6$& $54.1\pm0.7$& $42.3\pm0.7$ \\
		    Quick Draw & $53.9\pm1.0$& $51.0\pm1.0$& $38.8\pm1.0$& $38.2\pm1.0$& $36.8\pm0.9$& ${\bf 82.8\pm0.6}$& $37.7\pm0.9$& $39.7\pm1.0$ \\
		    Fungi & $41.6\pm1.0$&  $9.1\pm0.5$& $14.9\pm0.7$& $25.5\pm0.8$& $15.6\pm0.7$& $12.5\pm0.6$& ${\bf 65.8\pm0.9}$& $23.3\pm0.8$ \\
		    VGG Flower & ${\bf 87.0\pm0.6}$& $23.8\pm0.6$& $45.5\pm0.8$& $62.9\pm0.8$& $44.4\pm0.8$& $33.4\pm0.7$& $79.6\pm0.7$& $78.3\pm0.7$ \\
		    \midrule
		    Traffic Sign & ${\bf 47.4\pm1.1}$& $15.1\pm0.7$& $30.8\pm0.9$& $31.0\pm0.9$& $38.8\pm1.1$& $31.1\pm0.9$& $28.0\pm0.9$& $30.4\pm0.9$ \\
		    MSCOCO & ${\bf 53.5\pm1.0}$& $12.9\pm0.6$& $22.5\pm0.8$& $25.1\pm0.9$& $23.7\pm0.8$& $21.3\pm0.8$& $32.5\pm1.0$& $25.7\pm0.8$ \\
		    MNIST & $78.1\pm0.7$& ${\bf 89.8\pm0.5}$& $68.0\pm0.8$& $73.0\pm0.7$& $64.5\pm0.8$& $88.2\pm0.5$& $62.2\pm0.8$& $72.1\pm0.7$ \\
		    CIFAR-10 & ${\bf 67.3\pm0.8}$& $28.5\pm0.6$& $41.2\pm0.7$& $41.8\pm0.8$& $36.9\pm0.7$& $40.0\pm0.7$& $38.8\pm0.7$& $41.3\pm0.8$ \\
		    CIFAR-100 & ${\bf 56.6\pm0.9}$& $12.3\pm0.6$& $24.3\pm0.9$& $28.8\pm0.9$& $24.2\pm0.9$& $23.4\pm0.8$& $25.2\pm0.9$& $29.1\pm1.0$ \\
			\bottomrule
		\end{tabular}%
			}
		\caption{Results of all single domain learning models. Mean accuracy and 95\% confidence interval are reported. The first eight datasets are seen during training and the last five datasets are unseen for test only.}
		\label{tabapp:stl}
\end{table*}%

\begin{table*}[ht]
\vspace{-0.25cm}
	\centering
    \resizebox{0.85\textwidth}{!}
    {
		\begin{tabular}{ccccc|cccc}
			& \multicolumn{4}{c}{Five-Shot} & \multicolumn{4}{c}{Five-Way One-Shot} \\
		    \toprule
		    \multirow{2}{*}{Test Dataset} & Simple CNAPS & SUR & URT & \multirow{2}{*}{Ours} & Simple CNAPS & SUR & URT & \multirow{2}{*}{Ours}\\
		    &  \cite{bateni2020improved}  & \cite{dvornik2020selecting} & \cite{liu2020universal} & & \cite{bateni2020improved} & \cite{dvornik2020selecting}& \cite{liu2020universal} & \\
		    \midrule
		    ImageNet & $47.2\pm1.0$ & $46.7\pm1.0$ & $48.6\pm1.0$ & ${\bf 49.4\pm1.0}$ & $42.6\pm0.9$ & $40.7\pm1.0$ & $47.4\pm1.0$ & ${\bf 49.6\pm1.1}$ \\
			Omniglot & $95.1\pm0.3$ & $95.8\pm0.3$ & ${\bf 96.0\pm0.3}$ & ${\bf 96.0\pm0.3}$ & $93.1\pm0.5$ & $93.0\pm0.7$ & $95.6\pm0.5$ & ${\bf 95.8\pm0.5}$ \\
			Aircraft & $74.6\pm0.6$ & $82.0\pm0.6$ & $81.2\pm0.6$ & ${\bf 84.8\pm0.5}$ & $65.8\pm0.9$ & $67.1\pm1.4$ & $77.9\pm0.9$ & ${\bf 79.6\pm0.9}$ \\
			Birds & $69.6\pm0.7$ & $62.8\pm0.9$ & $71.2\pm0.7$ & ${\bf 76.0\pm0.6}$ & $67.9\pm0.9$ & $59.2\pm1.0$ & $70.9\pm0.9$ & ${\bf 74.9\pm0.9}$ \\
			Textures & $57.5\pm0.7$ & $60.2\pm0.7$ & $65.2\pm0.7$ & ${\bf 69.1\pm0.6}$ & $42.2\pm0.8$ & $42.5\pm0.8$ & $49.4\pm0.9$ & ${\bf 53.6\pm0.9}$ \\
			Quick Draw & $70.9\pm0.6$ & $79.0\pm0.5$ & ${\bf 79.2\pm0.5}$ & $78.2\pm0.5$ & $70.5\pm0.9$ & ${\bf 79.8\pm0.9}$ & $79.6\pm0.9$ & $79.0\pm0.8$ \\
			Fungi & $50.3\pm1.0$ & $66.5\pm0.8$ & $66.8\pm0.9$ & ${\bf 70.0\pm0.8}$ & $58.3\pm1.1$ & $64.8\pm1.1$ & $71.0\pm1.0$ & ${\bf 75.2\pm1.0}$ \\
			VGG Flower & $86.5\pm0.4$ & $76.9\pm0.6$ & $82.4\pm0.5$ & ${\bf 89.3\pm0.4}$ & ${\bf 79.9\pm0.7}$ & $65.0\pm1.0$ & $72.7\pm0.9$ & ${\bf 79.9\pm0.8}$ \\
			Traffic Sign & $55.2\pm0.8$ & $44.9\pm0.9$ & $45.1\pm0.9$ & ${\bf 57.5\pm0.8}$ & $55.3\pm0.9$ & $44.6\pm0.9$ & $52.6\pm0.9$ & ${\bf 57.9\pm0.9}$ \\
			MSCOCO & $49.2\pm0.8$ & $48.1\pm0.8$ & $52.3\pm0.8$ & ${\bf 56.1\pm0.8}$ & $48.8\pm0.9$ & $47.8\pm1.1$ & $56.9\pm1.1$ & ${\bf 59.1\pm1.0}$ \\
			MNIST & $88.9\pm0.4$ & ${\bf 90.1\pm0.4}$ & $86.5\pm0.5$ & $89.7\pm0.4$ & ${\bf 80.1\pm0.9}$ & $77.0\pm0.9$ & $75.6\pm0.9$ & $78.7\pm0.9$ \\
			CIFAR-10 & ${\bf 66.1\pm0.7}$ & $50.3\pm1.0$ & $61.4\pm0.7$ & $66.0\pm0.7$ & $50.3\pm0.9$ & $35.8\pm0.8$ & $47.3\pm0.9$ & ${\bf 54.7\pm0.9}$ \\
			CIFAR-100 & $53.8\pm0.9$ & $46.4\pm0.9$ & $52.5\pm0.9$ & ${\bf 57.0\pm0.9}$ & $53.8\pm0.9$ & $42.9\pm1.0$ & $54.9\pm1.1$ & ${\bf 61.8\pm0.9}$ \\
		    \midrule
		    Average Rank & $3.1$ & $3.0$ & $2.5$ & $1.3$ & $2.8$ & $3.5$ & $2.4$ & $1.2$ \\
			\bottomrule
		\end{tabular}%
			}
		\caption{Results of Five-Way One-Shot and Varying-Way Five-Shot settings. Mean accuracies are reported and the results with confidence interval are reported.}
		\label{tabapp:fixedshot}
\end{table*}%

To study the universal representation learning from multiple datasets, we train one network on each training dataset and use each single-domain network as the feature extractor and test it for few-shot classification in each dataset. This involves evaluating 8 single-domain networks on 13 datasets using Nearest Centroid Classifier (NCC). \Cref{tabapp:stl} shows the results of single domain learning models, where each column present the mean accuracy and 95\% confidence interval of a single-domain network trained on one dataset (\eg ImageNet) and evaluated on 13 test datasets. The average accuracy and 95\% confidence intervals computed over 600 few-shot tasks. The numbers in bold indicate that a method has the best accuracy per dataset.

As shown in \cref{tabapp:stl}, the feature of the ImageNet model generalizes well and achieves the best results on four out of eight seen datasets, \eg ImageNet, Birds, Texture, VGG Flower and four out of five previously unseen datasets, \eg Traffic Sign, MSCOCO, CIFAR-10, CIFAR-100. The models trained on Omniglot, Aircraft, Quick Draw, and Fungi perform the best on the corresponding datasets while the Omniglot model also generalizes well to MNIST which has the similar style images to Omniglot. We then pick the best performing model, forming the best single-domain model (Best SDL) which serves a very competitive baseline for universal representation learning.

\begin{table}[ht]
	\centering
    \resizebox{0.42\textwidth}{!}
    {
		\begin{tabular}{cccc}

		    \toprule
		    Test Dataset & Ours (CKA w/o $A_{\theta}$) & Ours (CKA) \\
		    \midrule
		    ImageNet & $58.3\pm1.0$ & ${\bf 59.0\pm1.0}$ \\
		    Omniglot & $94.4\pm0.4$ & ${\bf 94.7\pm0.4}$ \\
		    Aircraft & ${\bf 88.9\pm0.5}$ & ${\bf 88.9\pm0.4}$ \\
		    Birds & $78.7\pm0.8$ & ${\bf 80.4\pm0.7}$ & \\
		    Textures & ${\bf 74.8\pm0.7}$ & $74.5\pm0.7$ \\
		    Quick Draw  & ${\bf 82.1\pm0.6}$ & $81.9\pm0.6$ \\
		    Fungi & $65.4\pm0.9$ & ${\bf 66.4\pm0.9}$ \\
		    VGG Flower & $87.5\pm0.6$ & ${\bf 91.3\pm0.5}$ \\
		    \midrule
		    Traffic Sign & ${\bf 63.3\pm1.1}$ & $63.2\pm1.1$ \\
		    MSCOCO & $55.3\pm1.0$ & ${\bf 56.6\pm1.0}$ \\
		    MNIST & ${\bf 94.9\pm0.4}$ & $94.7\pm0.4$ \\
		    CIFAR-10 & $73.4\pm0.7$ & ${\bf 73.8\pm0.7}$ \\
		    CIFAR-100 & $61.8\pm1.0$ & ${\bf 62.1\pm1.0}$ \\
			\bottomrule
		\end{tabular}%
			}
		\caption{Results of our method using CKA, CKA without adaptors (\ie $A_{\theta}$). Mean accuracy and 95\% confidence interval are reported. Here, Ours (CKA w/o $A_{\theta}$) indicates that adaptors are not applied for aligning features. All results are obtained with feature adaptation during meta-test stage.}
		\label{apptab:lossf}
\end{table}%

\begin{table*}[ht]
	\centering
    \resizebox{1\textwidth}{!}
    {
		\begin{tabular}{c|cccc|cccc|cccc|cccc|cccc|cccc|cccc|cccc}

		    \toprule
		    Test Dataset & \multicolumn{4}{c|}{ImageNet} & \multicolumn{4}{c|}{Omniglot} & \multicolumn{4}{c|}{Aircraft} & \multicolumn{4}{c|}{Birds} & \multicolumn{4}{c|}{Textures} & \multicolumn{4}{c|}{Quick Draw} & \multicolumn{4}{c|}{Fungi} & \multicolumn{4}{c}{VGG Flower} \\
		    \midrule
		    Recall@$k$ & 1 & 2 & 4 & 8 & 1 & 2 & 4 & 8 & 1 & 2 & 4 & 8 & 1 & 2 & 4 & 8 & 1 & 2 & 4 & 8 & 1 & 2 & 4 & 8 & 1 & 2 & 4 & 8 & 1 & 2 & 4 & 8 \\
		    \midrule
		    Sum & $22.1$ & $30.3$ & $39.6$ & $50.0$ & $84.7$ & $91.8$ & $95.8$ & $97.8$ & $69.7$ & $80.7$ & $88.6$ & $94.5$ & $45.9$ & $59.7$ & $72.0$ & $84.1$ & $66.3$ & $78.2$ & $87.3$ & $94.0$ & $77.4$ & $84.3$ & $89.1$ & $92.1$ & $31.9$ & $42.9$ & $54.0$ & $65.4$ & $85.1$ & $92.1$ & $96.7$ & $98.6$ \\
		    Concate & $20.2$ & $28.0$ & $36.9$ & $47.8$ & $84.4$ & $91.5$ & $95.8$ & $97.8$ & $44.3$ & $58.1$ & $71.1$ & $82.9$ & $35.5$ & $48.8$ & $62.8$ & $76.0$ & $68.8$ & $78.2$ & $87.3$ & $93.9$ & $73.0$ & $80.8$ & $86.2$ & $90.6$ & $30.7$ & $40.4$ & $51.8$ & $63.0$ & $83.4$ & $91.3$ & $95.2$ & $98.2$ \\
		    MDL & $29.8$ & $39.6$ & $49.9$ & $60.9$ & ${\bf 89.8}$ & ${\bf 94.3}$ & $96.8$ & $98.2$ & $80.3$ & $87.1$ & $92.5$ & $95.9$ & $63.2$ & $75.9$ & $84.7$ & $91.6$ & $67.0$ & $77.1$ & $85.4$ & $92.9$ & $79.5$ & $85.4$ & $89.7$ & $92.8$ & $40.2$ & $51.7$ & $63.0$ & $72.4$ & $86.9$ & $93.3$ & $96.6$ & $98.4$ \\
		    Simple CNAPS~\cite{bateni2020improved} & $34.0$ & $43.8$ & $54.4$ & $65.1$ & $84.9$ & $91.6$ & $95.5$ & $97.5$ & $70.5$ & $82.5$ & $91.3$ & $96.1$ & $55.9$ & $70.5$ & $82.0$ & $90.2$ & $64.8$ & $76.9$ & ${\bf 87.6}$ & ${\bf 94.4}$ & $75.3$ & $83.0$ & $88.0$ & $91.7$ & $29.1$ & $39.0$ & $49.6$ & $61.5$ & $88.1$ & $94.1$ & ${\bf 97.6}$ & ${\bf 99.2}$ \\
		    Ours & ${\bf 36.1}$ & ${\bf 46.2}$ & ${\bf 56.3}$ & ${\bf 66.6}$ & $89.7$ & ${\bf 94.3}$ & ${\bf 97.2}$ & ${\bf 98.3}$ & ${\bf 83.3}$ & ${\bf 90.4}$ & ${\bf 93.7}$ & ${\bf 96.3}$ & ${\bf 66.7}$ & ${\bf 78.9}$ & ${\bf 87.9}$ & ${\bf 94.1}$ & ${\bf 70.2}$ & ${\bf 80.8}$ & $87.5$ & $93.8$ & ${\bf 79.9}$ & ${\bf 86.5}$ & ${\bf 90.5}$ & ${\bf 93.2}$ & ${\bf 44.5}$ & ${\bf 56.2}$ & ${\bf 67.3}$ & ${\bf 76.4}$ & ${\bf 90.0}$ & ${\bf 94.6}$ & $97.5$ & $98.9$ \\
			\bottomrule
		\end{tabular}%
			}
		\caption{Global retrieval performance on MetaDataset (seen datasets). In addition to few-shot learning experiments, we evaluate our method in a non-episodic retrieval task to further compare the generalization ability of our universal representations. }
		\label{tab:recallseen}
\end{table*}

\begin{table*}[ht]
	\centering
    \resizebox{0.8\textwidth}{!}
    {
		\begin{tabular}{c|cccc|cccc|cccc|cccc|cccc}

		    \toprule
		    Test Dataset & \multicolumn{4}{c|}{Traffic Sign} & \multicolumn{4}{c|}{MSCOCO} & \multicolumn{4}{c|}{MNIST} & \multicolumn{4}{c|}{CIFAR-10} & \multicolumn{4}{c}{CIFAR-100} \\
		    \midrule
		    Recall@$k$ & 1 & 2 & 4 & 8 & 1 & 2 & 4 & 8 & 1 & 2 & 4 & 8 & 1 & 2 & 4 & 8 & 1 & 2 & 4 & 8 \\
		    \midrule
		    Sum & $94.6$ & $97.2$ & $98.5$ & ${\bf 99.3}$ & $62.6$ & $71.2$ & $78.9$ & $85.0$ & $98.3$ & $99.2$ & ${\bf 99.6}$ & ${\bf 99.8}$ & $54.0$ & $68.9$ & $81.9$ & $90.6$ & $27.8$ & $37.4$ & $48.4$ & $60.4$ \\
		    Concate & ${\bf 95.1}$ & ${\bf 97.3}$ & ${\bf 98.6}$ & $99.2$ & $60.7$ & $69.8$ & $77.4$ & $83.6$ & ${\bf 98.7}$ & ${\bf 99.3}$ & ${\bf 99.6}$ & ${\bf 99.8}$ & $49.7$ & $65.3$ & $79.4$ & $88.9$ & $25.4$ & $34.6$ & $45.3$ & $57.2$ \\
		    MDL & $89.5$ & $94.1$ & $96.6$ & $98.3$ & $63.6$ & $72.6$ & $79.9$ & $86.0$ & $97.6$ & $98.8$ & $99.2$ & $99.6$ & $58.9$ & $72.9$ & $84.1$ & $92.2$ & $31.6$ & $42.0$ & $53.4$ & $64.8$ \\
		    Simple CNAPS~\cite{bateni2020improved} & $79.9$ & $86.9$ & $92.6$ & $96.2$ & $65.2$ & $73.8$ & $81.1$ & $86.6$ & $97.5$ & $98.8$ & $99.3$ & $99.7$ & ${\bf 66.2}$ & ${\bf 79.3}$ & ${\bf 88.5}$ & ${\bf 94.7}$ & $33.2$ & $44.2$ & $57.3$ & $68.7$ \\
		    Ours & $87.9$ & $93.0$ & $96.1$ & $98.2$ & ${\bf 67.4}$ & ${\bf 76.3}$ & ${\bf 83.0}$ & ${\bf 88.5}$ & $97.0$ & $98.4$ & $99.1$ & $99.5$ & $62.1$ & $76.5$ & $86.0$ & $93.3$ & ${\bf 35.1}$ & ${\bf 46.1}$ & ${\bf 57.8}$ & ${\bf 69.0}$ \\
			\bottomrule
		\end{tabular}%
			}
		\caption{Global retrieval performance on MetaDataset (unseen datasets). In addition to few-shot learning experiments, we evaluate our method in a non-episodic retrieval task to further compare the generalization ability of our universal representations. }
		\label{tab:recallunseen}
\end{table*}

\begin{table*}[ht]
	\centering
    \resizebox{1.0\textwidth}{!}
    {
		\begin{tabular}{cccccccccc}

		    \toprule
		    \multirow{2}{*}{Test Dataset} & Proto-MAML & BOHB-E & CNAPS & Simple CNAPS & SUR & URT & \multirow{2}{*}{Best SDL} & \multirow{2}{*}{MDL} & \multirow{2}{*}{Ours} \\
		     & \cite{triantafillou2019meta} & \cite{saikia2020optimized} & \cite{requeima2019fast} & \cite{bateni2020improved} & \cite{dvornik2020selecting} & \cite{liu2020universal} & & & \\
		    \midrule
		    ImageNet & $46.5\pm1.1$ & $51.9\pm1.1$ & $50.8\pm1.1$ & $56.5\pm1.1$ & $54.5\pm1.1$ & $55.0\pm1.1$ & $54.3\pm1.1$ & $52.9\pm1.2$ & ${\bf 57.5\pm1.1}$  \\
			Omniglot & $82.7\pm1.0$ & $67.6\pm1.2$ & $91.7\pm0.5$ & $91.9\pm0.6$ & $93.0\pm0.5$ & $93.3\pm0.5$ & $93.8\pm0.5$ & $93.7\pm0.5$ & ${\bf 94.5\pm0.4}$  \\
			Aircraft & $75.2\pm0.8$ & $54.1\pm0.9$ & $83.7\pm0.6$ & $83.8\pm0.6$ & $84.3\pm0.5$ & $84.5\pm0.6$ & $84.5\pm0.5$ & $84.9\pm0.5$ & ${\bf 88.6\pm0.5}$  \\
			Birds & $69.9\pm1.0$ & $70.7\pm0.9$ & $73.6\pm0.9$ & $76.1\pm0.9$ & $70.4\pm1.1$ & $75.8\pm0.8$ & $70.6\pm0.9$ & $79.2\pm0.8$ & ${\bf 80.5\pm0.7}$  \\
			Textures & $68.2\pm0.8$ & $68.3\pm0.8$ & $59.5\pm0.7$ & $70.0\pm0.8$ & $70.5\pm0.7$ & $70.6\pm0.7$ & $72.1\pm0.7$ & $70.9\pm0.8$ & ${\bf 76.2\pm0.7}$  \\
			Quick Draw & $66.8\pm0.9$ & $50.3\pm1.0$ & $74.7\pm0.8$ & $78.3\pm0.7$ & $81.6\pm0.6$ & $82.1\pm0.6$ & ${\bf 82.6\pm0.6}$ & $81.7\pm0.6$ & $81.9\pm0.6$  \\
			Fungi & $42.0\pm1.2$ & $41.4\pm1.1$ & $50.2\pm1.1$ & $49.1\pm1.2$ & $65.0\pm1.0$ & $63.7\pm1.0$ & $65.9\pm1.0$ & $63.2\pm1.1$ & ${\bf 68.8\pm0.9}$  \\
			VGG Flower & $88.7\pm0.7$ & $87.3\pm0.6$ & $88.9\pm0.5$ & $91.3\pm0.6$ & $82.2\pm0.8$ & $88.3\pm0.6$ & $86.7\pm0.6$ & $88.7\pm0.6$ & ${\bf 92.1\pm0.5}$  \\
			Traffic Sign & $52.4\pm1.1$ & $51.8\pm1.0$ & $56.5\pm1.1$ & $59.2\pm1.0$ & $49.8\pm1.1$ & $50.1\pm1.1$ & $47.1\pm1.1$ & $49.2\pm1.0$ & ${\bf 63.3\pm1.2}$  \\
			MSCOCO & $41.7\pm1.1$ & $48.0\pm1.0$ & $39.4\pm1.0$ & $42.4\pm1.1$ & $49.4\pm1.1$ & $48.9\pm1.1$ & $49.7\pm1.0$ & $47.3\pm1.1$ & ${\bf 54.0\pm1.0}$  \\
			MNIST & - & - & - & $94.3\pm0.4$ & ${\bf 94.9\pm0.4}$ & $90.5\pm0.4$ & $91.0\pm0.5$ & $94.2\pm0.4$ & $94.5\pm0.5$  \\
			CIFAR-10 & - & - & - & ${\bf 72.0\pm0.8}$ & $64.2\pm0.9$ & $65.1\pm0.8$ & $65.4\pm0.8$ & $63.2\pm0.8$ & $71.9\pm0.7$  \\
			CIFAR-100 & - & - & - & $60.9\pm1.1$ & $57.1\pm1.1$ & $57.2\pm1.0$ & $56.2\pm1.0$ & $54.7\pm1.1$ & ${\bf 62.6\pm1.0}$  \\
		    \midrule
		    Average Rank & $7.7$ & $8.0$ & $6.8$ & $4.8$ & $5.4$ & $4.2$ & $4.8$ & $4.8$ & $1.2$ \\
			\bottomrule
		\end{tabular}%
			}
		\caption{Comparison to baselines and state-of-the-art methods on MetaDataset. Mean accuracy, 95\% confidence interval are reported. The first eight datasets are seen during training and the last five datasets are unseen and used for test only. Average rank is computed according to first 10 datasets as some methods do not report results on last three datasets.}
		\label{apptab:currmethod}
\end{table*}%

\subsubsection{Effect of adaptors in knowledge distillation}
In this section, we evaluate our method with adaptors or without adaptors for aligning features when we use CKA for knowledge distillation. From \cref{apptab:lossf}, We can see that using adaptors can improve the performance, such as Birds (+1.7) and VGG Flower (+3.6), MSCOCO (+1.3). This indicates that the adaptors $A_{\theta}$ help align features between multi-domain and single-domain learning networks which are learned from very different domains.

\subsubsection{Complete results of varying-way five-shot and five-way one-shot}

We further analyze our method for 5-shot setting with varying number of categories.
To this end, we follow the setting in \cite{doersch2020crosstransformers}, compare our method to the best three state-of-the-art methods including Simple CNAPS, SUR and URT. In this setting, we sample a varying number of ways in MetaDataset the same as the standard setting but a fixed number of shots to form balanced support and query sets. The mean accuracy and 95\% confidence interval of our method and compared approaches are depicted in \cref{tabapp:fixedshot}. 
As shown in Table~\ref{tabapp:fixedshot}, overall performance for all methods decreases in most datasets compared to results in the conventional setting shown in Table 1 in the paper, indicating that this is a more challenging setting. It is due to that five-shot setting samples much less support images than the standard setting.
While both Simple CNAPS and SUR obtain 3.1 and 3.0 average rank, respectively. SUR performs the best on MNIST, Simple CNAPS outperforms others on CIFAR-10 and URT is top-1 on Quick Draw. 
Ours still achieves significant better performance than other methods on the rest ten datasets.

\paragraph*{Results in five-way one-shot setting.}
Next we test an extremely challenging five-way one-shot setting on MetaDataset.
For each task, only one image per class is seen as support set.
This setting is often used in evaluating different methods in a single domain~\cite{Lake1332,ren2018meta,vinyals2016matching}, while we adopt it for multiple domains.
As shown in Table~\ref{tabapp:fixedshot}, our method achieves consistent gain as observed in previous two settings, which validates the importance of good universal representations in case of limited labeled samples in meta-test. Interestingly, Simple CNAPS achieves better rank than SUR in this setting, which is opposite in previous settings.

\subsubsection{Results evaluated with updated evaluation protocol.}
As the code from MetaDataset has been updated, we evaluate all methods with the updated evaluation protocol from the MetaDataset~\footnote{As mentioned in \url{https://github.com/google-research/MetaDataset/issues/54}, we also set the shuffle\_buffer\_size as 1000 to evaluate all methods and report the results in \cref{apptab:currmethod}. This change does not affect much on the results as the datasets we used were shuffled using the latest data convert code from \href{https://github.com/google-research/MetaDataset}{MetaDataset}.} and report the results~\footnote{Results of Proto-MAML~\cite{triantafillou2019meta}, BOHB-E~\cite{saikia2020optimized}, and CNAPS~\cite{requeima2019fast} are obtained from \href{https://github.com/google-research/MetaDataset}{MetaDataset}. The results of Simple CNAPS~\cite{bateni2020improved} are reproduced by the authors and reported at \url{https://github.com/peymanbateni/simple-cnaps}. We reproduce the results of SUR~\cite{dvornik2020selecting} and URT~\cite{liu2020universal} with the updated evaluation protocol for fair comparison.} in \cref{apptab:currmethod}. As shown in \cref{apptab:currmethod}, the update does not affect much on the results and our method rank 1.2 in average and the state-of-the-art methods SUR and URT rank 5.4 and 4.2, respectively. More specifically, we obtain significantly better results than the second best approach on Aircraft (+4.1), Birds (+1.3), Texture (+4.1), and Fungi (+2.9) for seen domains and Traffic Sign (+4.1) and MSCOCO (+4.3). 
The results show that jointly learning a single set of representations provides better generalization ability than fusing the ones from multiple single-domain feature extractors as done in SUR and URT.
Notably, our method requires less parameters and less computations to run during inference than SUR and URT, as it runs only one universal network to extract features, while both SUR and URT need to pass the query set to multiple single-domain network.

\begin{table}[t]
	\centering
    \resizebox{0.42\textwidth}{!}
    {
		\begin{tabular}{cccc}

		    \toprule
		    Test Dataset & CTX~\cite{doersch2020crosstransformers} & TCNAPS~\cite{bateni2020enhancing} & Ours \\
		    \midrule
			ImageNet & ${\bf 62.8\pm1.0}$ & $57.9\pm1.1$ & $57.5\pm1.1$  \\
			Omniglot & $82.2\pm1.0$ & $94.3\pm0.4$ & ${\bf 94.5\pm0.4}$  \\
			Aircraft & $79.5\pm0.9$ & $84.7\pm0.5$ & ${\bf 88.6\pm0.5}$  \\
			Birds & ${\bf 80.6\pm0.9}$ & $78.8\pm0.7$ & $80.5\pm0.7$  \\
			Textures & $75.6\pm0.6$ & $66.2\pm0.8$ & ${\bf 76.2\pm0.7}$  \\
			Quick Draw & $72.7\pm0.8$ & $77.9\pm0.6$ & ${\bf 81.9\pm0.6}$  \\
			Fungi & $51.6\pm1.1$ & $48.9\pm1.2$ & ${\bf 68.8\pm0.9}$  \\
			VGG Flower & ${\bf 95.3\pm0.4}$ & $92.3\pm0.4$ & $92.1\pm0.5$  \\
			Traffic Sign & ${\bf 82.7\pm0.8}$ & $59.7\pm1.1$ & $63.3\pm1.2$  \\
			MSCOCO & ${\bf 59.9\pm1.0}$ & $42.5\pm1.1$ & $54.0\pm1.0$  \\
			MNIST & - & - & ${\bf 94.5\pm0.5}$  \\
			CIFAR-10 & - & - & ${\bf 71.9\pm0.7}$  \\
			CIFAR-100 & - & - & ${\bf 62.6\pm1.0}$  \\
			\bottomrule
		\end{tabular}%
			}
		\caption{Comparison to CrossTransformer (CTX) and TransductiveCNAPS (TCNAPS) on MetaDataset. Mean accuracy, 95\% confidence interval are reported. The first eight datasets are seen during training and the last five datasets are unseen and used for test only. Note that TCNAPS and CTX are \emph{not directly comparable} to our method.}
		\label{apptab:ctx}
\end{table}%

\subsubsection{Comparison to Cross-Transformer~\cite{doersch2020crosstransformers} and Transductive CNAPS~\cite{jiang2020transductive}.}
Here we compare our method to CTX~\cite{doersch2020crosstransformers} and TCNAPS~\cite{bateni2020enhancing}\footnote{Results of CTX and TCNAPS are from \url{https://github.com/google-research/MetaDataset}} in \cref{apptab:ctx}. 
Note that TCNAPS and CTX are \emph{not directly comparable} to our method.
TCNAPS extends the Simple CNAPS~\cite{bateni2020improved} to a more favorable \emph{transductive inference} setting and exploits the query set at test time which is in contrast to the inductive learning in our submission.
CTX~\cite{doersch2020crosstransformers} focuses on learning from \emph{a single domain} (ImageNet), while our method is proposed to learn a single set of universal representation from multiple domains.
In addition, CTX is built on a heavier network (ResNet-34) and larger resolution images ($224\times 224$) than the one (ResNet-18, $84\times 84$ images) in ours.
Nevertheless, as shown in \cref{apptab:ctx}, our method still outperforms TCNAPS and CTX on most of the domains (8 out of 10 and 5 out of 10 respectively).
Both the transductive learning in TCNAPS and the cross-attention mechanism in CTX are potentially orthogonal to our universal representation learning and thus can be incorporated to ours, while we leave this as future work.
We will include the results and detailed discussion in the final version.

\subsubsection{Qualitatively results}
\begin{figure}[ht]
\begin{center}
\includegraphics[width=0.98\linewidth]{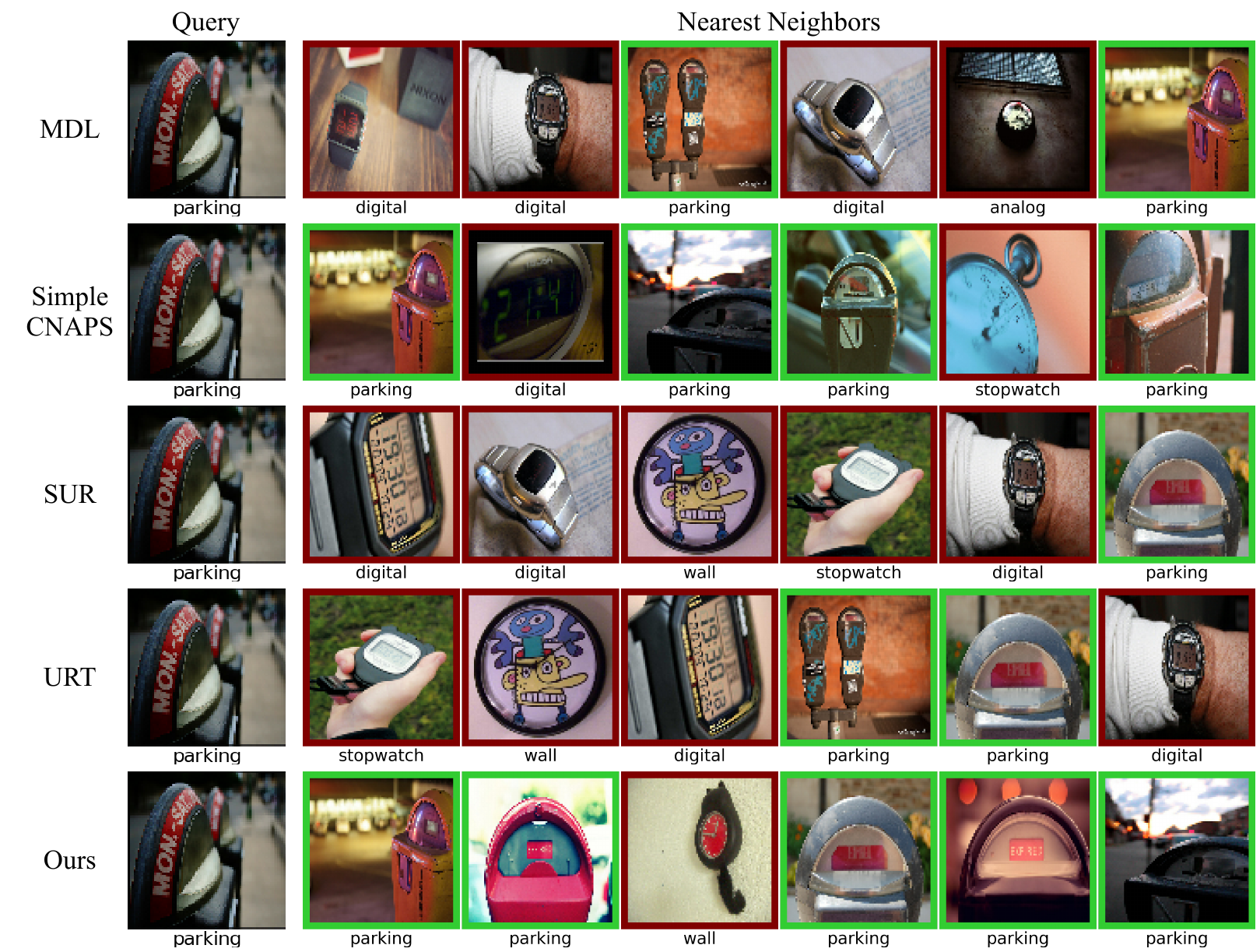}
\end{center}
\vspace{-0.3cm}
\caption{Qualitative comparison to MDL, Simple CNAPS~\cite{bateni2020improved}, SUR~\cite{dvornik2020selecting}, and URT~\cite{liu2020universal} in ImageNet. Green and red colors indicate correct and false predictions respectively.}
\label{fig:imagenet}
\end{figure}

\begin{figure}[ht]
\begin{center}
\includegraphics[width=0.98\linewidth]{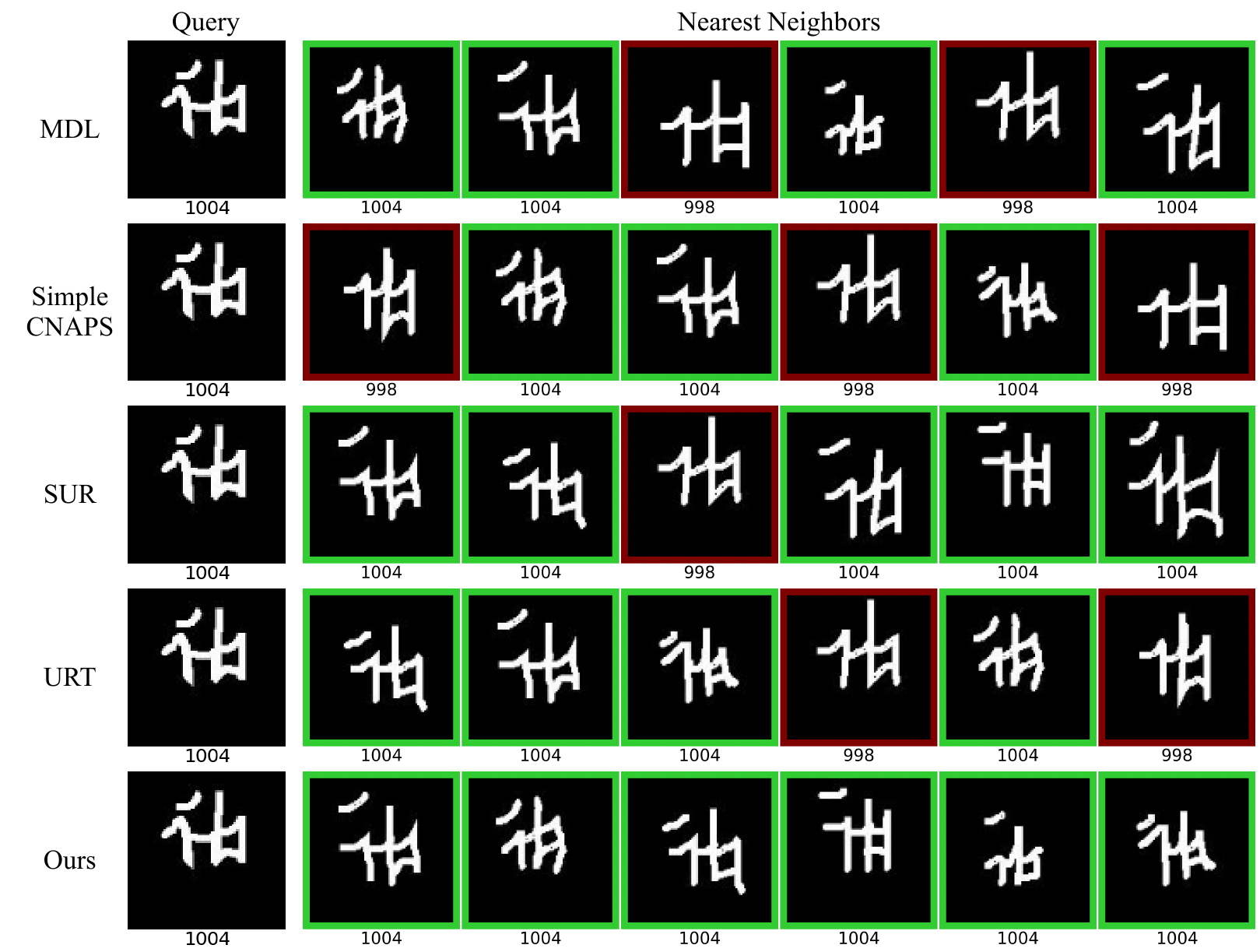}
\end{center}
\vspace{-0.3cm}
\caption{Qualitative comparison to MDL, Simple CNAPS~\cite{bateni2020improved}, SUR~\cite{dvornik2020selecting}, and URT~\cite{liu2020universal} in Omniglot. Green and red colors indicate correct and false predictions respectively.}
\label{fig:omniglot}
\end{figure}

\begin{figure}[ht]
\begin{center}
\includegraphics[width=0.98\linewidth]{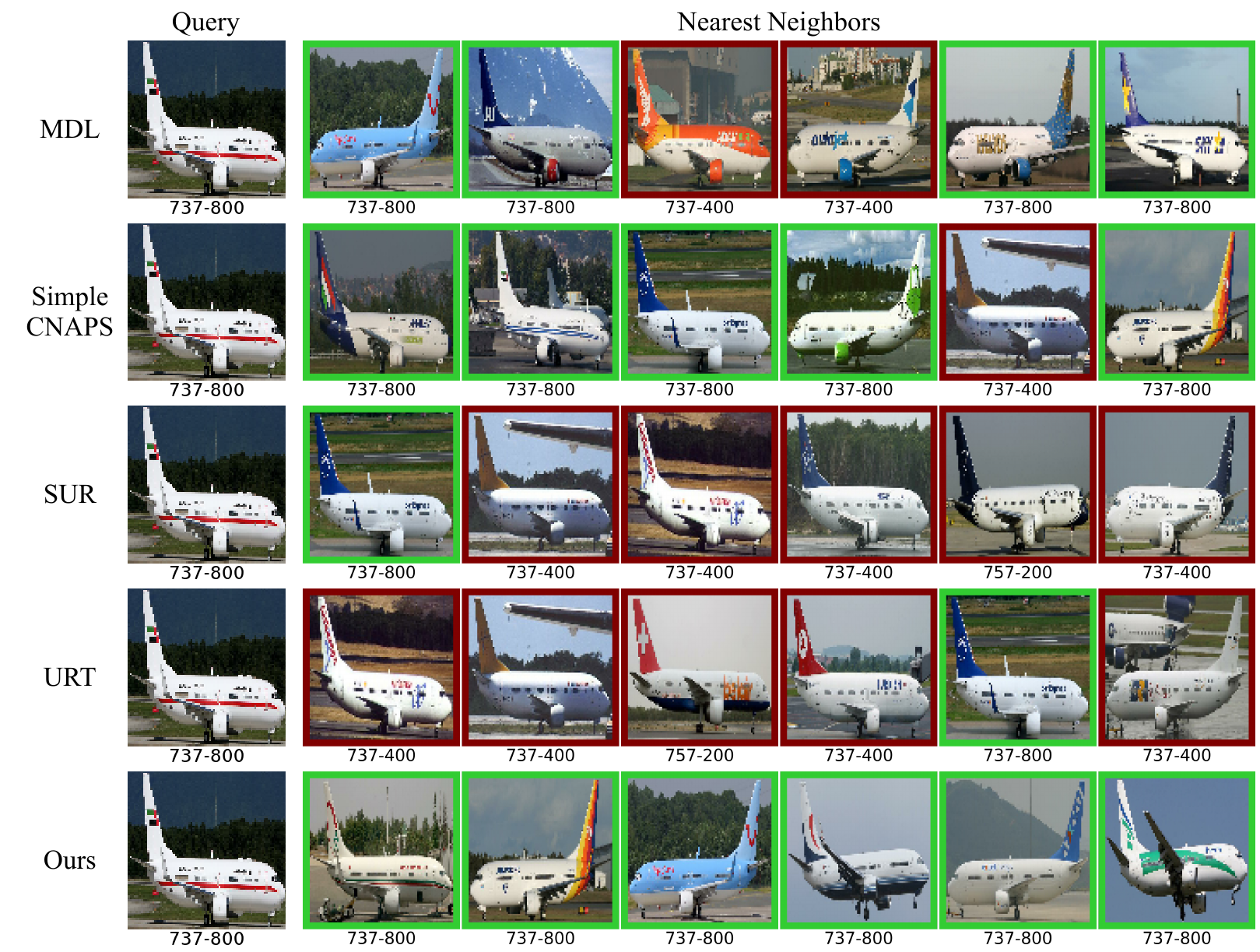}
\end{center}
\vspace{-0.3cm}
\caption{Qualitative comparison to MDL, Simple CNAPS~\cite{bateni2020improved}, SUR~\cite{dvornik2020selecting}, and URT~\cite{liu2020universal} in Aircraft. Green and red colors indicate correct and false predictions respectively.}
\label{fig:aircraft}
\end{figure}

\begin{figure}[ht]
\begin{center}
\includegraphics[width=0.98\linewidth]{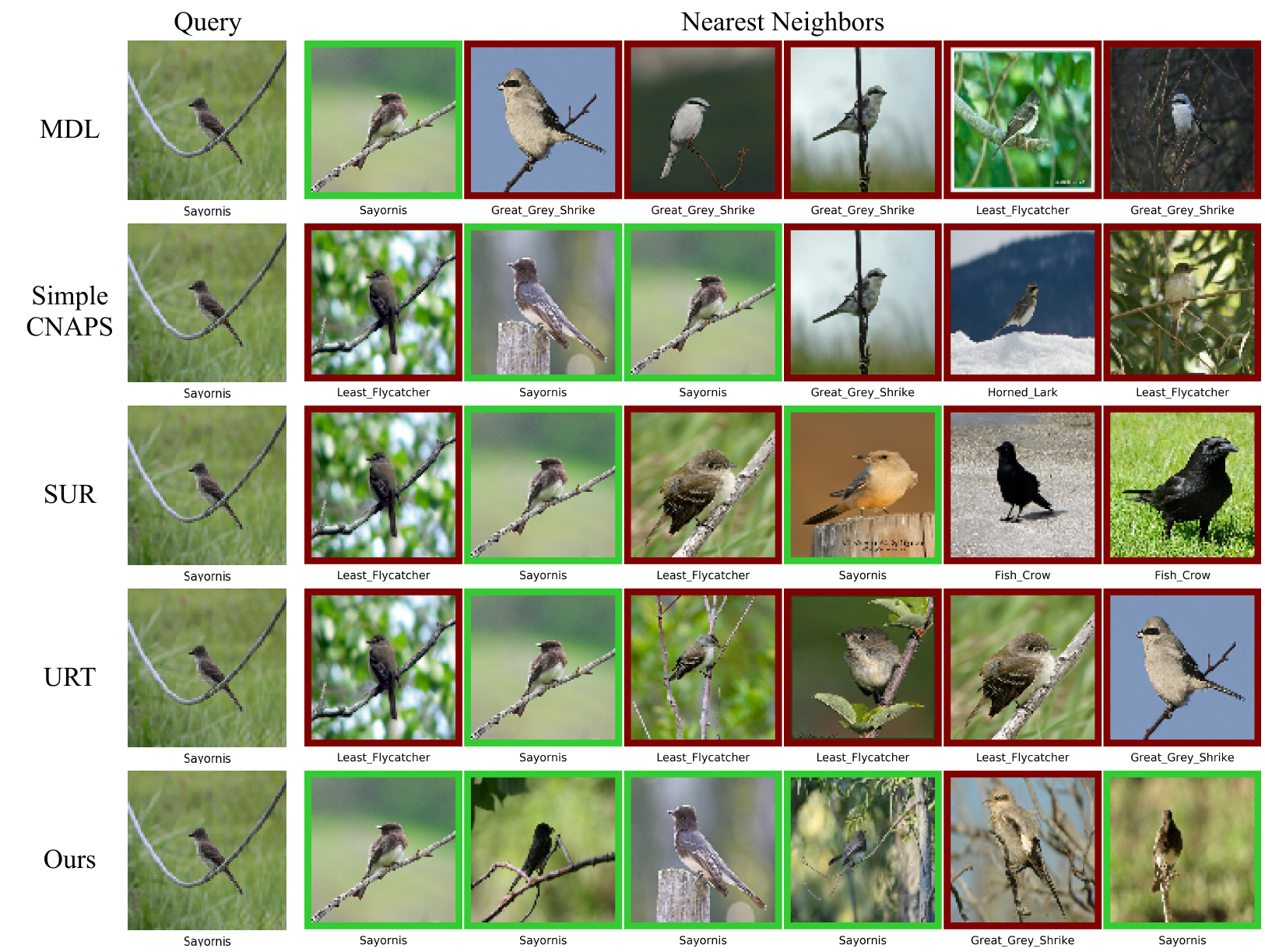}
\end{center}
\vspace{-0.3cm}
\caption{Qualitative comparison to MDL, Simple CNAPS~\cite{bateni2020improved}, SUR~\cite{dvornik2020selecting}, and URT~\cite{liu2020universal} in Birds. Green and red colors indicate correct and false predictions respectively.}
\label{fig:birds}
\end{figure}

\begin{figure}[ht]
\begin{center}
\includegraphics[width=0.98\linewidth]{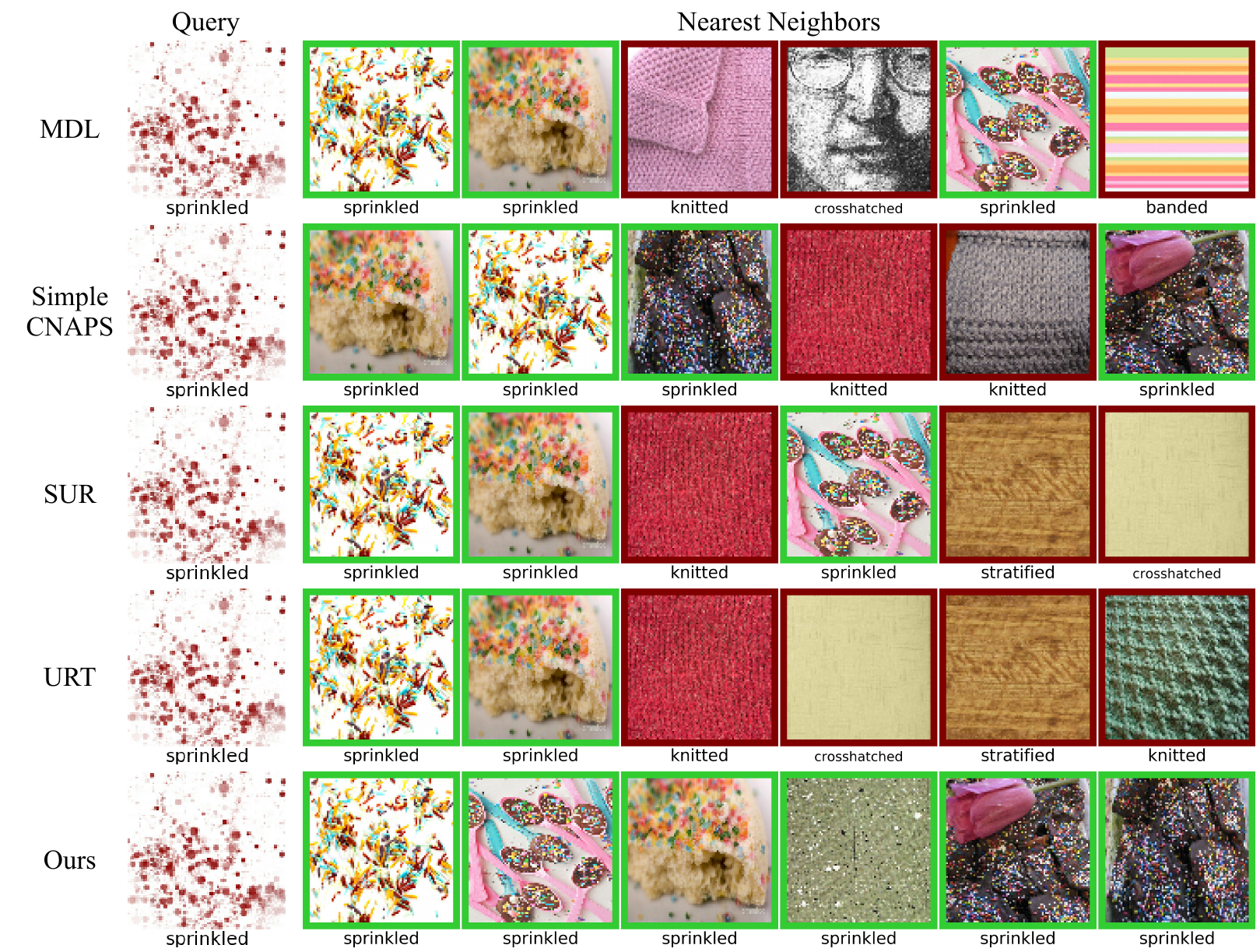}
\end{center}
\vspace{-0.3cm}
\caption{Qualitative comparison to MDL, Simple CNAPS~\cite{bateni2020improved}, SUR~\cite{dvornik2020selecting}, and URT~\cite{liu2020universal} in Textures. Green and red colors indicate correct and false predictions respectively.}
\label{fig:texture}
\end{figure}

\begin{figure}[ht]
\begin{center}
\includegraphics[width=0.98\linewidth]{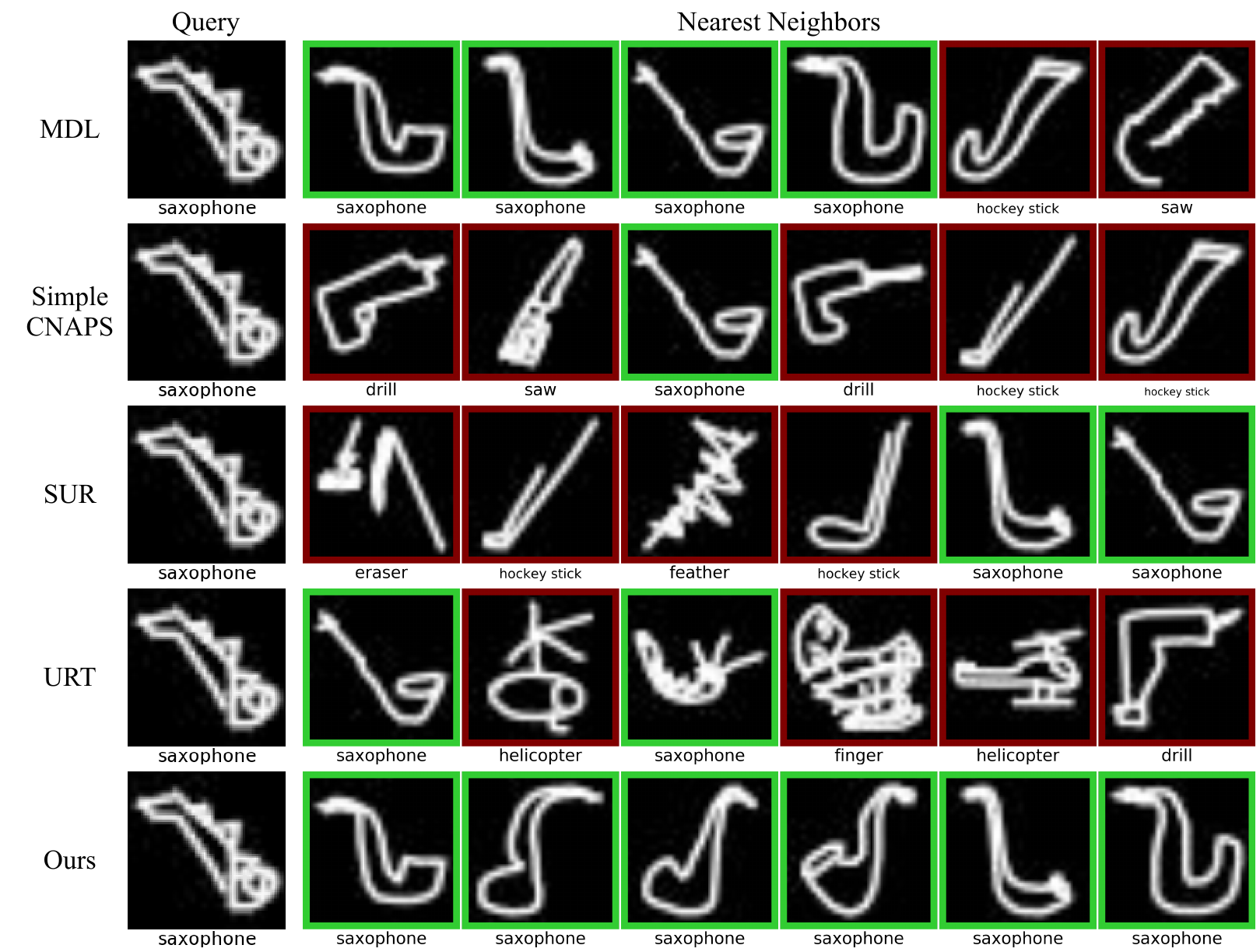}
\end{center}
\vspace{-0.3cm}
\caption{Qualitative comparison to MDL, Simple CNAPS~\cite{bateni2020improved}, SUR~\cite{dvornik2020selecting}, and URT~\cite{liu2020universal} in Quick Draw. Green and red colors indicate correct and false predictions respectively.}
\label{fig:quickdraw}
\end{figure}

\begin{figure}[ht]
\begin{center}
\includegraphics[width=0.98\linewidth]{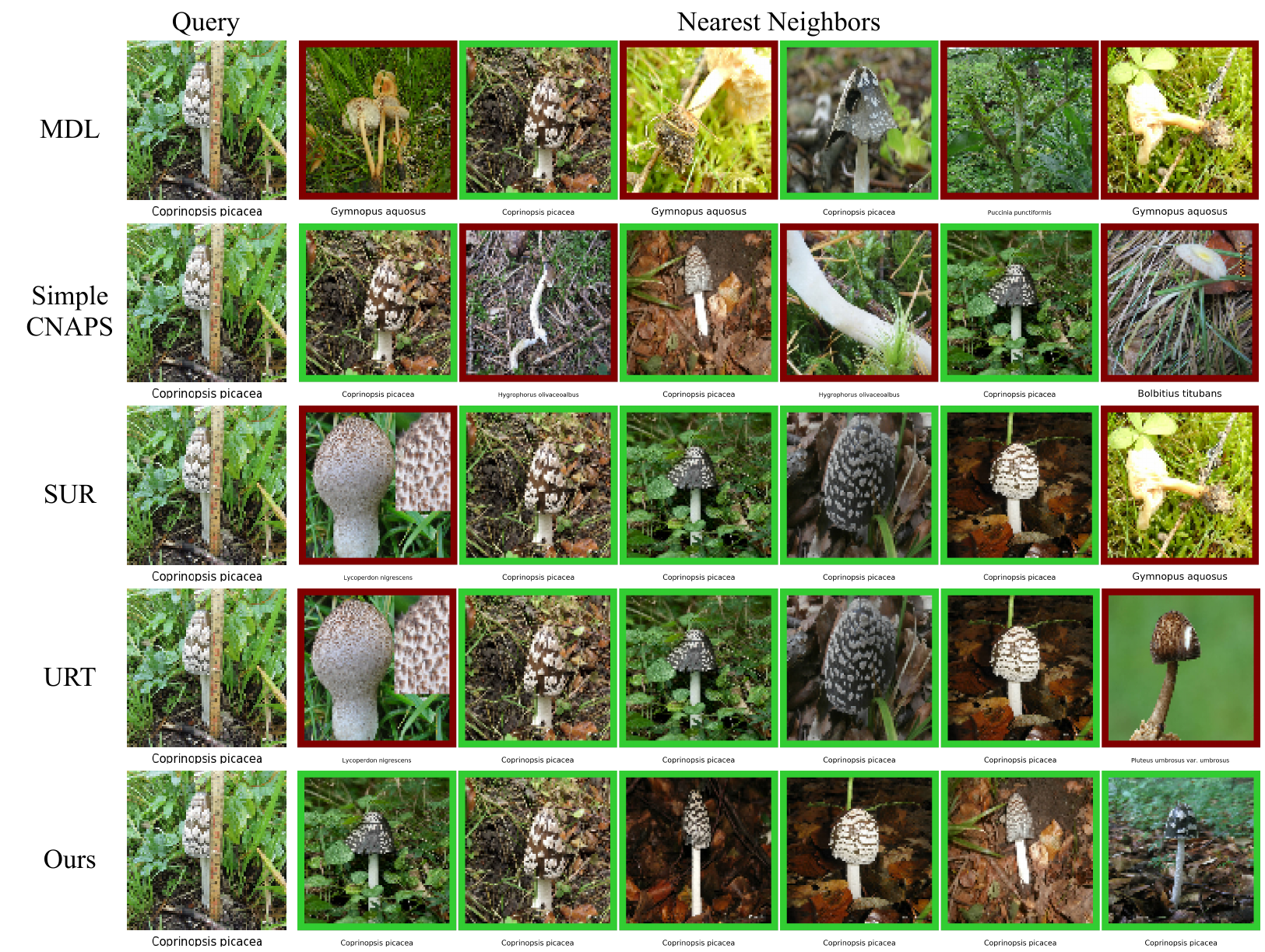}
\end{center}
\vspace{-0.3cm}
\caption{Qualitative comparison to MDL, Simple CNAPS~\cite{bateni2020improved}, SUR~\cite{dvornik2020selecting}, and URT~\cite{liu2020universal} in Fungi. Green and red colors indicate correct and false predictions respectively.}
\label{fig:fungi}
\end{figure}

\begin{figure}[ht]
\begin{center}
\includegraphics[width=0.98\linewidth]{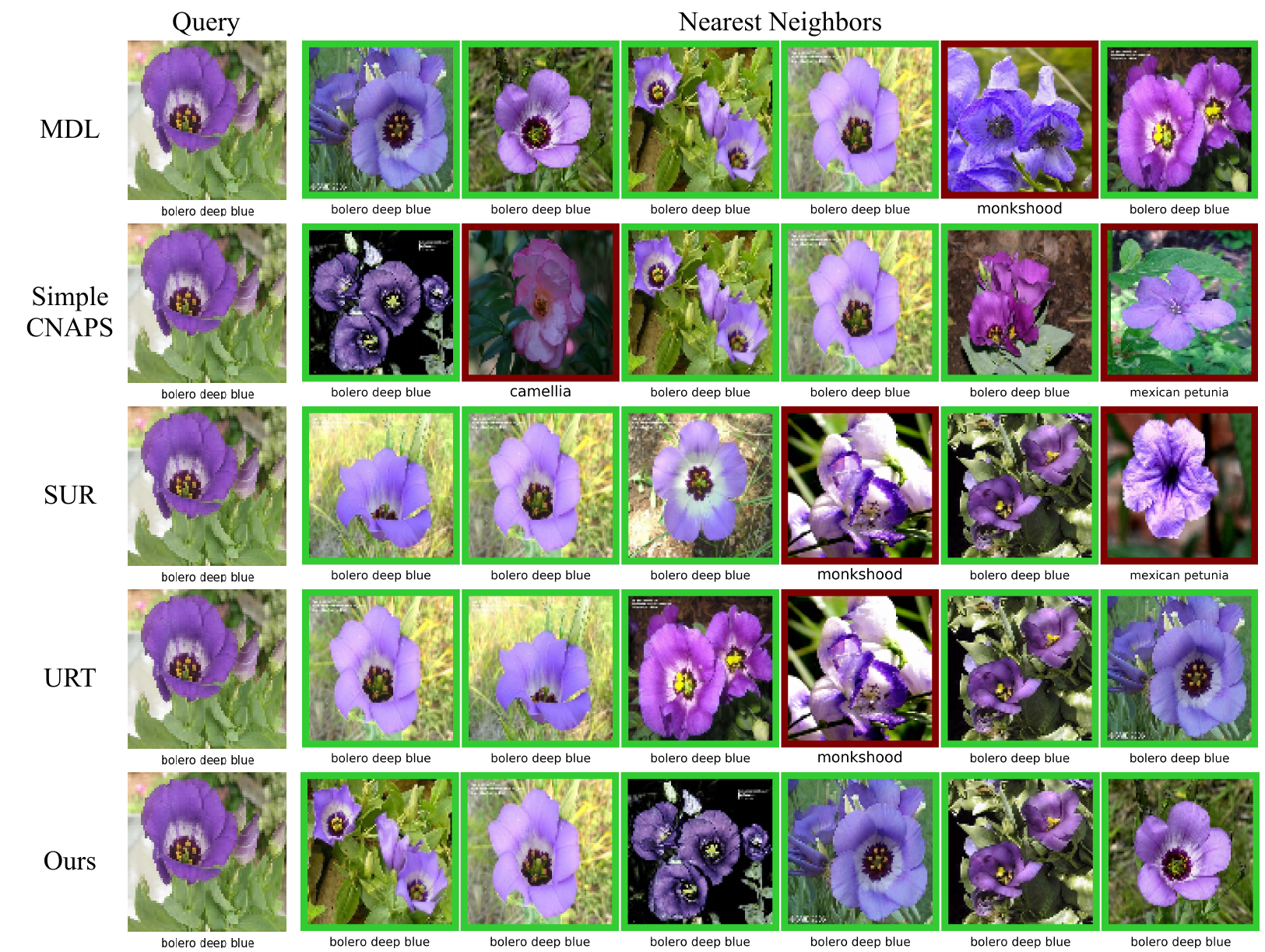}
\end{center}
\vspace{-0.3cm}
\caption{Qualitative comparison to MDL, Simple CNAPS~\cite{bateni2020improved}, SUR~\cite{dvornik2020selecting}, and URT~\cite{liu2020universal} in VGG Flower. Green and red colors indicate correct and false predictions respectively.}
\label{fig:flower}
\end{figure}

\begin{figure}[ht]
\begin{center}
\includegraphics[width=0.98\linewidth]{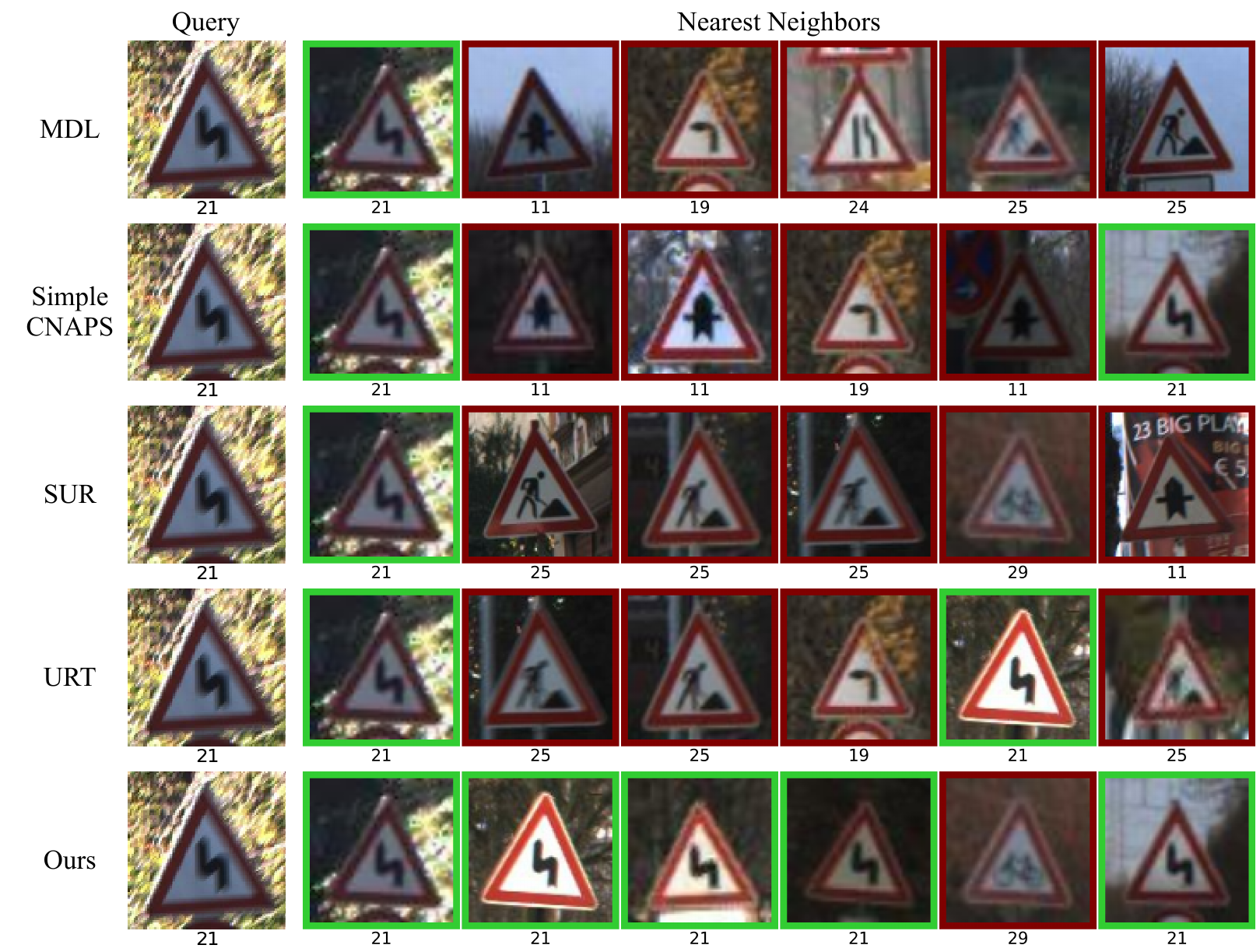}
\end{center}
\vspace{-0.3cm}
\caption{Qualitative comparison to MDL, Simple CNAPS~\cite{bateni2020improved}, SUR~\cite{dvornik2020selecting}, and URT~\cite{liu2020universal} in Traffic Sign. Green and red colors indicate correct and false predictions respectively.}
\label{fig:traffic}
\end{figure}

\begin{figure}[ht]
\begin{center}
\includegraphics[width=0.98\linewidth]{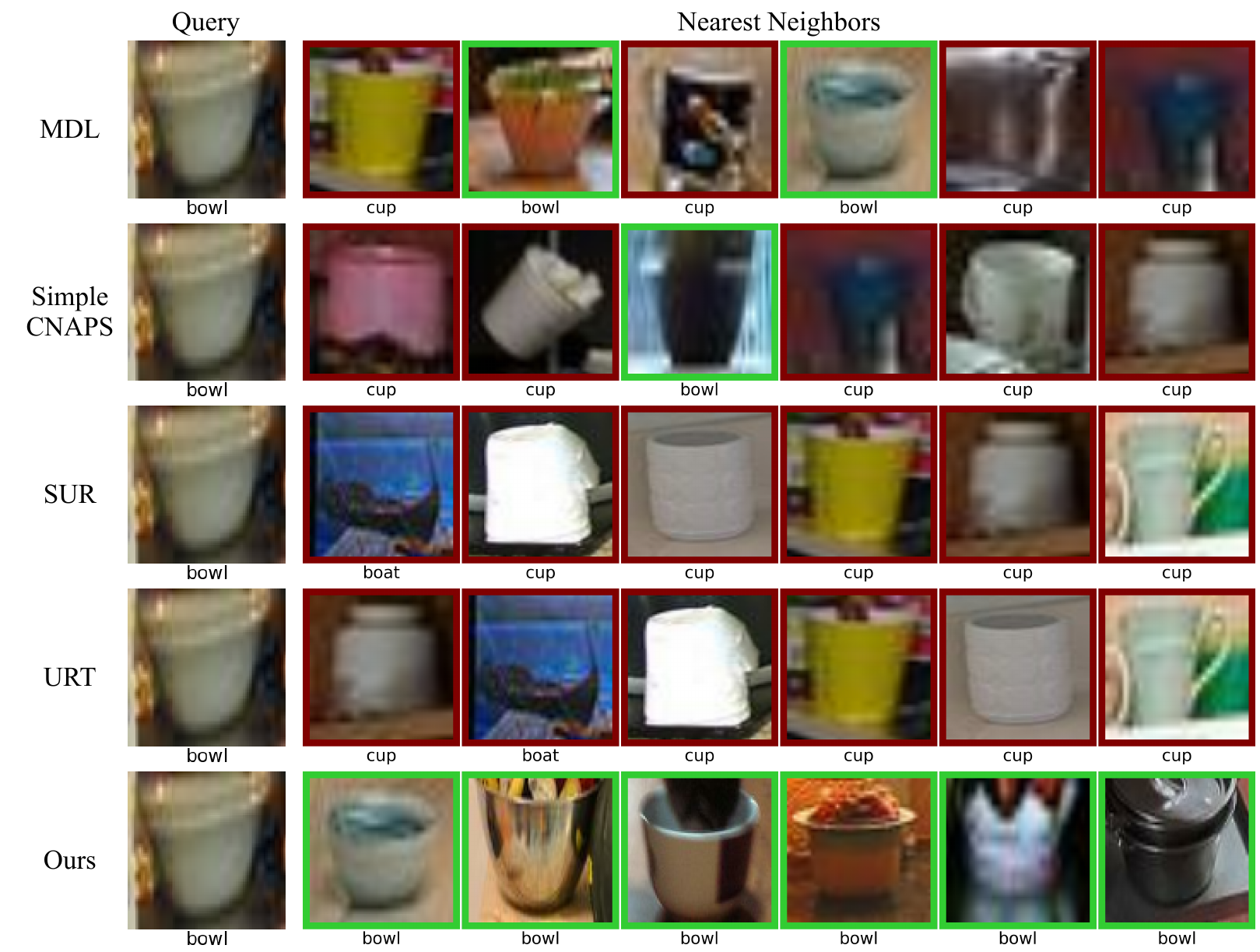}
\end{center}
\vspace{-0.3cm}
\caption{Qualitative comparison to MDL, Simple CNAPS~\cite{bateni2020improved}, SUR~\cite{dvornik2020selecting}, and URT~\cite{liu2020universal} in MSCOCO. Green and red colors indicate correct and false predictions respectively.}
\label{fig:mscoco}
\end{figure}

\begin{figure}[ht]
\begin{center}
\includegraphics[width=0.98\linewidth]{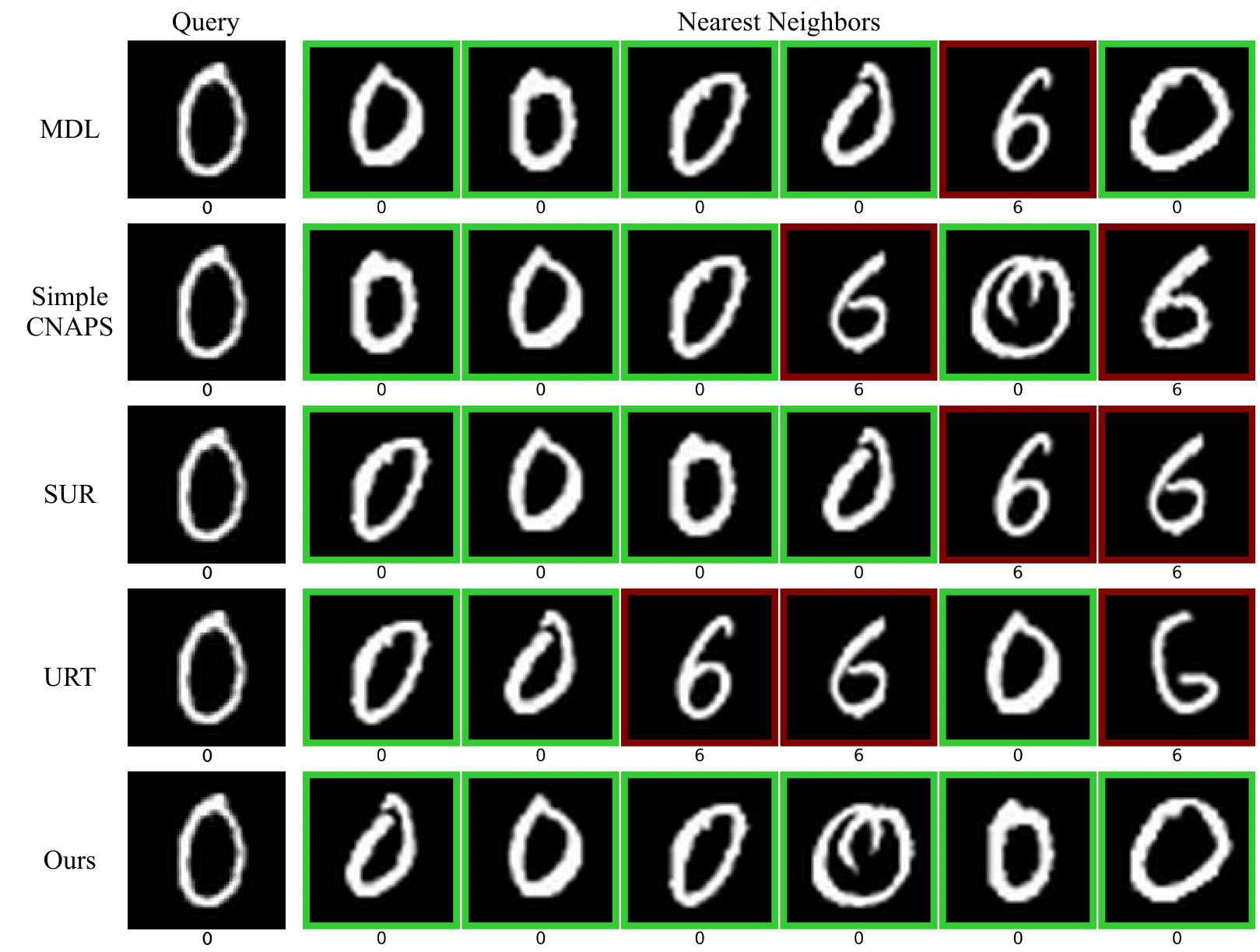}
\end{center}
\vspace{-0.3cm}
\caption{Qualitative comparison to MDL, Simple CNAPS~\cite{bateni2020improved}, SUR~\cite{dvornik2020selecting}, and URT~\cite{liu2020universal} in MNIST. Green and red colors indicate correct and false predictions respectively.}
\label{fig:mnist}
\end{figure}

\begin{figure}[ht]
\begin{center}
\includegraphics[width=0.98\linewidth]{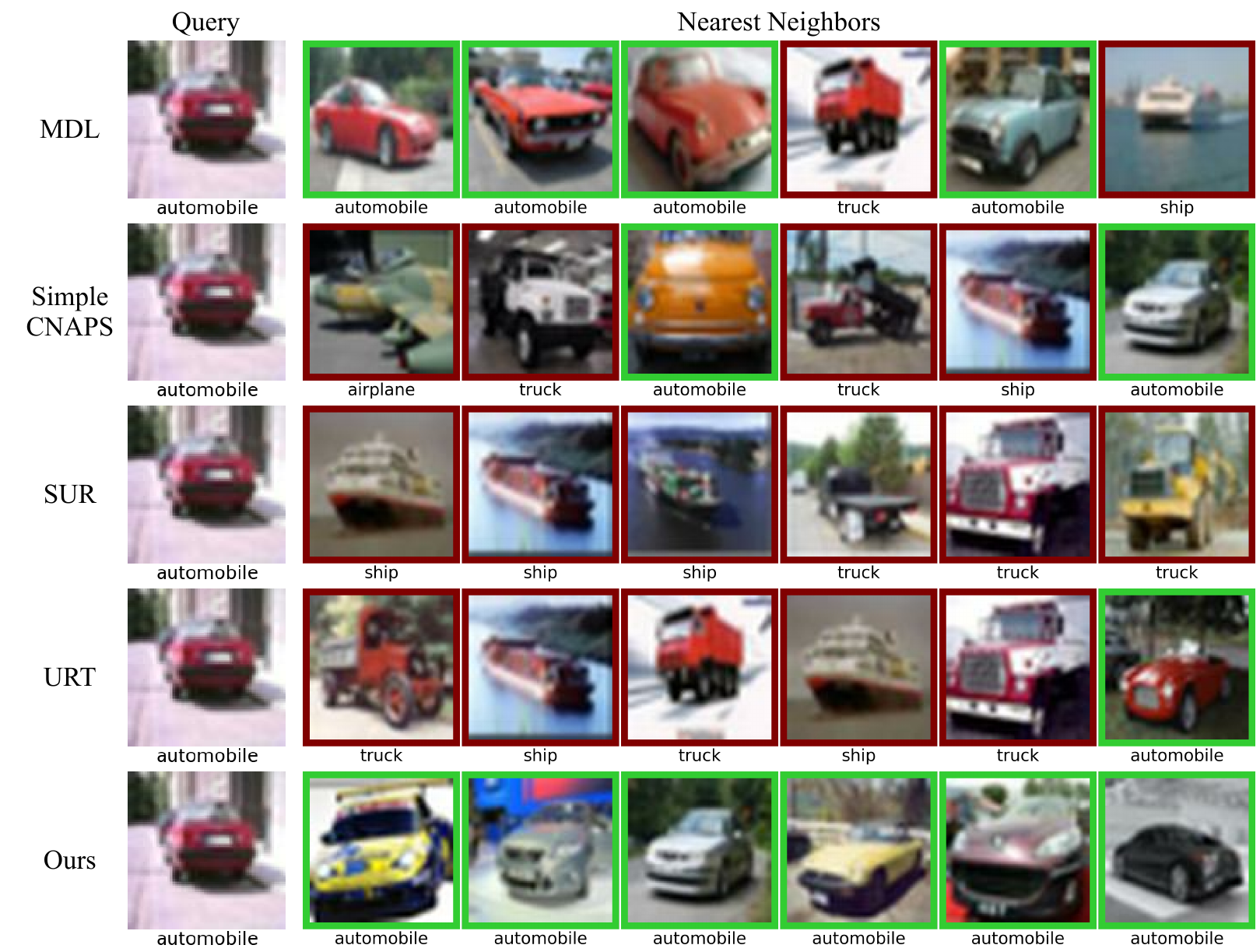}
\end{center}
\vspace{-0.3cm}
\caption{Qualitative comparison to MDL, Simple CNAPS~\cite{bateni2020improved}, SUR~\cite{dvornik2020selecting}, and URT~\cite{liu2020universal} in CIFAR-10. Green and red colors indicate correct and false predictions respectively.}
\label{fig:cifar10}
\end{figure}

\begin{figure}[ht]
\begin{center}
\includegraphics[width=0.98\linewidth]{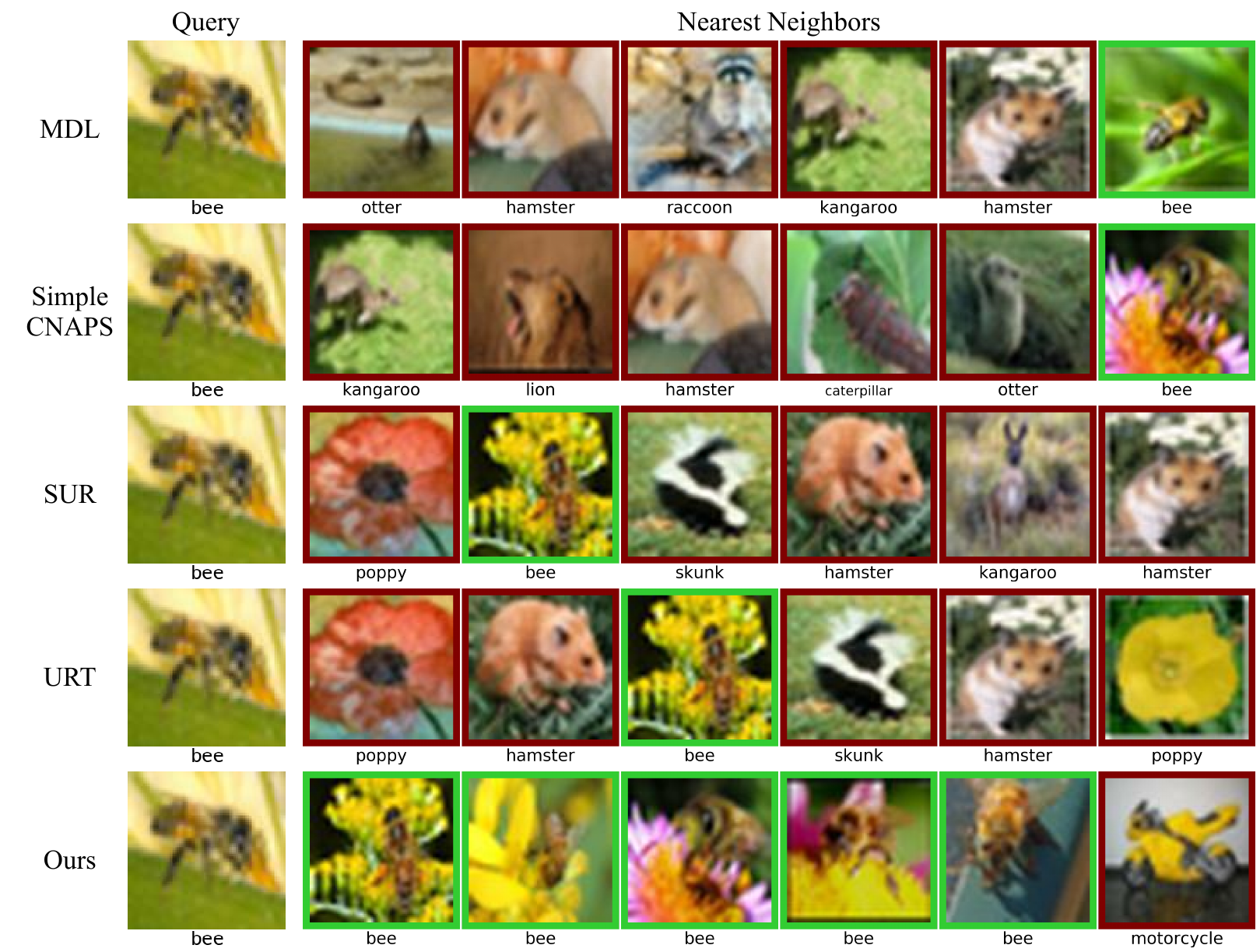}
\end{center}
\vspace{-0.3cm}
\caption{Qualitative comparison to MDL, Simple CNAPS~\cite{bateni2020improved}, SUR~\cite{dvornik2020selecting}, and URT~\cite{liu2020universal} in CIFAR-100. Green and red colors indicate correct and false predictions respectively.}
\label{fig:cifar100}
\end{figure}

We qualitatively analyze our method and compare it to the vanilla multi-domain leanring (MDL) baseline, Simple CNAPS~\cite{bateni2020improved}, SUR~\cite{dvornik2020selecting} and URT~\cite{liu2020universal} in \cref{fig:imagenet,fig:omniglot,fig:aircraft,fig:birds,fig:texture,fig:quickdraw,fig:fungi,fig:flower,fig:traffic,fig:mscoco,fig:mnist,fig:cifar10,fig:cifar100} by illustrating the nearest neighbors in all test datasets given a query image.
It is clear that our method produces more correct neighbors than other methods. 
While other methods retrieves images with more similar colors, shapes and backgrounds, \eg in \cref{fig:traffic,fig:mscoco}, our method is able to retrieve semantically similar images. It again suggests that our method is able to learn more useful and general representations.

\subsubsection{Complete global retrieval results}
Here we go beyond the few-shot classification experiments and evaluate the generalization ability of our representations that are learned in the multi-domain network in a retrieval task, inspired from metric learning literature~\cite{oh2016deep,yu2019learning}.
To this end, for each test image, we find the nearest  images in entire test set in the feature space and test whether they correspond to the same category.
For evaluation metric, we use Recall@$k$ which considers the predictions with one of the $k$ closest neighbors with the same label as positive.
In \cref{tab:recallseen,tab:recallunseen}, we compare our method with Simple CNAPS in Recall@1, Recall@2, Recall@4 and Recall@8. URT and SUR require adaption using support set and no such adaptation in retrieval task is possible, we replace them with two baselines that concatenate or sum features from multiple domain-specific networks. 
Our method achieves the best performance in ten out of thirteen domains with significant gains in Aircraft, Birds, Textures and Fungi.
This strongly suggests that our multi-domain representations are the key to the success of our method in the previous few-shot classification tasks.

\clearpage
\bibliographystyle{spmpsci}      
\bibliography{ref,ref_fsl}

\end{document}